\documentclass[a4paper, 10pt]{article}
\usepackage[english]{babel}
\usepackage{fullpage}
\usepackage[table,dvipsnames,x11names]{xcolor}
\usepackage{graphicx}
\usepackage{amsmath,amssymb,amsfonts,amsthm}
\usepackage{thmtools}
\usepackage[ruled,vlined]{algorithm2e}
\usepackage{esint}
\usepackage{quiver}
\usepackage{array}
\usepackage{multirow}
\usepackage{booktabs}
\usepackage{enumerate}
\usepackage{adjustbox}
\usepackage{setspace}

\usepackage{soul}
\soulregister\cite7
\soulregister\ref7
\soulregister\eqref7

\sethlcolor{accent-red!20}

\usepackage[hidelinks]{hyperref}
\hypersetup{
 colorlinks=true,
 citecolor=accent-blue,
 linkcolor=accent-blue,
 urlcolor=accent-blue,
} 

\usepackage{natbib}
\theoremstyle{plain}
\newtheorem{theorem}{Theorem}[section]
\newtheorem{proposition}[theorem]{Proposition}
\newtheorem{lemma}[theorem]{Lemma}

\theoremstyle{definition}
\newtheorem{definition}{Definition}

\newtheorem{assumption}{Assumption}

\newtheorem{remark}{Remark}

\newcommand{\lip}[1]{\operatorname{Lip}[#1]}

\newcommand{\leb}[1]{\operatorname{Leb}_{#1}}

\DeclareMathOperator*{\esssup}{ess\,sup}

\DeclareMathOperator*{\dist}{dist}
\DeclareMathOperator*{\conv}{conv}

\newcommand{\var}{\operatorname{Var}}
\newcommand{\E}{\mathbb{E}}
\newcommand{\R}{\mathbb{R}}
\newcommand{\N}{\mathbb{N}}
\newcommand{\C}{\mathbb{C}}

\newcommand{\X}{\mathcal{X}}

\renewcommand{\L}{\mathcal{L}}

\newcommand{\lx}{\ell_{x}}
\newcommand{\Lx}{\mathcal{L}_{x}}

\newcommand{\lf}{\ell_{f}}
\newcommand{\Lf}{\mathcal{L}_{f}}

\newcommand{\tlf}{\tilde{\ell}_{f,\Nt}}
\newcommand{\tLf}{\tilde{\mathcal{L}}_{f,N,\Nt}}

\newcommand{\lVM}{\ell_{f,\Nt}^{\rm{VM}}}

\newcommand{\LVM}{\mathcal{L}_{f,N,\Nt}^{\rm{VM}}}

\newcommand{\lTF}{\ell_{f,\Nt}^{\rm{TF}}}

\newcommand{\dt}{{\Delta t}}
\newcommand{\Nt}{{N_t}}

\renewcommand{\Re}{\operatorname{Re}}

\definecolor{accent-grey}{HTML}{000000}
\definecolor{accent-blue}{HTML}{3366cc}
\definecolor{accent-red}{HTML}{842a81}

\usepackage[most]{tcolorbox}

\newtcolorbox{emphbox}{
  breakable,
  colback=accent-grey!5!white,
  colframe=white,
  boxrule=0pt,
  arc=4pt,
  boxsep=5pt,
  left=5pt,
  right=5pt,
  top=4pt,
  bottom=4pt
}

\newtcolorbox{eqbox}{
  colback=accent-red!8!white,
  colframe=accent-red,
  boxrule=0.8pt,
  arc=4pt,
  boxsep=5pt,
  left=5pt,
  right=5pt,
  top=4pt,
  bottom=4pt
}

\title{Learning to evolve by matching velocities}

\author{
Nicola Farenga\thanks{MOX, Department of Mathematics, Politecnico di Milano, Milan, Italy.
} 
\and Andrea Manzoni\footnotemark[1]
}

\date{}

\begin{document}

\maketitle

\begin{abstract}
Learning the continuous-time evolution of nonlinear dynamical systems commonly relies on trajectory-based objectives, which require simulating the learned dynamics during training and backpropagating through the numerical scheme, at a cost that grows with the rollout horizon. 
We show that, beyond this cost, trajectory-based objectives degenerate as the rollout horizon increases, hindering gradient-based optimization, and that the numerical scheme biases the learned dynamics by constraining the spectrum of the learned Jacobian to its absolute stability region. 
To overcome these limitations, we introduce \textit{Velocity Matching} (VM), a simulation-free objective that relies on piecewise-linear interpolation and stochastic collocation, and approximates the velocity population risk to second order in the sampling step.
We develop a theoretical framework for velocity-based learning, introducing stability estimates that control trajectory errors in terms of velocity errors, thereby justifying simulation-free training via velocity supervision. These estimates are used to establish universal and continuous-time approximation results when the learned velocity field is parameterized by deep neural networks.
Building on a bias-variance decomposition of the VM objective, we show that its higher-order approximation of the velocity population risk yields a consistent minimizer; that its bias term relates to the residual of an A-stable continuous-stage Runge--Kutta scheme, thus connecting VM to implicit integration without requiring any implicit solve; and that it implicitly matches the action of the learned Jacobian to that of the true one along the observed trajectories, without requiring access to the underlying velocity field or its Jacobian.
Numerical experiments on nonlinear ordinary differential equations and semi-discretized time-dependent partial differential equations show that VM improves accuracy over trajectory-based objectives, while avoiding both the cost of simulation and the instabilities of long rollouts.
Empirical assessments of the continuous-time approximation and implicit Jacobian matching properties show that the models learned with VM can be simulated with schemes and step sizes different from those used during training without loss of accuracy, and that they preserve invariants and reproduce chaotic attractors that trajectory-based strategies fail to capture.

\bigskip

\noindent
\href{https://github.com/farenga/velocity-matching}{github.com/farenga/velocity-matching}
\end{abstract}

\newpage
\tableofcontents

\newpage
\section{Introduction}

\subsection{Background and motivation}

Recent advancements in the field of deep learning have enabled to address the fundamental problem of modeling the evolution of complex physical processes governed by nonlinear ordinary differential equations (ODEs) and time-dependent partial differential equations (PDEs), by building efficient surrogates that drastically reduce the computational costs required for high-fidelity simulations.
Although significant progress has been made in terms of architectures, ranging from graph-based models \citep{sanchez-gonzalez18a, pfaff21, lam23} to operator learning frameworks \citep{li21, kovachki23}, transformer-based architectures \citep{geneva22, li23b, herde24, bodnar25, holzschuh25}, and generative approaches \citep{price25, kohl26, kossaifi26}, autoregression remains the core modeling paradigm for handling temporal dependencies. Consequently, training objectives are typically based on trajectory supervision, thus requiring simulation of the learned dynamics during training.
\textit{Simulation-based} learning requires unrolling the learned dynamics over multiple time steps, which, from the neural ODEs perspective \citep{chen18, kidger22}, translates into the need to numerically integrate the learned dynamics over the rollout horizon, and relying on backpropagation through time (BPTT) or adjoint-based methods to compute gradients, thus introducing substantial computational overhead, often prohibitive for fine spatial and temporal discretizations, long time horizons, and large-scale models.
While rollout objectives are often employed to improve long-term predictive stability \citep{uy23}, undesired coupling effects between the learned dynamics and the integration scheme used during training may arise \citep{lu25}, shaping the loss landscape and inducing overfitting to the numerical method and the temporal discretization, thus undermining the continuous-time approximation capabilities of the learned model \citep{krishnapriyan23}.

\medskip

\textit{Simulation-free} alternatives commonly rely on teacher forcing (TF) schemes, in which autoregression is avoided during training by supervising the learned dynamics to predict the next state from the current ground truth, rather than regressing over multi-step rollouts. Such an approach, however, does not account for the distributional shift induced by error accumulation along inference-time rollouts, exposing a gap between one-step and multi-step objectives \citep{mccabe23, koehler24}, that has motivated the adoption of adversarial training strategies \citep{stachenfeld22, brandstetter22} and stabilization techniques for error correction \citep{pedersen25}.
Given the residual nature of autoregressive updates, commonly reading as explicit Euler updates, velocity supervision represents a natural alternative to trajectory-based objectives \citep{zhou25}, reframing the learning problem as a regression over velocity fields. However, velocity-based learning is hindered by the inaccessibility of the true velocities, typically estimated from discrete observations via finite differences or interpolation techniques. In this context, \citet{hou26} extend flow matching \citep{lipman23, tong24} to the operator learning setting, regressing the neural approximation of the right-hand side of a semi-discretized PDE onto the estimated velocities obtained by differentiating higher-order spline interpolants fitted to the observed trajectories. This construction mirrors the simulation-based and simulation-free duality in modern dynamical approaches to generative modeling \citep{albergo25}, and empirically demonstrates improved long-term stability and time-resolution invariance capabilities. Yet the relationship between trajectory-based and velocity-based learning objectives still lacks a comprehensive characterization, leaving the theoretical justification for the effectiveness of velocity-based approaches an open question.

\begin{figure}[t]
    \centering
    \includegraphics[width=1\textwidth]{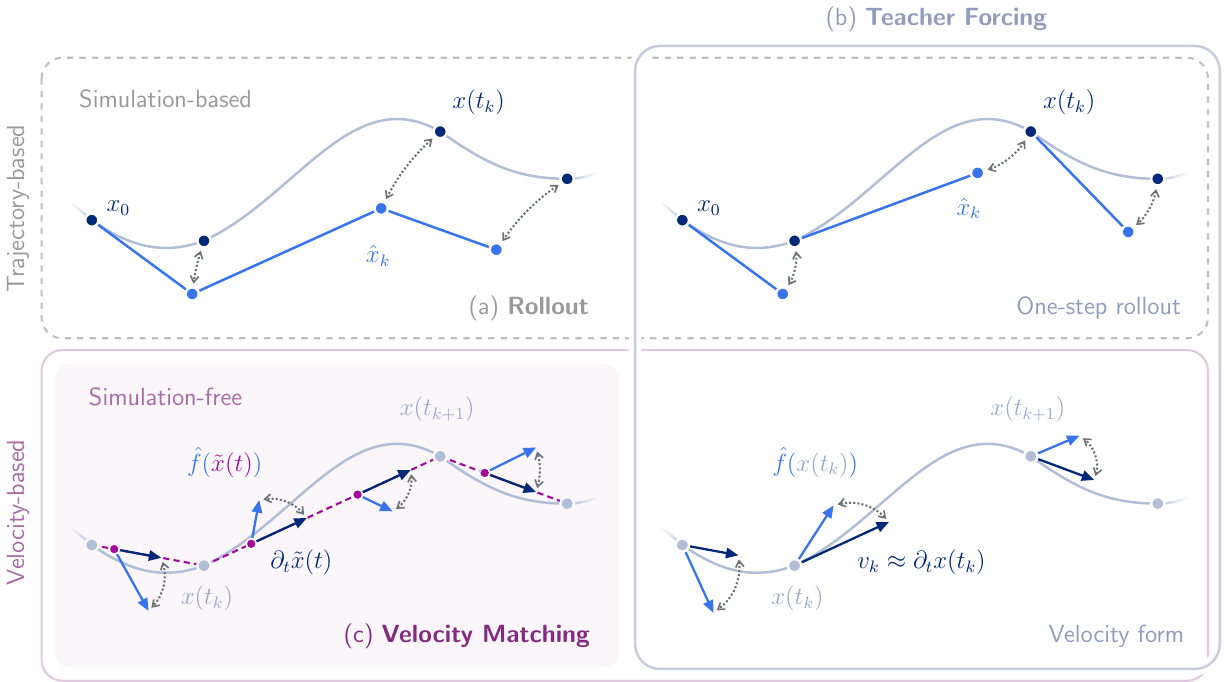}
    \caption{\textbf{Learning paradigms for dynamical systems.} 
    Rows separate trajectory-based (top) from velocity-based objectives (bottom).
    (a)~\textit{Rollout} training simulates the learned dynamics via numerical integration, and minimizes the discrepancy between the predicted  $\hat{x}_k$ and ground truth $x(t_k)$ states, with gradients backpropagated through the entire unrolled trajectory.
    (b)~\textit{Teacher forcing} spans both rows, as it admits either reading: it replaces multi-step rollouts with one-step predictions, avoiding backpropagation through time.
    As discussed in Section \ref{sec:trajectory-learning}, its loss can equivalently be read as a regression of the learned increment onto finite difference estimates $v_k \approx \partial_t x(t_k)$, at the observed states; the bottom panel shows the explicit Euler case, where the increment is $\hat{f}$ itself.
    (c)~\textit{Velocity Matching} instead  evaluates the learned velocity field $\hat{f}$ along the piecewise-linear interpolant $\tilde{x}(t)$ of the observations, at times $t$ sampled uniformly over the observation window, and regresses it onto the derivative $\partial_t \tilde{x}(t)$ of the interpolant.
    }
    \label{fig:objectives-comparison}
\end{figure}

\medskip
Whether simulation can be avoided during training by regressing directly onto velocity fields, while controlling the trajectory error and preserving continuous-time approximation properties, is a question that requires us to \emph{(i)} identify what separates trajectory-based from velocity-based learning, that is, how the two objectives relate; and \emph{(ii)} devise a suitable empirical objective that enables estimation of the velocity field while meeting these desiderata.
To this end we develop a theoretical framework for velocity-based learning, characterizing the velocity field of the underlying ODE as the minimizer of a quadratic functional over the space of admissible vector fields. We relate this objective to its trajectory-based counterpart by introducing suitable stability estimates mirroring the Lax equivalence theorem~\citep{lax56}; these estimates justify simulation-free training based on velocity supervision, and transfer approximation guarantees on the vector field to the flow it generates. 
Building on this analysis, we introduce the \textbf{Velocity Matching} (VM) objective, which combines stochastic collocation in time with piecewise-linear interpolation of the observed states to supervise the learned velocity field over the continuum of the time interval. By contrast, TF supervises the learned field at the observed states only, as illustrated in Figure~\ref{fig:objectives-comparison}. A further theoretical analysis identifies stochastic collocation as the mechanism behind three properties of VM: a higher-order approximation of the velocity population risk, an implicit regularization of the Jacobian along observed trajectories, and a correspondence with implicit integration schemes. The main contributions of this work and the organization of the paper are summarized next.

\subsection{Main contributions and organization} 

This work develops a theoretical framework for velocity-based learning of dynamical systems, which characterizes the gap between trajectory-based and velocity-based learning, and serves as the foundation of a simulation-free empirical objective that we introduce, named \textit{Velocity Matching}.

First, Section~\ref{sec:trajectory-learning} introduces the problem of learning dynamical systems in the form of ODEs, together with trajectory-based training objectives, ranging from multi-step rollouts to teacher forcing (TF), and discusses the challenges associated with simulation-based learning.
In particular, we study the objective defined over trajectories in a simplified continuous-time linear setting, where analytical expressions for the loss and its gradient can be derived. These expressions highlight the unsuitability of the objective for gradient-based optimization over long time horizons.
Starting from a similar setting and then extending to the more general nonlinear case, we discuss the role of the numerical scheme used to simulate the learned dynamics during training, showing how it shapes the resulting loss landscape and, consequently, constrains the spectrum of the Jacobian of the learned dynamics to the stability region of the scheme.

In Section~\ref{sec:velocity-learning}, we develop the theoretical framework for velocity-based learning that underlies the simulation-free objective introduced next, whose main contributions are summarized as follows:
\begin{itemize}
    \item In Section \ref{sec:velocity-learning-1}, we reframe the problem of learning dynamical systems in continuous-time as a regression over velocity fields, characterizing the true velocity field as the unique minimizer of a quadratic objective defined with respect to the occupation measure of the observed trajectories.
    
    \item In Section \ref{sec:velocity-learning-2}, we then show, by introducing a set of stability estimates,
    that this theoretical velocity objective controls the one defined over trajectories, up to a stability constant depending on the Lipschitz constant of the learned field, thus providing the theoretical justification for simulation-free learning. These estimates extend to a discrete-time setting, where the step size and the order of the numerical method employed to simulate the learned dynamics are explicitly accounted for, highlighting the role of numerical consistency and stability in the estimation of the underlying dynamics.
    
    \item Building on the stability estimates in continuous-time, in Section \ref{sec:velocity-learning-3} we provide a universal approximation result for flow maps associated with the true dynamics, when the learned velocity field is parameterized by deep neural networks (NNs) with smooth activation functions, such as GELU. 
    Relying on this result and a further error decomposition, we then show that the learned dynamics satisfies a \textit{continuous-time approximation} property, namely that it can be simulated, at inference, on arbitrarily fine temporal grids without loss of accuracy.
\end{itemize}
Section~\ref{sec:velocity-matching} addresses empirical estimation, building on the developed theoretical framework, by introducing the \textbf{Velocity Matching} (VM) objective. VM is a simulation-free objective that extends naive velocity regression by evaluating the learned velocity field at points sampled uniformly along the piecewise-linear interpolant of the observed trajectories, and by supervising it with the velocities obtained by differentiating the interpolant. As a velocity-based objective, VM does not require numerical integration of the learned dynamics.
Section \ref{sec:velocity-matching-1} introduces the objective and its algorithmic implementation. We then provide a theoretical analysis of its approximation properties, specifically:
\begin{itemize}
    \item In Sections~\ref{sec:velocity-matching-2} and~\ref{sec:velocity-matching-3}, we relate the VM loss to the residual of the average vector field method, an A-stable continuous-stage Runge--Kutta method, through a bias-variance decomposition of the VM objective with respect to the stochastic offset along the interpolant, so that VM inherits the properties of an implicit integration scheme without requiring any implicit solve. We then interpret the VM objective as a stochastic quadrature formula yielding an $O(\dt^2)$ approximation of the velocity population risk, whereas TF attains only an $O(\dt)$ approximation. Combined with the stability estimates of Section~\ref{sec:velocity-learning}, this result yields a simulation-free bound on the trajectory loss, together with the consistency of the VM minimizer as the step size vanishes.
    \item In Section~\ref{sec:velocity-matching-4}, we show that the VM objective admits an equivalent formulation as a regularized velocity regression problem, in which the bias term implicitly matches the action of the Jacobian of the learned velocity field to that of the true one along the flow, without requiring access to the latter, while the variance term penalizes the variation of the learned field along the flow. This characterization provides an interpretation of the improved long-term behavior of the learned dynamics, and connects VM to Jacobian matching approaches for chaotic systems.
\end{itemize}

In Section~\ref{sec:numerical-results}, we compare VM with TF and rollout-based strategies on nonlinear ODEs and semi-discretized time-dependent PDEs. Under a common iteration budget, VM attains errors lower than those of the baselines by more than an order of magnitude, at the training cost of TF, and trains stably where rollout objectives become unstable. A comparable reduction is observed for the Jacobian error, consistently with the implicit regularization result, although no Jacobian information is used during training. We further verify that the velocity fields learned by VM satisfy the continuous-time approximation property, and can thus be simulated with schemes and step sizes different from those used during training, whereas the models learned by trajectory-based strategies overfit both the scheme and the step size of the training discretization. On Hamiltonian and chaotic benchmarks, VM preserves the qualitative structure of the dynamics, namely the invariants and the chaotic attractors, which trajectory-based strategies fail to recover. Finally, we draw conclusions in Section~\ref{sec:conclusions}.

\subsection{Related work}

We review prior work on autoregressive surrogates for dynamical systems, whose residual updates admit a continuous-time interpretation, thus connecting them to neural ODEs and simulation-free velocity supervision. Within the latter, we consider two-stage approaches, flow matching, and collocation methods. We conclude by reviewing universal approximation for neural ODEs, together with continuous-time approximation results and analyses of the numerical bias induced by simulation-based training.

\paragraph{Autoregressive surrogates and velocity supervision.}
Autoregression is a fundamental paradigm for modeling temporal dependencies and underlies many recent deep learning approaches to dynamical systems and PDE surrogate modeling, regardless of the model architecture or specific application domain \citep{koehler24, herde24, bodnar25, holzschuh25, mccabe24}. 
While multi-step rollout objectives are employed to improve long-term predictive stability \citep{pathak22, lam23, uy23}, simulation-based training objectives might be computationally prohibitive to evaluate because of the high dimensionality of the state space, the long temporal horizons encountered in many applications, and the large parameter count of recent architectures. These factors motivate the introduction of simulation-free alternatives based on teacher forcing (TF). 
Such autoregressive updates commonly adopt a \textit{residual} formulation, so that the learned model is typically trained to predict the increment rather than the full state at the next time step, with the resulting update reading as an explicit Euler step \citep{mccabe25, kossaifi26}. Generalizing this residual update to higher-order Runge--Kutta (RK) schemes \citep{sanchez-gonzalez19} naturally leads to a neural ODE interpretation \citep{chen18, rackauckas21, kidger22}, so that the learned model can be interpreted as the velocity field governing the evolution of the system. This continuous-time interpretation, together with the residual structure of autoregressive updates, motivates velocity supervision as an alternative to trajectory-based objectives. In this context, \citet{zhou25} demonstrate the empirical effectiveness of velocity supervision, yielding a smoother optimization landscape and improved long-term stability. Velocity supervision, however, requires estimating the target velocities from discrete observations, leading to two-stage approaches that we discuss next.

\paragraph{Two-stage approaches, gradient matching and flow matching.} 
The reliance on prior velocity estimation from discrete state observations to supervise vector field learning, thus avoiding the simulation of the learned ODE, has its roots in the context of structural system identification \citep{bellman71, varah82, brunton16}. \textit{Two-stage} approaches, also known as \textit{gradient matching} methods \citep{ellner02}, first construct a sufficiently regular interpolant of the observed state trajectory from discrete observations and, in a second stage, regress the velocity field onto the velocities obtained by differentiating this interpolant.
For regression splines, the statistical properties of two-stage estimators, including consistency and convergence rates, have been established by \citet{brunel08}.
The two-stage approach was extended to neural ODEs by \citet{bonnaffe23} for nonparametric population dynamics modeling, where the interpolant is itself parameterized by a neural network rather than a spline.
The reliance on a prior interpolation step to supervise the learning of the underlying velocity field shares strong similarities with \textit{flow matching} approaches for simulation-free training of continuous normalizing flows \citep{lipman23, chen18}, where the neural ODE is trained by regressing onto a prescribed velocity field generating the conditional probability path. A related perspective has recently been explored for autoregressive neural operators by \citet{hou26}, but differs from our approach in two key respects. First, it relies on quintic spline interpolation, whereas we show that piecewise-linear interpolation suffices to attain a higher-order approximation of the velocity population risk. Second, it adopts a stochastic interpolant between consecutive observations~\citep{albergo25}, which perturbs the deterministic interpolant with isotropic noise according to a prescribed schedule. Notably, their own results indicate that isotropic noise injection is not essential in this setting, as the best performance is attained at low noise levels. This observation is consistent with our analysis, which identifies perturbations \textit{along} the interpolant, rather than in isotropic directions, as the mechanism underlying the higher-order accuracy and the implicit regularization of the VM objective, established in Sections~\ref{sec:velocity-matching-3} and~\ref{sec:velocity-matching-4}, respectively.

\paragraph{Collocation methods, Jacobian matching and chaotic dynamics.} 
Collocation approaches for system identification and parameter estimation in ODEs share similarities with the outlined two-stage approach, but reframe the problem under a regularized regression perspective \citep{ramsay07}. Specifically, \textit{generalized smoothing} techniques \citep{ramsay17} consists in jointly optimizing with respect to the interpolant and the velocity field parameters, rather than adopting a separate two-stage approach. In particular, the solution is approximated by a basis function expansion, whose coefficients are considered to be functions of the velocity field parameters, while a regularization term in the form of ODE residual, enforces velocity consistency and  acts as a smoothing. Variants of this technique have been explored in the context of neural ODEs \citep{shapovalova25}, alongside the use of collocation-based training approaches \citep{quaglino20, roesch21, norcliffe23}. Objectives and regularization terms that enforce consistency between the Jacobians of the learned and true velocity fields, termed \textit{Jacobian matching}, have been shown to improve the performance of neural ODEs on chaotic dynamics~\citep{park24, kang26}, enabling the recovery of ergodic properties of the target system, such as statistical moments and Lyapunov exponents. These results are closely related to our finding, reported in Section~\ref{sec:velocity-matching-4}, that the VM objective, through its stochastic collocation mechanism, implicitly enforces a directional form of Jacobian matching along the observed trajectories, without requiring access to the true Jacobian.

\paragraph{Universal approximation, continuous-time approximation and numerical bias.} 
Approximation properties of neural ODEs have been extensively investigated \citep{zhang20, li23, ruizbalet23, celledoni23}. Within the specific context of learning dynamical systems, \citet{li26} establish universal approximation properties for semi-autonomous neural ODEs, by leveraging universal approximation results for shallow NNs \citep{pinkus99}. Following a similar approximation framework, resting on the stability estimates we derive in Section \ref{sec:velocity-learning-2}, we extend this result to velocity fields parameterized by deep NNs, by drawing on recent approximation results for deep NNs in Sobolev norms \citep{guhring20, guhring21}. Universal approximation property of the learned dynamics, is closely related to the notion of continuous-time approximation (CTA), which concerns the ability of simulating the learned dynamics at arbitrary temporal resolutions, without the need for retraining, hence overcoming the limitations of fixed time-step autoregressive models. A first investigation appeared in \citet{krishnapriyan23}, with empirical evidence suggesting that the choice of the numerical method used during training governs the emergence of this property, improving with the order of the employed RK scheme. A theoretical framing was provided by \citet{farenga25}, in the context of latent neural ODEs for model order reduction of parameterized time-dependent PDEs, where a universal approximation result has been established and related to CTA, highlighting the need for a sufficiently small time discretization step size to guarantee such a property.
The ability to recover the underlying dynamics, and hence to achieve CTA, is however hindered by the numerical bias introduced by the integration scheme used during training. In this context, \citet{lu25} provide an extensive analysis relating underdamping and overdamping effects in the learned dynamics to the choice of the numerical method, tracing such effects back to the stability region of the employed scheme. The use of linear multistep methods has been addressed in \citet{keller21}, where a characterization of the notions of consistency, stability and convergence in the context of dynamics discovery is provided.

\subsection{Notation}

Throughout the paper, we denote state vectors by $x \in \R^n$, with $n\in \N$ being the dimension of the state space, and $|\cdot|$ the Euclidean norm. 
Regarding function spaces, $C^k(\Omega; \R^n)$ denotes the space of $k$-times continuously differentiable functions, from $\Omega\subseteq \R^m$ with values in $\R^n$. The space of $p$-integrable functions from $\Omega$ to $\R^n$ is denoted by $L^p(\Omega; \R^n)$, with $\|\cdot\|_{L^p(\Omega;\R^n)}$ its norm and $p\in[1,\infty]$. When a different measure $\mu$ over $\Omega$
 is considered, rather than the Lebesgue measure, it will be indicated as $L^p(\mu; \R^n)$.
Analogously, $W^{k,p}(\Omega; \R^n)$ denotes the Sobolev space of functions with weak derivatives up to order $k\in \N$ in $L^p$, with $\|\cdot\|_{W^{k,p}(\Omega;\R^n)}$ its norm. In particular, for $W^{k,\infty}$-norms, the sum convention is adopted \citep{evans10, brezis11}. 
The Lipschitz constant of a function $f:\Omega \to \R^n$ is denoted by $\lip{f}$. In the case of additional time dependence $f:[0,T] \times \Omega \to \R^n$, the Lipschitz constant with respect to the spatial variable, uniformly in time, is considered as
 $\lip{f} = \esssup_{t\in[0,T]} \lip{f(t,\cdot)}$. Finally, for a restriction of $f$ to a subset $K\subset \Omega$, we denote $\lip{f|_K} = \esssup_{t\in[0,T]} \lip{f(t,\cdot)|_K}$.
To indicate that a random variable $Z$ is distributed according to a probability distribution $\rho$, we write $Z\sim \rho$, and we denote by $\E_{Z\sim \rho}$ the expectation with respect to the distribution $\rho$. The pushforward of a measure $\mu$ through a measurable map $T$ is denoted by $T_\# \mu$. With a slight abuse of notation, the notion of measure and probability density are used interchangeably. Further notation, specifically related to the problem of learning dynamical systems, is introduced in the following sections as needed.

\section{Trajectory-based learning}
\label{sec:trajectory-learning}

This section formulates the problem of learning the dynamics in a continuous-time setting and introduces the trajectory-based learning objective in both theoretical and empirical settings. We then discuss the sources of approximation that arise in the latter, along with different learning strategies commonly adopted in the context of rollout-based training. Finally, the challenges of trajectory-based learning are addressed by studying the degeneracies of the trajectory objectives in both continuous-time and discrete-time, motivating a departure from this approach.

\subsection{Theoretical objective and problem formulation}

We proceed to formulate the problem of learning the dynamics of a nonlinear system in a continuous-time setting. In particular, nonlinear dynamical systems are considered in the form of first-order ordinary differential equations (ODEs), reading as
\begin{equation}
    \partial_t x(t,x_0) = f(t, x(t,x_0)), \quad x(0,x_0) = x_0,
    \label{eq:ode}
\end{equation}
with $x(t, x_0) \in \mathbb{R}^n$ being the solution at time $t \in [0,T]$,  associated to the initial condition $x_0 \in \X_0$, with $\X_0 \subset \R^n$ compact. Here, $T>0$ is a finite time horizon within which the solution is well-defined and remains contained in a compact set of $\R^n$, for any initial condition $x_0 \in \X_0$. The velocity field $f$ governing the dynamics is generally assumed to be continuous in time and locally Lipschitz in space, namely $f \in C^0([0,T]; W^{1,\infty}_{\rm loc}(\R^n; \R^n))$, to comply with the standard assumptions of Cauchy--Lipschitz (Picard--Lindel\"of) theory \citep{teschl12, brezis11}. The resulting flow map $x(t, \cdot): \X_0 \to \R^n$ then depends continuously on the initial condition. When the dependence on the initial time needs to be made explicit, we denote the solution of \eqref{eq:ode} as $x(t, t_0, x_0)$, with $t_0 \in [0,T]$ being the initial time. Having used a flow notation that makes the parameterization of the solution by the initial condition explicit, the time derivative of the solution is denoted by $\partial_t x(t,x_0)$, although, when it is clear from the context, we may also use the notation $\dot{x}$ for the sake of brevity.
We denote by $\X\subset \R^n$ the reachable set of states visited by the solution of \eqref{eq:ode} over the time interval $[0,T]$, starting from admissible initial conditions in $\X_0$, defined as
\begin{equation}
    \mathcal{X} := \{x(t,x_0) : t \in [0,T], \, x_0 \in \X_0\} \subset \R^n,
     \label{eq:reachable-set}
\end{equation}
compact by the above assumptions.
Hence, learning the dynamics of \eqref{eq:ode} in a continuous-time setting amounts to approximating the velocity field $f$ with $\hat{f}$, within a suitable class of admissible functions, solely relying on trajectory observations. The resulting learned flow $\hat{x}(t,x_0)$ is then obtained by integrating
\begin{equation}
    \partial_t\hat{x}(t,x_0) = \hat{f}(t, \hat{x}(t,x_0)), \quad \hat{x}(0,x_0) = x_0,
    \label{eq:neural-ode}
\end{equation}
which we refer to as the learned ODE.
In particular, when $\hat{f}$ is sought within the parametric class of neural networks, \eqref{eq:neural-ode} is referred to as a \textit{neural} ODE \citep{chen18,kidger22}. Assuming access to the full trajectories $x(t, x_0)$, and denoting by $\rho_0$ a probability distribution supported on $\X_0$, the learning problem can then be framed as a risk minimization problem over the space of admissible velocity fields $\hat{f}$, where the objective is the expected discrepancy between the true and learned trajectories, namely
\begin{equation}
    \L_x[\hat{f}] = \E_{x_0 \sim \rho_0} \ell_x(\hat{f};x_0), \quad \ell_x(\hat{f};x_0) = \frac{1}{T}\|x(\cdot,x_0)-\hat{x}(\cdot,x_0)\|^2_{L^2([0,T];\R^n)}.
    \label{eq:theoretical-trajectory-objective}
\end{equation}
which we refer to as the \textit{theoretical trajectory objective}.
The expectation argument $\ell_x(\hat{f};x_0)$ denotes the per-trajectory loss, defined as a mean $L^2$ error between the true $x(\cdot, x_0)$ and learned $\hat{x}(\cdot, x_0)$ trajectories over $[0,T]$. The dependence of the theoretical objective $\L_x[\hat{f}]$ on the learned model $\hat{f}$ is implicit through the learned solution $\hat{x}$, making the objective nonlinear in $\hat{f}$ even in the case of a linear dynamics, as discussed in Section~\ref{sec:loss-degeneracy-continuous-time}. As a notational remark, we use the subscript $x$ in $\L_x$ to indicate that the objective is defined over trajectories, as opposed to velocity-based objectives, denoted by $\L_f$ and introduced in Section \ref{sec:velocity-matching}, which directly supervise the learned velocity field $\hat{f}$ rather than the resulting trajectories.

\subsection{Empirical objective and learning strategies}
\label{sec:learning-strategies}

    When addressing the learning problem in an empirical setting, the theoretical trajectory objective $\L_x[\hat{f}]$ defined in \eqref{eq:theoretical-trajectory-objective} is commonly approximated by its empirical counterpart $\L_{x,N,\Nt}[\hat{f}]$, involving a three-fold approximation. First, the expected risk is approximated through a Monte Carlo estimate using a finite number $N$ of trajectories corresponding to independent realizations of the initial condition $x_0^{(i)}\sim\rho_0$, for $i=1,\dots,N$; second, the discrete-time nature of the available observations at time instances $\{t_k\}_{k=0}^{\Nt}$ induces an approximation of the $L^2$ norm over the time interval $[0,T]$ in $\ell_{x}$; third, the need to compute the mismatch between the true and learned trajectories at the discrete time instances $\{t_k\}_{k=0}^{\Nt}$, coupled with the nonlinear nature of $\hat{f}$, requires to rely on numerical integration of \eqref{eq:neural-ode} to simulate the learned dynamics and obtain $\hat{x}_k(x_0)\approx \hat{x}(t_k, x_0)$. Hence, the empirical counterpart of the theoretical trajectory objective reads as
\begin{equation}
    \L_{x,N,\Nt}[\hat{f}] = \frac{1}{N} \sum_{i=1}^N \ell_{x,\Nt}(\hat{f};x_0^{(i)}), \quad \ell_{x,\Nt}(\hat{f};x_0) = \frac{1}{\Nt} \sum_{k=1}^{\Nt} \left|x(t_k, x_0) - \hat{x}_k(x_0)\right|^2,
    \label{eq:rollout-loss}
\end{equation}
where subscripts $N$ and $\Nt$ in $\L_{x,N,\Nt}$ and $\ell_{x,\Nt}$ are used to indicate the dependence of the empirical objective on the finite number of trajectories and time instances, respectively.  
The need for simulating the learned dynamics, whether through a simple residual update or higher-order numerical schemes, leads to a \textit{rollout-based training} strategy, where the trajectory predictions are obtained by rolling out the learned model over the time horizon of interest, and model weights are updated by backpropagating gradients through the autoregressive steps, commonly referred to as \textit{backpropagation through time} (BPTT) \citep{werbos90}. 

While rollout-based training strategies have been widely adopted for learning autoregressive neural surrogates for dynamical systems, due to their ability to provide stable long-term behavior and mitigate error accumulation over time \citep{uy23, koehler24}, they are commonly associated to significant computational and memory costs, which are further induced by larger model sizes, often prohibitive for recently proposed large-scale neural surrogates \citep{mccabe25}. 
The issues associated with BPTT, arising from long rollouts and gradient degradation, have been widely studied in the context of recurrent neural networks (RNNs) \citep{bengio94, pascanu13, bengio12, bengio15, gruslys16}, and naturally carry over to the continuous-time ODE setting, which we elaborate on next. 
To mitigate such issues, mainly related to long rollout horizons, and to reduce memory and computational costs, several strategies have been proposed in the literature, which we briefly review in the following. In particular,
a \textit{multiple shooting} formulation of the problem is commonly adopted, where the full trajectory is split into sub-trajectories of length $L\leq\Nt$, yielding a per-trajectory loss of the form
\begin{equation}
    \ell_{x,\Nt,L}(\hat{f}; x_0) = \frac{1}{\Nt} \sum_{j=0}^{\frac{\Nt}{L}-1} \sum_{k=1}^L \left|x(t_{jL + k}, x_0) - \hat{x}_{jL + k}(x(t_{jL}, x_0))\right|^2,
    \label{eq:rollout-loss-multiple-shooting}
\end{equation}
with $L$ being the shooting length, which, for $L = \Nt$, reduces to the standard rollout loss over the full trajectory. The empirical risk in this setting is denoted by $\L_{x,N,\Nt,L}[\hat{f}]$.
Depending on the adopted shooting length $L$, different training strategies arise, which we briefly recall in the following. 

\paragraph{Multiple shooting ($L<\Nt$).} 
Originating in optimal control and system identification \citep{bock83,bock84,baake92}, \textit{multiple shooting} partitions the full trajectory into sub-trajectories of length $L \ll \Nt$, over which the loss is evaluated, possibly enforcing continuity constraints at the interval boundaries. A parallel development of multiple shooting has been proposed in the RNNs literature, under the name of \textit{truncated} BPTT \citep{werbos90,jaeger02,tallec17}, with recent extensions to neural ODEs \citep{turan22,iakovlev23}, including stochastic scheduling of the integration horizon \citep{ghosh20}. Common implementations of multiple shooting typically feature overlapping sub-trajectories, leading to a variance reduction effect \citep{beintema23}.
Beyond empirical considerations, \citet{ribeiro20} provide a theoretical justification for multiple shooting in discrete-time state-space models by analyzing the smoothness of the loss landscape as a function of the simulation horizon. In particular, they show that both the loss and its gradient admit Lipschitz constants that grow exponentially with the rollout length for non-contractive dynamics, thereby motivating the adoption of shorter horizons. In this view, in Section~\ref{sec:loss-degeneracy-continuous-time} we provide a complementary analysis of the loss landscape in a continuous-time linear setting, where explicit rates of vanishing and exploding gradients, as well as the sharpening of the loss minimum as a function of the rollout horizon, are derived.

\paragraph{Teacher forcing ($L=1$).} Setting $L=1$ in \eqref{eq:rollout-loss-multiple-shooting} yields the limiting case of multiple shooting, where the loss is computed over one-step predictions, with the model being fed with ground-truth observations at each time step, rather than its own predictions. This one-step strategy, known as \textit{teacher forcing} (TF) \citep{williams89}, ditches autoregression during training, and hence the need for backpropagation through the integrator, avoiding BPTT. In particular, when the autoregressive update is formulated as an explicit Euler step -- being the one commonly adopted in neural PDE surrogates -- TF admits an equivalent reading as a velocity-based objective, with the per-trajectory loss that can be rewritten as
\begin{equation}
    \ell_{x,\Nt,1}(\hat{f}; x_0) = \frac{\dt^2}{\Nt} \sum_{k=0}^{\Nt-1} \left|\frac{x(t_{k+1}, x_0) - x(t_k, x_0)}{\dt} - \hat{f}(t_k, x(t_k, x_0))\right|^2.
     \label{eq:teacher-forcing-velocity-formulation}
\end{equation}
In this view, TF represents a velocity-based learning strategy, where the learned velocity field $\hat{f}$ is supervised pointwise in time, with the target velocity obtained by means of finite differences approximation.
While the resulting per-step regression problem is well-conditioned, it introduces a discrepancy between training and inference regimes, known as \textit{exposure bias} \citep{ranzato16}. Since the model is never exposed to its own predictions during training, it is vulnerable to distribution shifts induced by error accumulation associated to autoregression at inference time, potentially resulting in unstable long-term behavior. Narrowing the training-inference gap has become a central issue in sequence modeling, motivating approaches such as \textit{scheduled sampling} \citep{bengio15, vlachas24} and \textit{professor forcing} \citep{goyal16}. 
In the specific context of dynamical systems and PDEs surrogate modeling, the exposure bias can be understood through the lens of \textit{zero-stability} \citep{quarteroni07, hairer08}, with distribution shifts causing the model to diverge from the solution manifold. Recent benchmarking studies \citep{koehler24,ohana24} have shown that models trained with one-step objectives struggle to perform autoregressive rollouts across different architectures, suggesting that the observed limitations are inherent to the training strategy, rather than the model architecture.
This observation motivates adversarial training strategies that promote stability of the learned dynamics \citep{sanchez-gonzalez18a, sanchez-gonzalez20a, pfaff21, stachenfeld22}, by evaluating the loss \eqref{eq:teacher-forcing-velocity-formulation} at perturbed states, either isotropically sampled from a Gaussian $\mathcal{N}(x(t_k, x_0), \sigma^2 I)$, whose variance $\sigma^2$ is a hyperparameter requiring tuning, or from the pushforward of the distribution of ground-truth states through the learned dynamics \citep{brandstetter22}, namely $\hat{x}(t_k, t_{k-1}, \cdot)_\# \rho(t_{k-1})$, with $\rho(t_{k-1})$ being the distribution of ground-truth states at time $t_{k-1}$.
Short rollout horizons, up to the TF limiting case $L=1$, are also adopted for learning chaotic dynamics,  where the Lyapunov time scale governs the length of the rollout horizon for which BPTT is feasible \citep{mikhaeil22}, motivating a \textit{sparsely forced} BPTT strategy, where the model is exposed to ground-truth observations at sparse time instances matching the Lyapunov time scale. Similarly, \textit{generalized teacher forcing} evaluates the model at convex combinations of observed and predicted states, with the mixing coefficient being scheduled during training \citep{hess23}. While different in their implementation, these schemes share a common reading as a TF strategy coupled with a state perturbation mechanism, which can be interpreted as a form of regularization, promoting stability of the learned dynamics. Such a perspective is shared by the \textbf{Velocity Matching} (VM) objective which we introduce in Section \ref{sec:velocity-matching}.

\paragraph{Curriculum learning ($L=1\to\Nt$).} 
A possible strategy, known as \textit{curriculum learning}, avoids committing to a fixed shooting length and instead schedules $L$ to progressively increase during training \citep{bengio09, bengio15}. Such schedules can be either deterministic or stochastic, and enable to interpolate between TF and single-shooting regimes, focusing on short-term predictions at the beginning of training, and gradually increasing the temporal horizon up to the full horizon $\Nt$ as training proceeds. The adoption of curriculum strategies has been often regarded as necessary to achieve stable long-term rollouts, showing improved performance in the context of autoregressive PDE surrogates \citep{takamoto23, koehler24}.  

\subsection{Degeneracies of the trajectory objective}
\label{sec:degeneracies-trajectory-objective}

Trajectory-based learning unrolls the learned dynamics to obtain model predictions over a time horizon $T$ and compares them with reference observations at discrete-time instances. The cost of long rollouts motivates the adoption of multiple shooting and teacher forcing strategies, which in turn give rise to the exposure bias issue, feeding the common belief that long rollouts are necessary to achieve stable long-term behavior. Empirical evidence, however, suggests a more nuanced picture: longer rollouts may induce training instabilities, leading to a less smooth loss landscape and higher-variance gradients \citep{metz22}, thus requiring the adoption of curriculum strategies to stabilize the learning process \citep{koehler24, list25}.
Here we build understanding on these issues from an optimization standpoint by analyzing the loss landscape of the trajectory objective both in continuous and discrete-time, highlighting its dependence on the target dynamics regime and temporal horizon $T$, as well as the role of the numerical scheme adopted for simulating the learned dynamics.
The analysis is deliberately carried out in a simplified linear setting, as it allows to derive explicit results thanks to closed-form derivations, and to isolate the analysis of the loss landscape from the complexity of the approximating function class.
In particular, in a continuous-time setting, it is shown that the trajectory objective exhibits vanishing and exploding gradients together with a sharpening of its minimum as $T \to +\infty$, with such degeneracies developing exponentially fast in $T$. In discrete-time, the interaction between the rollout loss \eqref{eq:rollout-loss} and the numerical integrator causes the landscape to collapse, in the infinite-horizon limit, onto the absolute stability region of the chosen numerical scheme, admitting a continuum of global minima, thereby hindering proper identification of the underlying system. Such a degeneracy translates into a constraint on the Jacobian spectrum of the learned dynamics in a nonlinear setting.

\subsubsection{Loss degeneracy in continuous-time} 
\label{sec:loss-degeneracy-continuous-time}

\begin{figure}[t]
\centering
\raisebox{4.75mm}{\includegraphics[width=.53\textwidth]{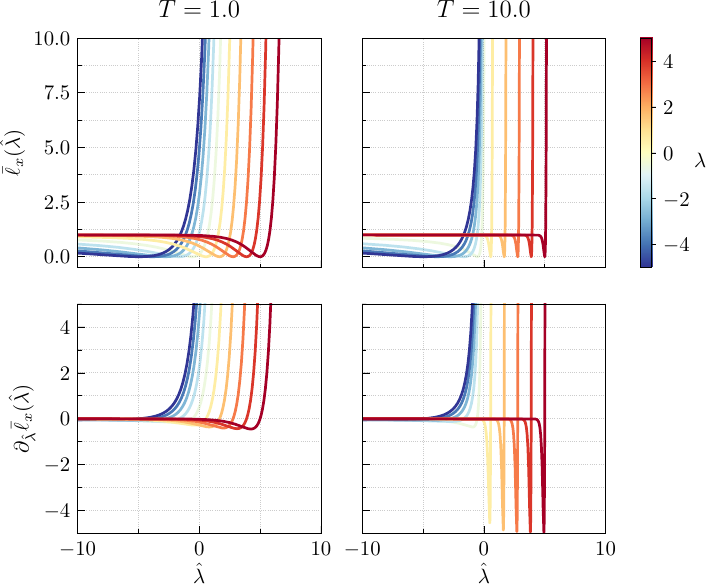}}
\hfill
\includegraphics[width=.462\textwidth]{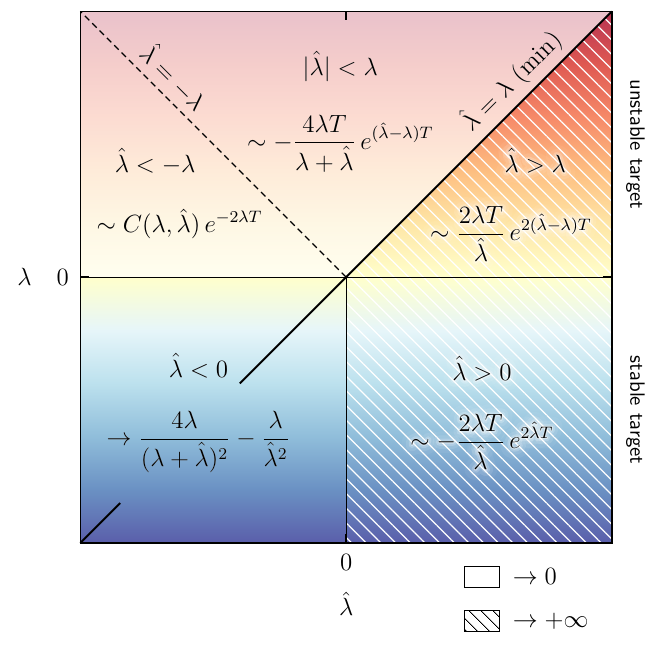}
\caption{\textbf{Trajectory loss landscape and gradient behavior.} (\emph{Left}) The energy-normalized trajectory objective $\bar{\ell}_x(\hat{\lambda})$ (top) and its gradient $\partial_{\hat{\lambda}}\bar{\ell}_x(\hat{\lambda})$ (bottom) for the scalar linear system $\dot{x}=\lambda x$, over learnable parameters $\hat{\lambda}\in[-10,10]$ and target parameters $\lambda\in[-5,5]$, at horizons $T=1$ and $T=10$.
(\emph{Right}) Asymptotic behavior of the gradient $\partial_{\hat{\lambda}} \bar{\ell}_x(\hat{\lambda})$ in the infinite-horizon limit $T \to +\infty$, over the $(\hat{\lambda},\lambda)$ plane, with the leading order behavior reported in each region.
In particular, the hatched right region indicates exploding gradients, while the top white faded region indicates vanishing gradients, both developing exponentially fast in $T$. In the remaining lower-left quadrant, the gradient converges to a $T$-independent finite value, vanishing along the line $\hat{\lambda} = \lambda$.
}
\label{fig:trajectory-loss-landscape}
\end{figure}

The theoretical trajectory objective \eqref{eq:theoretical-trajectory-objective}, despite being a quadratic functional in the solution $\hat{x}$, is nonlinear in $\hat{f}$ even in the case of linear dynamics, as the map $\hat{f}\mapsto \hat{x}$ implicitly defined by \eqref{eq:neural-ode} is nonlinear, thus yielding a non-convex loss landscape. Given the empirical evidence of training instabilities in the long rollouts regime, it is of interest to analyze how such a landscape depends on the regime of the target dynamics, as well as on the temporal horizon $T$, and to what extent these dependencies may hurt gradient-based optimization.
To quantify these considerations, we analyze the loss landscape of the theoretical trajectory objective, normalized by its average energy, which we denote by $\bar{\ell}_x(\hat{\lambda})$. A simplified linear setting is adopted, considering a scalar target system $\dot{x} = \lambda x$, with $\lambda \in \R$, to be learned from a fixed initial condition $x_0$ by the parametric family $\dot{\hat{x}} = \hat{\lambda} \hat{x}$. While not representative of the complexity of the general nonlinear case, such a linear setting is enough to expose the ill-suitedness of the trajectory objective to gradient-based optimization, allowing to provide explicit rates at which vanishing and exploding gradients develop, as well as the sharpening of the loss minimum as $T \to +\infty$. 
We defer detailed results and derivations to Appendix~\ref{app:continuous-time-loss-landscape-analysis}, while here we summarize the main findings.

\paragraph{Regime dependence.} For every fixed $T>0$, the loss landscape $\bar{\ell}_x(\hat{\lambda})$ is smooth and admits a unique global minimum at $\hat{\lambda} = \lambda$ at which $\bar{\ell}_x(\lambda) = 0$; it flattens for $\hat{\lambda} \to -\infty$, while it grows exponentially for $\hat{\lambda} \to +\infty$. The presence of a positive horizontal asymptote and of the vanishing minimum imply non-convexity. Our interest in long rollouts motivates the analysis of the infinite-horizon limit $T \to +\infty$, for which the loss exhibits a regime-dependent, discontinuous behavior. In the stable regime $\lambda < 0$, the loss landscape converges, for $\hat{\lambda}<0$, to a smooth non-convex profile, which diverges for $\hat{\lambda} \to 0^-$, while the loss blows up for $\hat{\lambda} \geq 0$. In the unstable regime $\lambda > 0$, the loss landscape converges to a discontinuous profile: constant for $\hat{\lambda} < \lambda$, vanishing at $\hat{\lambda} = \lambda$, and diverging for $\hat{\lambda} > \lambda$. Such a discontinuous profile for unstable dynamics, reported in Figure~\ref{fig:trajectory-loss-landscape} for $T=10$, denotes how ill-conditioning may develop in the presence of non-contractive dynamics, and motivates the adoption of shorter rollout horizons, as also suggested in \cite{ribeiro20}. Hence, in view of gradient-based optimization, it is worth discussing at which rate the smoothness of the loss landscape deteriorates as $T \to +\infty$.

\paragraph{Gradient degeneracy and sharpness.} As a consequence of previous considerations, the gradient $\partial_{\hat{\lambda}} \bar{\ell}_x(\hat{\lambda})$ inherits the regime-dependent behavior of the loss landscape,  provided in Proposition~\ref{prop:gradient-behavior}. 
Its asymptotic behavior as $T \to +\infty$ is summarized in Figure~\ref{fig:trajectory-loss-landscape}, which highlights the presence of regions of vanishing and exploding gradients over the $(\hat{\lambda}, \lambda)$ plane. Besides the stable quadrant $\hat{\lambda}, \lambda < 0$, where the gradient converges to a finite non-zero value, the degeneracies observed 
over the remaining quadrants develop exponentially fast in $T$. In particular, over the positive quadrant $\hat{\lambda}, \lambda > 0$, the behavior of the gradient is discontinuous over the line $\hat{\lambda} = \lambda$, with vanishing gradients for $\hat{\lambda} < \lambda$ and exploding gradients for $\hat{\lambda} > \lambda$. 
Hence, away from the minimizer, gradients become exponentially uninformative as $T \to +\infty$, pointing to possible convergence issues of gradient-based optimization. Furthermore, as provided in Proposition~\ref{prop:hessian-behavior}, the sharpness at the minimum diverges as $2T^2$ in the unstable regime.

\medbreak

The development of vanishing and exploding gradients at an exponential rate in $T$ is coherent with \citet[Theorems 1, 2]{mikhaeil22}, who relate the gradient behavior to the maximal Lyapunov exponent of the learned recursive map in the RNN setting, reducing here to $\hat{\lambda}$. The gradient degeneracy of the unstable regime in Proposition~\ref{prop:gradient-behavior} thus connects to the chaotic setting, with $\hat{\lambda}$ and $\lambda$ reading as the Lyapunov exponents of the learned and target dynamics in such a prototypical linear setting \citep{chicone24}.
In view of the above considerations, enlarging the temporal horizon $T$ clearly hinders gradient-based optimization. In particular, for non-contractive dynamics, the sharpening of the minimum implies that, for gradient descent to converge locally, the learning rate must scale as $T^{-2}$, while the radius of guaranteed local convergence shrinks as $T^{-1}$.

\subsubsection{Loss degeneracy in discrete-time} 
\label{sec:loss-degeneracy-discrete-time}

\begin{figure}[t]
    \centering
    \includegraphics[width=\textwidth]{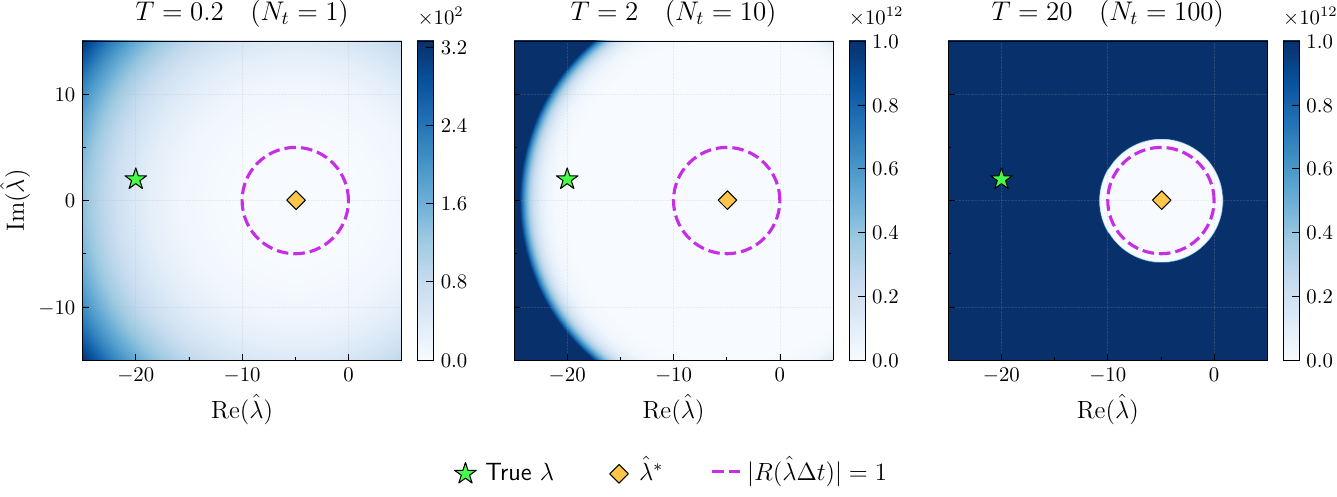}
    \caption{\textbf{Discrete-time trajectory loss degeneracy.} Representation of the loss collapse phenomenon in the infinite-horizon limit in the case of explicit Euler integration with step size $\dt = 0.2$, for $T=0.2, 2, 20$, resulting in $\Nt = 1, 10, 100$ time steps, respectively. The loss landscape is shown as a function of the real and imaginary parts of the learned parameter $\hat{\lambda}$, with the true parameter being $\lambda = -20 + 2\mathrm{i}$. The dashed region represents the stability region $\mathcal{A}_\dt$ of explicit Euler for the given step size, towards which the loss landscape collapses as $T\to +\infty$. 
    }
    \label{fig:discrete-time-loss-degeneracy}
\end{figure}

The challenges of trajectory-based learning are further compounded in the empirical setting, where simulation-based training requires the adoption of numerical integration to unroll the learned dynamics. Recent works have addressed the interplay between numerical integration and rollout-based training, highlighting the role of numerical stability in shaping the loss landscape and its implications for gradient-based optimization. In particular, \citet{fronk25} isolate the role of numerical integration in causing vanishing gradients for stiff neural ODEs, with numerical stability attenuating the parameter sensitivity associated with stiff modes. The central object is the stability function $R(z)$ associated with the numerical scheme. Indeed, for the prototypical case of linear dynamics (e.g Dahlquist test equation), the rollout loss \eqref{eq:rollout-loss} is minimized exactly at the learned parameters $\hat{\lambda}$ satisfying $R(\hat{\lambda} \dt) = e^{\lambda \dt}$, implying that $|R(\hat{\lambda} \dt)| < 1$ in the case of stable dynamics \citep[Theorem 2.1]{lu25}.
While these results have been derived in the context of neural ODEs, they are not specific to this setting, but rather arise in the broader context of system identification \citep[Section 7.4.5]{ramsay17}.

\paragraph{Loss collapse in the infinite-horizon limit.}
In this view, the following result characterizes the limiting behavior of the loss landscape \eqref{eq:rollout-loss} in the long rollout regime, namely as $\Nt \to +\infty$, showing that the loss landscape collapses onto such a stability region $\mathcal{A}_\dt = \{\hat{\lambda} \in \C : |R(\hat{\lambda} \dt)| < 1\}$, for a fixed time step $\dt$, and quantifying the rate at which such a collapse occurs.

\begin{emphbox}
\begin{restatable}[Loss degeneracy in discrete-time]{proposition}{lossdegeneracydiscretetime}
\label{prop:loss-degeneracy-discrete-time}
Fix $\Delta t > 0$, and $x_0 \in \C\setminus\{0\}$. Let the target dynamics be given by the linear system $\dot{x} = \lambda x$, $x(0) = x_0$, with $\lambda \in \C$, $\Re(\lambda) < 0$, observed at $t_k = k \dt$, $k=0,\dots,\Nt$. Let the parametric family $\dot{\hat{x}} = \hat{\lambda} \hat{x}$, $\hat{x}(0) = x_0$, with $\hat{\lambda} \in \C$, be simulated by a one-step Runge-Kutta method with constant step size $\dt$, and stability function $R$. Then, 
\begin{equation}
    \lim_{\Nt \to +\infty}\ell_{x,\Nt}(\hat{\lambda};x_0) = \begin{cases}
    0, & |R(\hat{\lambda} \dt)| < 1,\\
    |x_0|^2, & |R(\hat{\lambda} \dt)| = 1,\\
    +\infty, & |R(\hat{\lambda} \dt)| > 1,
    \end{cases}
    \label{eq:loss-degeneracy-discrete-time}
\end{equation}
and the collapse of the loss within the stability region $\mathcal{A}_\dt$ occurs at rate $\Nt^{-1}$, for every fixed $\hat\lambda$.
\end{restatable}
\end{emphbox}

The proof, given in Appendix~\ref{app:loss-degeneracy-discrete-time}, relies on the explicit expression of the rollout loss \eqref{eq:rollout-loss}, amounting to summing three geometric series whose convergence depends on the stability function $R$. As a consequence, the limiting objective \eqref{eq:loss-degeneracy-discrete-time} is blind to the target dynamics on the whole $\mathcal{A}_\dt$ region, admitting a continuum of global minima, despite the fact that the true parameter $\lambda$ may lie outside such a region, as illustrated in Figure~\ref{fig:discrete-time-loss-degeneracy} for the case of explicit Euler integration. This is a statement about the limit, and not about the finite-horizon minimizers, which remain confined to the finite set $\{\hat{\lambda} \in \C : R(\hat{\lambda} \dt) = e^{\lambda \dt}\}$, at every finite $\Nt$, as shown in \citet[Lemma 2.1]{lu25}. Such a collapse highlights that long rollouts do not displace the optimum, but rather flatten the loss landscape around it,
with loss values on $\mathcal{A}_\dt$ being pointwise $O(\Nt^{-1})$ and thus indistinguishable from the minimal value at that rate, representing a possible concern for gradient-based optimization.

\paragraph{Jacobian spectral constraint in the infinite-horizon limit.}
These arguments extend to the nonlinear case, governed by \eqref{eq:ode} and \eqref{eq:neural-ode} for the target and learned dynamics, respectively; we restrict to the autonomous setting, in which $f$ and $\hat{f}$ do not depend explicitly on time. Assuming that the target dynamics admits an asymptotically stable equilibrium, the following result shows that convergence of the series defining the infinite-horizon rollout loss translates into a spectral constraint on the Jacobian $\partial_x \hat{f}$ of the learned vector field.

\begin{emphbox}
\begin{restatable}[Jacobian spectral constraint]{proposition}{jacobianspectralconstraint}
\label{prop:jacobian-spectral-constraint}
Fix $\dt > 0$, and denote by
$\overline{\mathcal{A}}_\dt := \{\hat{\lambda} \in \C : |1 + \hat{\lambda} \dt| \leq 1\}$.
Let $f, \hat{f} \in C^1(\R^n;\R^n)$ with $\dt\lip{\hat{f}} < 1$, and
$\bar{x} \in \R^n$ be an asymptotically stable equilibrium of $f$. Let the learned dynamics
be rolled out by explicit Euler, with $\hat{x}_{k+1} = \hat{x}_k + \dt \hat{f}(\hat{x}_k)$.
Take $\delta > 0$, such that $B(\bar{x}, \delta)$ is contained in the basin of attraction of $\bar{x}$ for $f$, and the rollout loss satisfies
\begin{equation}
    \lim_{\Nt \to +\infty} \Nt \ell_{x,\Nt}(\hat{f}; x_0) < +\infty,
    \qquad \forall \, x_0 \in B(\bar{x}, \delta).
    \label{eq:bounded-rollout-loss}
\end{equation}
Then, $\hat{f}(\bar{x}) = 0$ and $\sigma\big(\partial_x \hat{f}(\bar{x})\big) \subseteq \overline{\mathcal{A}}_\dt$.
\end{restatable}
\end{emphbox}

The proof is given in Appendix~\ref{app:loss-degeneracy-discrete-time}. 
The two results, Proposition~\ref{prop:loss-degeneracy-discrete-time} and Proposition~\ref{prop:jacobian-spectral-constraint}, are in a sense converse to one another at an equilibrium, even stated in different settings.
By the former, membership in $\mathcal{A}_\dt$ yields a rollout loss decaying as $\Nt^{-1}$, that is, square-summable residuals; by the latter, square-summable residuals imply membership in $\overline{\mathcal{A}}_\dt$.
Hence, rollout training shapes the learned Jacobian $\partial_x \hat{f}(\bar{x})$, constraining its spectrum to lie within the closure of the stability region of the numerical scheme $\overline{\mathcal{A}}_\dt$, a set determined by the integrator and the step size alone, independent of the target dynamics. 
As a consequence, the stabilization induced by long rollouts training is merely a numerical effect, rather than evidence of fidelity to the true dynamics.

\medskip 

While carried out in a simplified setting, these arguments clarify that the challenges of trajectory-based learning are not restricted to the computational and memory costs of long rollouts, but concern the ill-suitedness of the trajectory objective to gradient-based optimization, 
further compounded by numerical simulation of the learned dynamics. The above considerations thus motivate alternative, simulation-free, learning strategies, which we address in the following sections.

\section{Velocity-based learning}
\label{sec:velocity-learning}
To address the challenges of trajectory-based learning, and motivated by the empirical effectiveness of recent velocity-based approaches \citep{zhou25, hou26}, we develop a theoretical framework for the latter, which will serve as a foundation for the proposed Velocity Matching (VM) objective.
We introduce the theoretical objective over velocities and provide a variational characterization of the velocity field of the observed flow as the unique minimizer of this quadratic objective. A series of stability estimates, in both continuous and discrete time, relate the trajectory-based objective to the velocity-based one. These estimates (i) characterize the gap between the two objectives, both theoretically and empirically; (ii) yield a simulation-free objective, evaluating velocity fields only along observed trajectories and thereby avoiding backpropagation through time; and (iii) reduce the approximation of the true dynamics to the approximation of its velocity field, yielding a universal approximation result for dynamical systems through neural ODEs whose velocity fields are parameterized by deep neural networks. This result, in turn, is shown to guarantee a continuous-time approximation property, which can be interpreted as a time super-resolution property of the learned system.

\subsection{Theoretical objective and variational characterization}
\label{sec:velocity-learning-1}

Trajectory-based learning minimizes the discrepancy between true and learned trajectories. The nonlinear dependence of the learned trajectories $\hat{x}(t,x_0)$ on the learned velocity field $\hat{f}$, through the solution of the initial value problem \eqref{eq:neural-ode}, makes the trajectory-based objective $\Lx$ non-convex in $\hat{f}$, yielding a challenging optimization problem. 
Since the object of interest is the velocity field itself, it is natural to cast learning directly as a regression problem over velocities, by regressing the $\hat{f}$ against the true field $f$ along observed trajectories, avoiding simulation of the learned system during training. We refer to the resulting population risk over velocities as the \textit{theoretical velocity objective},
\begin{equation}
    \Lf[\hat{f}] = \E_{x_0 \sim \rho_0}\lf(\hat{f};x_0), \quad \lf(\hat{f};x_0) = \frac{1}{T}\|\underbrace{f(\cdot, x(\cdot, x_0))}_{\partial_t x(\cdot, x_0)} - \hat{f}(\cdot, x(\cdot, x_0))\|^2_{L^2([0,T];\R^n)}.
    \label{eq:theoretical-velocity-objective}
\end{equation}
The objective $\Lf$ is a quadratic functional of the learned velocity field $\hat{f}$, whereas $\Lx$ is not.
Before addressing the relationship between the two objectives, we provide a variational characterization of the velocity field associated with the observed flow, showing that the true velocity field $f$ is the unique minimizer of $\Lf$, which provides a first justification for casting learning as a regression problem over velocities. The arguments are carried out in the autonomous setting, in which $f$ does not depend explicitly on time and the flow map $x(t,x_0)$ solves
\begin{equation}
    \partial_t x(t,x_0) = f(x(t,x_0)), \quad x(0,x_0) = x_0 \in \X_0,
    \label{eq:ode-autonomous}
\end{equation}
over $[0,T]$, for all $x_0 \in \X_0$. Recalling that $\rho_0$ denotes the distribution of initial conditions, compactly supported on $\X_0$, we introduce the occupation measure $\mu$ associated with the flow of \eqref{eq:ode-autonomous}, defined as
\begin{equation}
    \mu = \frac{1}{T}\int_0^T x(t,\cdot)_\#\rho_0 \, dt,
    \label{eq:occupation-measure}
\end{equation}
namely, the time-averaged push-forward of $\rho_0$ under the flow map, so that $\mu(A)$ represents the fraction of time spent in $A \subset \R^n$ by trajectories starting from $x_0 \sim \rho_0$. 
Note that $\mu$ is a probability measure, and is the normalized, $\rho_0$-averaged state marginal of the occupation measure of the process $(t,x(t,x_0))$ used in the context of optimal control
\citep{lasserre08}. As shown below, $\mu$ turns the time-averaged regression problem over velocities into a space-averaged one, more amenable to a variational characterization of the velocity field.

\begin{emphbox}    
\begin{restatable}[Velocity regression via occupation measure]{proposition}{velocityregressionoccupationmeasure}
\label{prop:velocity-regression-occupation-measure}
Let $f\in W^{1,\infty}_{\rm loc}(\R^n;\R^n)$, $x(t,x_0)$ be the flow of \eqref{eq:ode-autonomous}, and $\mu$ be the associated occupation measure \eqref{eq:occupation-measure}. Then, for every Borel measurable $\hat{f}:\R^n \to \R^n$, it holds that 
\begin{equation}
    \Lf[\hat{f}] = \E_{x\sim \mu} |f(x) - \hat{f}(x)|^2.
\end{equation}
\end{restatable}
\end{emphbox}

As a consequence of the above result, $\Lf$ can be identified with the population risk of a noise-free regression problem over the velocity field $f$, with respect to the occupation measure $\mu$. In particular, 
the true velocity field $f$ can be characterized as the unique minimizer of the theoretical velocity objective $\Lf$ over the $\mu$-weighted $L^2(\R^n;\R^n)$ space, denoted by $L^2(\mu;\R^n)$, as stated in the following result.

\begin{emphbox}    
\begin{restatable}[Variational characterization]{proposition}{variationalcharacterization}
\label{prop:variational-characterization}
Let $f\in W^{1,\infty}_{\rm loc}(\R^n;\R^n)$, and $\mu$ be the occupation measure associated with the flow of \eqref{eq:ode-autonomous}. Then, for $\hat{f}\in L^2(\mu;\R^n)$, it holds that
\begin{equation}
\Lf[\hat{f}] = 0 \iff \hat{f} = f \ \mu\text{-a.e.}
\end{equation}
In particular, $f$ is the unique minimizer of $\Lf$ over $L^2(\mu;\R^n)$.
\end{restatable}
\end{emphbox}

The proofs are given in Appendix~\ref{app:velocity-regression}.
While Proposition \ref{prop:velocity-regression-occupation-measure} only requires $\hat{f}$ to be Borel measurable, which suffices for the  subsequent variational characterization, well-posedness of the learned system \eqref{eq:neural-ode} requires $\hat{f}$ to be locally Lipschitz continuous.
Proposition \ref{prop:variational-characterization} determines $f$ only on the support of $\mu$, namely the region of the state space visited by trajectories starting from $\X_0$. This mirrors the correspondence between velocity fields and flow maps arising from elementary arguments. Indeed, by Cauchy--Lipschitz theory, a locally Lipschitz continuous velocity field $f$ generates a unique flow map $x(t, x_0)$, solution of \eqref{eq:ode-autonomous}; conversely, a flow map determines its velocity field pointwise through the relation $f(x(t,x_0)) = \partial_t x(t,x_0)$, but only at points visited by the flow.

To a first approximation, the above consideration thus suggests minimizing a Monte Carlo estimate of $\Lf$, which regresses $\hat{f}$ onto $f$ at points sampled from the occupation measure $\mu$. This would require access to pointwise evaluations of the true velocity field $f$ at sampled points, which are not available in practice, as only flow observations are accessible. 
In other words, $\Lf$ is a population risk whose empirical counterpart cannot be directly computed from data, motivating the introduction of suitable surrogate objectives involving velocity estimation from flow observations; these are the subject of Section~\ref{sec:velocity-matching}. Prior to that, we relate trajectory-based and velocity-based objectives.

\subsection{Stability estimates and objective control}
\label{sec:velocity-learning-2}

In this section we address the relationship between the trajectory-based objective $\Lx$ and the velocity-based one $\Lf$, by deriving stability estimates, in both continuous and discrete-time, that allow to control the former in terms of the latter, up to a stability constant quantifying the gap between the two.

\subsubsection{Continuous-time and discrete-time stability estimates}
\label{sec:velocity-learning-2-1}

In a first instance, we provide a series of estimates in a continuous-time setting which are closely related to well-posedness results in the classical ODE theory \citep[Theorem 2.8]{teschl12}. The main goal is to bound the error between true $x(t,x_0)$ and learned trajectories $\hat{x}(t,x_0)$, in terms of the error between true $f$ and learned velocity fields $\hat{f}$ along the true trajectory, with the aim of then using the latter as a learning objective. 
Evaluation of the velocity fields only along the true trajectory turns out to be the key feature enabling to shift from a simulation-based framework to a simulation-free one, as the absence of $\hat{x}(t,x_0)$ in the right-hand side of the estimates allows to evaluate the velocity error without simulating the learned system, hence avoiding the need for backpropagation through time (i.e. through the numerical solver steps). As a result, the price to pay is the presence of a stability constant depending on $\hat{f}$ through its Lipschitz constant, which is a classical feature of Gronwall estimates. However, this is not a limitation in practice, and is actually the gap for which proper control is required, and which the proposed \textit{velocity matching} (VM) objective aims to address.

The result is stated in a general setting, allowing for time-dependent velocity fields, under the assumption of local Lipschitz continuity in the state variable, uniformly in time, which is sufficient to ensure well-posedness of the target and learned systems. The estimates are provided in different norms, denoting by $L^p$-$L^q$ the estimate of the $L^p$-norm of the trajectory error in terms of the $L^q$-norm of the velocity error, for $p,q\in\{2,\infty\}$ which are the most relevant in our context.

\begin{emphbox}    
\begin{restatable}[Continuous-time stability estimates]{lemma}{continuoustimestabilityestimates}
    \label{lemma:continuous-time-stability-estimates}
    Let $\X_0\subset \R^n$ be compact and $T>0$ finite. 
    Let $f,\hat{f} \in C^0([0,T];W^{1,\infty}_{\rm loc}(\R^n; \R^n))$, and assume that for every $x_0\in\X_0$ the initial value
    problems \eqref{eq:ode}, \eqref{eq:neural-ode} admit classical solutions on $[0,T]$. Then there exists a compact set $K\subset\R^n$ containing $x(t,x_0)$ and $\hat{x}(t,x_0)$ for every $t\in[0,T]$ and every $x_0\in\X_0$, such that, setting $\Lambda := \lip{\hat{f}|_K}$, the $L^p$-$L^q$ estimate
    \begin{equation}
        \|x(\cdot, x_0) - \hat{x}(\cdot, x_0)\|_{L^p([0,T];\R^n)} \leq C_{p,q}(\Lambda) \|f(\cdot, x(\cdot, x_0)) - \hat{f}(\cdot, x(\cdot, x_0))\|_{L^q([0,T];\R^n)}, \quad \forall x_0 \in \X_0,
        \label{eq:stability-estimate-Lp-Lq}
    \end{equation}
    holds for the following $(p,q)$-pairs, with the corresponding constants:
    \begin{enumerate}
        \item $p=q=\infty$, with $C_{\infty,\infty}(\Lambda) = (e^{\Lambda T} - 1)/\Lambda$;
        \item $p=\infty$, $q=2$, with $C_{\infty,2}(\Lambda) = \sqrt{(e^{2\Lambda T} - 1)/(2\Lambda)}$;
        \item $p=q=2$, with $C_{2,2}(\Lambda) = \sqrt{e^{2\Lambda T} - 2\Lambda T - 1}/(2\Lambda)$.
    \end{enumerate}
\end{restatable}
\end{emphbox}

The three estimates of the above result, whose proof is provided in Appendix~\ref{app:stability-estimates}, serve different purposes. First, the $L^\infty$-$L^\infty$ estimate, which provides a uniform bound on the trajectory error in terms of the uniform error between true and learned velocity fields along the true trajectory, is central to establishing universal and continuous-time approximation results in the context of neural ODEs, as discussed in Section~\ref{sec:velocity-learning-3}. The $L^2$-$L^2$ estimate is relevant in practice, as it allows to relate the two theoretical objectives $\Lx$ and $\Lf$, as they are defined in terms of $L^2$-norms over time, hence providing a theoretical justification for the adoption of the $L^2$ velocity-based objective $\Lf$, discussed in Section~\ref{sec:velocity-learning-2-2}. Finally, the $L^\infty$-$L^2$ estimate shows that the $L^2$ velocity-based objective $\Lf$ is sufficient to control the trajectory error uniformly, a fact which is relevant in practice, as it will be used to establish control of the theoretical trajectory loss $\lx$ in terms of the proposed \textit{velocity matching} loss, as discussed in Section~\ref{sec:velocity-matching-3}.

\begin{figure}
    \centering
    \includegraphics[width=.95\textwidth]{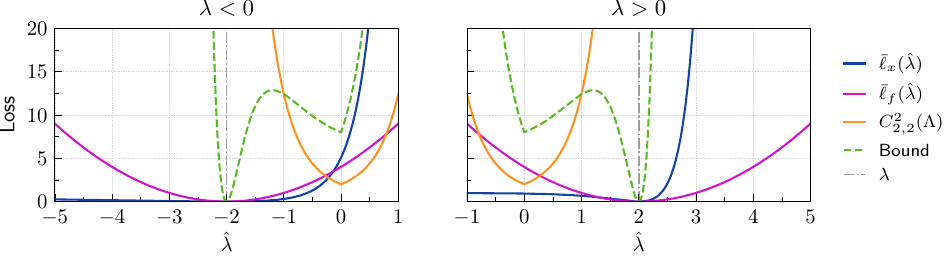}
    \caption{\textbf{Continuous-time stability estimate in a linear setting.} Representation of the different contributions appearing in the continuous-time stability estimates in a simplified linear setting, with $f(t,x) = \lambda x$, $\hat{f}(t,x) = \hat{\lambda} x$, and $\lambda, \hat{\lambda} \in \R$. The trajectory $\bar{\ell}_x$ and velocity $\bar{\ell}_f$ losses are normalized by the squared $L^2$-norm of the reference trajectory, and shown as a function of the learned parameter $\hat{\lambda}$, with true parameter $\lambda = -2$ (left) and $\lambda = 2$ (right). The bound (dashed green) decomposes into a product of two terms, namely $\bar{\ell}_f$ and the stability constant $C_{2,2}^2(\Lambda)$, with the former being convex in $\hat{\lambda}$, and sharing the same minimum with $\bar{\ell}_x$.}
    \label{fig:stability-estimates-bound-decomposition}
\end{figure}

The dependence of the stability constant $C_{p,q}(\Lambda)$ on the Lipschitz constant of the learned velocity field $\hat{f}$ points towards the need of controlling $\lip{\hat{f}}$ during training. The scalar linear setting provides further insights, as shown in Figure \ref{fig:stability-estimates-bound-decomposition}. The overall bound, being the product of $C_{p,q}(\Lambda)$ and the velocity error, is non-convex in $\hat{\lambda}$, whereas the velocity error alone is convex and attains its minimum at $\hat{\lambda} = \lambda$, as does the trajectory error. The fact that the stability constant is minimized at $\hat{\lambda} = 0$, highlights the fact that arbitrary regularization may pull the learned velocity field towards a trivial solution.

\medbreak

A discrete-time counterpart of the continuous-time stability estimates can be established as well, when the learned system is simulated with a one-step Runge-Kutta (RK) scheme. Here, the idea is to relate the trajectory error to the one-step residual $\varepsilon_{k+1}$, which measures the mismatch between true and learned dynamics over a single time step, when the current reference state $x(t_k)$ is fed as input to scheme, and is defined as
\begin{equation}
    \varepsilon_{k+1}(x_0) := x(t_{k+1},x_0) - x(t_k,x_0) - \dt \, F(t_k, x(t_k,x_0), \dt; \hat{f}), \quad k=0,\ldots, \Nt-1,
    \label{eq:tf-residual}
\end{equation}
with $F$ being the increment function associated with the RK scheme, in Henrici's notation \citep{quarteroni07, hairer08}. The proof of the following result is provided in Appendix~\ref{app:stability-estimates}.

\begin{emphbox}
\begin{restatable}[Discrete-time stability estimate]{lemma}{discretetimestabilityestimate}
\label{lemma:discrete-time-stability-estimate}

Let $\X_0\subset \R^n$ be compact, $T>0$ finite, $\Nt \in \N$, and $\dt = T/\Nt$. Let $f,\hat{f} \in C^0([0,T];W^{1,\infty}_{\rm loc}(\R^n; \R^n))$, and assume that for every $x_0\in\X_0$ the initial value problem \eqref{eq:ode} admits a classical solution on $[0,T]$. Let $\{\hat{x}_k(x_0)\}_{k=0}^{\Nt}$ be the numerical solution obtained from integrating $\hat{f}$ with a one-step explicit Runge-Kutta scheme with increment function $F$, and step size $\dt$ from the initial condition $x_0$.
Then there exists a compact set $K\subset\R^n$ containing $x(t,x_0)$ and $\hat{x}_k(x_0)$ for every $t\in[0,T]$, $k=0,\ldots,\Nt$, and $x_0\in\X_0$, such that 
    \begin{equation}
         \max_{0\leq k \leq \Nt} |x(t_k,x_0) - \hat{x}_k(x_0) |\leq C_{\dt}(\Lambda) \sum_{k=1}^{\Nt} | \varepsilon_{k}(x_0) |, \quad \forall\, x_0\in\X_0,
         \label{eq:discrete-stability-estimate}
    \end{equation}
with $C_{\dt}(\Lambda) = e^{T\Lambda}$ and $\Lambda := \sup_{t\in[0,T]}\lip{F(t,\cdot\,;\dt,\hat{f})|_{K}}$.
\end{restatable}
\end{emphbox}

Corollary~\ref{cor:empirical-objective-control} leverages the relation between the one-step residual $\varepsilon_k$ and the \textit{local truncation error} (LTE), used to characterize the consistency of numerical methods for ODEs, to explicitly relate the empirical rollout objective~\eqref{eq:rollout-loss} to the one-step teacher forcing objective~\eqref{eq:teacher-forcing-velocity-formulation}.

\begin{remark}[Lax equivalence perspective]
\label{rmk:lax-equivalence-perspective}    
The stability estimates provided in Lemma~\ref{lemma:continuous-time-stability-estimates} and Lemma~\ref{lemma:discrete-time-stability-estimate} can be interpreted as a Lax-Richtmyer, or, in an ODE context, Dahlquist-type equivalence result \citep{lax56,dahlquist56}.
In classical convergence theory for numerical methods for ODEs, stability estimates relate the convergence of the numerical solution to the notion of consistency, defined by means of the local truncation error, through a stability constant depending on the Lipschitz constant of the vector field, with the source of error being due to a numerical discretization. Here, the provided estimates mimic the same argument, with the difference that the source of error is due to an approximation $\hat{f}$ of the true vector field $f$, rather than a numerical discretization. The notion of consistency is here read in an infinitesimal way, as the error between true and learned velocity fields along the true trajectory, while convergence is read in terms of trajectory error over $[0,T]$.
The role of numerical schemes, which inevitably enter the picture when simulating the learned dynamics, is made explicit in Lemma~\ref{lemma:discrete-time-stability-estimate}, which more naturally fits into the classical Lax-Dahlquist framework.
\end{remark}

\subsubsection{Objective control}
\label{sec:velocity-learning-2-2}

Here a series of corollaries are provided, which are a direct consequence of the stability estimates established in the previous section, and relate the trajectory-based objective to the velocity-based one. 
While the results of this section hold at fixed $\hat{f}$, reading them as a justification for minimizing the velocity objective in place of the trajectory one, however, clearly requires the stability constant not to degrade as $\hat{f}$ varies. Therefore, by restricting the hypothesis class to velocity fields of uniformly bounded Lipschitz constant, the following results can be read as a justification for minimizing the velocity objective $\Lf$ in place of the trajectory one $\Lx$, given uniformly bounded stability constants $C_{p,q}(\Lambda)$. Such a restriction is natural in practice, and will be made explicit in the following sections.
First, the continuous-time $L^2$-$L^2$ stability estimate is leveraged to control the theoretical trajectory-based objective in terms of the theoretical velocity-based one, as stated in the following result.

\begin{emphbox}
\begin{restatable}[Theoretical objective control]{corollary}{theoreticalobjectivecontrol}
\label{cor:theoretical-objective-control}    
Let $\Lx$ and $\Lf$ denote the theoretical objectives over trajectories and velocities, respectively defined in \eqref{eq:theoretical-trajectory-objective} and \eqref{eq:theoretical-velocity-objective}. Then, under the same assumptions of Lemma~\ref{lemma:continuous-time-stability-estimates}, it holds that
\begin{equation}
    \Lx[\hat{f}] \leq C_{2,2}^2(\Lambda) \Lf[\hat{f}].
\end{equation}
\end{restatable}
\end{emphbox}

The proof is a direct application of the $L^2$-$L^2$ stability estimate in Lemma~\ref{lemma:continuous-time-stability-estimates}, and is provided in Appendix~\ref{app:objective-control}.
The result provides a theoretical justification for the adoption of a velocity-based objective $\Lf$ as a surrogate objective for the trajectory-based objective $\Lx$. The former is quadratic in $\hat{f}$ and simulation-free, since $\hat{f}$ is evaluated only along the reference trajectory $x(t,x_0)$, thus making it more amenable to optimization.

As a further consequence of the continuous-time stability estimates, a bound on the 2-Wasserstein distance between the push-forward of the distribution over initial conditions $\rho_0$ under the true and learned flow maps, namely $\rho_t = x(t,\cdot)_\#\rho_0$ and $\hat{\rho}_t = \hat{x}(t,\cdot)_\#\rho_0$, can be established as well. 
While weaker than the previous ones, the following result is relevant when the goal consists in recovering statistical properties of the system, rather than pointwise quantities, as it may be the case in the chaotic setting.

\begin{emphbox}
\begin{restatable}[2-Wasserstein stability estimate]{corollary}{wassersteinstabilityestimate}
\label{cor:wasserstein-stability-estimate}    
Let $\rho_0$ denote a distribution compactly supported on the set of admissible initial conditions $\X_0\subset \R^n$. Let $\rho_t$ and $\hat{\rho}_t$ be defined as the push-forward of $\rho_0$ under the flow maps $x(t,\cdot)$ and $\hat{x}(t,\cdot)$, respectively. Then, under the same assumptions of Lemma~\ref{lemma:continuous-time-stability-estimates}, the following $2$-Wasserstein stability estimate holds
\begin{equation}
    \mathcal{W}_2^2(\rho_t, \hat{\rho}_t) \leq T C_{\infty,2}^2(\Lambda) \Lf[\hat{f}], \quad \forall t\in[0,T].
    \label{eq:2-wasserstein-stability-estimate}
\end{equation}
\end{restatable}
\end{emphbox}

The proof is provided in Appendix~\ref{app:objective-control}. The estimate relates to \citet[Theorem 2.7]{li26}, who establishes universal approximation rates for transport equations via neural ODE approximators, namely \textit{neural transport equations}, in $1$-Wasserstein distance. The result builds upon the duality between the ODEs \eqref{eq:ode} and \eqref{eq:neural-ode} and the associated continuity equation \citep[Proposition 16.4]{ambrosio21}, with the densities $\rho_t$ and $\hat{\rho}_t$ being its weak solutions, while the flow maps $x(t,\cdot)$ and $\hat{x}(t,\cdot)$ are its characteristics. In this view, the result finds further application in the context of flow-based generative modeling \citep{gao24, ding26}, where the learned flow map transports a source distribution to match the target distribution at some final time $t=T$. In particular, the bound \eqref{eq:2-wasserstein-stability-estimate} connects to flow matching objectives \citep{lipman23, tong24}, where minimization of $\mathcal{W}_2(\rho_T, \hat{\rho}_T)$ is achieved by regressing the learned velocity field $\hat{f}$ to the true one $f$, through the minimization of a velocity-based objective $\Lf$.

\medbreak

Adopting the notation of Section~\ref{sec:learning-strategies}, the teacher forcing (TF) objective at time $t_k$ admits a natural numerical interpretation. Defining the local truncation error (LTE) as $\tau_{k+1} = \varepsilon_{k+1}/\dt$, the TF objective \eqref{eq:teacher-forcing-velocity-formulation} is, up to a rescaling factor $\dt^2$, the mean squared LTE.
By classical numerical analysis, the LTE measures the consistency of a scheme, and its decay rate in $\dt$ characterizes the order of the scheme itself. Here the residual, and hence the LTE, carries an additional contribution arising from the evaluation of $F$ at the learned velocity field $\hat{f}$, rather than the true one $f$, as the following decomposition makes explicit
\begin{equation}
    \tau_{k+1} = \left (\frac{x(t_{k+1}) - x(t_k)}{\dt}
    - F\big(t_k, x(t_k), \dt; f\big)\right )
    + \left (F\big(t_k, x(t_k), \dt; f\big)
    - F\big(t_k, x(t_k), \dt; \hat{f}\big) \right ).
    \label{eq:tf-residual-decomposition}
\end{equation}
The first term in \eqref{eq:tf-residual-decomposition} is the classical LTE associated with the numerical scheme applied to the true velocity field $f$. Independent of the learned velocity field $\hat{f}$, it can be made arbitrarily small by reducing the step size $\dt$, being of order $O(\dt^p)$ for a scheme of order $p\geq 1$. 
At a fixed $\dt$ this term is governed solely by the order of the scheme, and may introduce a bias dominating the second term, hence motivating the adoption of higher-order schemes during training \citep{krishnapriyan23, zhou25}. 
The second term accounts instead for the mismatch between the true and learned velocity fields along the reference trajectory $x(t_k)$, by means of the increment function $F$ of the adopted scheme. TF thus minimizes such a second term, which for the case of explicit Euler reads as $f(t_k, x(t_k))-\hat{f}(t_k, x(t_k))$, thereby yielding an empirical counterpart of the continuous-time velocity-based objective $\Lf$ in \eqref{eq:theoretical-velocity-objective}. 

The following result establishes a connection between the rollout empirical risk \eqref{eq:rollout-loss} and the TF empirical risk \eqref{eq:teacher-forcing-velocity-formulation}, by leveraging the discrete-time stability estimate in Lemma~\ref{lemma:discrete-time-stability-estimate}. 

\begin{emphbox}
\begin{restatable}[Empirical objective control]{corollary}{empiricalobjectivecontrol}
\label{cor:empirical-objective-control}    
Let $\L_{x,N,\Nt}[\hat{f}]$ denote the rollout empirical objective defined in \eqref{eq:rollout-loss}, and $\L_{x,N,\Nt,1}[\hat{f}]$ denote the empirical objective associated with the teacher forcing loss defined in \eqref{eq:teacher-forcing-velocity-formulation}. Then, under the same assumptions of Lemma~\ref{lemma:discrete-time-stability-estimate}, it holds that
\begin{equation}
    \L_{x,N,\Nt}[\hat{f}] \leq N_t^2 C^2_{\dt}(\Lambda) \, \L_{x,N,\Nt,1}[\hat{f}].
\end{equation}
\end{restatable}
\end{emphbox}

The proof is provided in Appendix~\ref{app:objective-control}. While the right-hand side of the above bound still reads as a trajectory-based quantity, the object at play is the velocity-based formulation of the TF loss \eqref{eq:teacher-forcing-velocity-formulation}, with the stability constant $C_{\dt}(\Lambda)$ being inherited from the discrete-time stability estimate of Lemma~\ref{lemma:discrete-time-stability-estimate}. 

\medskip

The results of this section thus close the chain of arguments bridging the gap between trajectory-based and velocity-based objectives. In continuous-time, the theoretical trajectory objective $\Lx$ is controlled by the quadratic theoretical velocity objective $\Lf$, quadratic in $\hat{f}$; in discrete-time, the multi-step rollout loss is controlled by the one-step TF loss, with suitable stability constants quantifying the gap in either case. Simulation is thereby removed from training, at the price of a constant depending on the Lipschitz constant of the learned velocity field $\hat{f}$, growing with the time horizon $T$. From an empirical perspective, what remains to be addressed is the estimation of velocities themselves from discrete observations, leading to the proposed \textit{velocity matching} (VM) objective, discussed in Section~\ref{sec:velocity-matching}.

\subsection{Universal and continuous-time approximation}
\label{sec:velocity-learning-3}

In this section we consider the case when the learned velocity field $\hat{f}$ is parametrized by a deep neural network, and show that, under suitable regularity assumptions on the true vector field $f$, the associated flow map $x(t,x_0)$ can be approximated with arbitrary accuracy by a flow map $\hat{x}(t,x_0)$ generated by a deep neural network $\hat{f}$ that approximates $f$ in $W^{1,\infty}$-norm. 
The result has twofold implications. First, it provides insights on the role of the Lipschitz constant of the learned vector field $\hat{f}$, which characterizes the stability constant in the provided estimates, hinting towards the adoption of a suitable empirical objective, which, in addition to minimizing the velocity error along observed trajectories, steers the Lipschitz constant of the learned vector field $\hat{f}$ towards the one of the target vector field $f$, rather than relying on arbitrary regularization strategies. Second, it enables to establish a continuous-time approximation result for neural ODEs, characterizing time super-resolution capabilities of the learned system at inference time, when simulating the learned dynamics with arbitrary time step size. 

\subsubsection{Universal approximation of flow maps}
Before establishing continuous-time approximation properties of the learned flow, we prove that the flow map of \eqref{eq:ode} is universally approximated by flow maps associated with neural ODEs \eqref{eq:neural-ode}, whose learned velocity field $\hat{f}$ is parametrized by a deep neural network.
The argument rests on the $L^\infty$-$L^\infty$ stability estimate of Lemma \ref{lemma:continuous-time-stability-estimates}, which bounds the trajectory error between true and learned flow maps in terms of the error between true and learned velocity fields along the reference trajectory, and thereby transfers the approximation problem from the space of flow maps to the space of velocity fields. The latter is then addressed by relying on universal approximation results for deep neural networks in Sobolev spaces \citep{guhring21}. Two aspects require care for the application of the stability estimate, and for the argument to go through. First, given the dependence of the estimate on the Lipschitz constant of the learned velocity field $\hat{f}$, through the stability constant $C_{\infty,\infty}(\Lambda)$, the approximation must be carried out in $W^{1,\infty}$-norm, rather than in $L^\infty$-norm, so that joint control over velocity error and Lipschitz constants can be established. Second, both flows must be confined to a common compact set, which for the learned flow is not given a priori and can be established by means of an escape argument, standing on the following two assumptions.

\begin{assumption}
\label{ass:vl-regularity-1}
Denoting by $K$ the closed $\delta$-neighborhood\footnote{We define the closed $\delta$-neighborhood of a set $\Omega\subset\R^n$ as the set $\{x \in \R^n : \inf_{y \in \Omega} |x - y| \leq \delta\}$.} of the convex hull of the reachable set $\X$ defined in \eqref{eq:reachable-set}, the target velocity field $f$ belongs to
\begin{equation}
        \mathcal{F} = \left \{g \in W^{2,\infty}_{\rm loc}(\R^{n+1}; \R^n) :  \|g\|_{W^{2,\infty}([0,T]\times K; \R^n)} \leq M\right \},
        \label{eq:target-velocity-field-class}
\end{equation}
for some $\delta>0$ and $M>0$.
\end{assumption}

\begin{assumption}
\label{ass:vl-regularity-2}
The activation function $\varrho: \R \to \R$ belongs to $W^{1,\infty}_{\rm loc}(\R)$, and is admissible for \citet[Proposition 4.8, Theorem 4.9]{guhring21}.
\end{assumption}

Denote by $\mathcal{N}_{\varrho}(\R^{n+1}; \R^n)$ the class of feedforward neural networks with input dimension $n+1$, output dimension $n$, and activation function $\varrho$. Assumption \ref{ass:vl-regularity-2} then guarantees that $\mathcal{N}_{\varrho}(\R^{n+1}; \R^n) \subset W^{1,\infty}_{\rm loc}(\R^{n+1}; \R^n)$, by the same argument as in \citet[Remark 2.12]{guhring20}, yielding local Lipschitz continuity of the learned velocity field $\hat{f}$, and hence local existence and uniqueness of the associated flow map $\hat{x}(t,x_0)$.
The admissibility conditions required by Assumption \ref{ass:vl-regularity-2} involve a technical condition on the differentiability of the activation function, requiring non-vanishing second derivative at some point, and that $\varrho$ admits an approximate partition of unity, in the sense of \citet[Definition 4.2]{guhring21}, whose decay defines in turn the approximation rate of the associated neural network class.  A wide class of activation functions, including exponential linear units (ELUs) and sigmoidal linear units (SiLU), satisfy these conditions.
We note that rectified linear units (ReLU), while in $W^{1,\infty}_{\rm loc}(\R)$, are not admissible for the results of \citet{guhring21}, and hence excluded from Assumption \ref{ass:vl-regularity-2}. These results can be eventually recovered by relying on \citet{yarotsky17}, as discussed in \citet[Remark 4.10]{guhring21}, at the cost of a depth bound which is no longer independent of the approximation accuracy. This is not a limitation in our context, as further regularity assumptions on the activation function $\varrho$ will be required in Section~\ref{sec:velocity-learning-3-2} to establish continuous-time approximation results, and will be made explicit therein.
We proceed to establish the result, whose proof is provided in Appendix~\ref{app:universal-continuous-time-approximation}.

\begin{emphbox}
\begin{restatable}[Universal approximation of flow maps]{theorem}{universalapproximationflow}
\label{thm:universal-approximation-flow}
Let $f$ satisfy Assumption \ref{ass:vl-regularity-1}, and $\varrho$ satisfy Assumption \ref{ass:vl-regularity-2}. Then, for every $\epsilon > 0$ there exists a deep neural network $\hat{f} \in \mathcal{N}_\varrho(\R^{n+1}; \R^n)$ such that the associated flow map $\hat{x}(t,x_0)$, solving \eqref{eq:neural-ode}, exists and remains in $K$ for all $t\in[0,T]$ and $x_0 \in \X_0$, and satisfies
\begin{equation}
    \|x - \hat{x}\|_{L^\infty([0,T]\times \X_0; \R^n)} \leq \epsilon.
\end{equation}
Moreover, for $\varrho$ being a GELU activation function, $\hat{f}$ can be chosen with a constant number of layers, independent of $\epsilon$, and $O(n\epsilon^{-(n+1)})$ nonzero weights.
\end{restatable}
\end{emphbox}

The result establishes that the flow map of \eqref{eq:ode} can be uniformly approximated, over $[0,T]$ and the set of admissible initial conditions $\X_0$, with arbitrary accuracy by a flow map generated by a deep neural network $\hat{f}$, which approximates $f$ in $W^{1,\infty}$-norm.
Although the theorem allows for a wide class of activation functions satisfying Assumption \ref{ass:vl-regularity-2}, the final statement provides a more explicit characterization for the case of GELU activation \citep{hendrycks23}, which we adopt in our numerical experiments. The corresponding approximation rate follows from showing that GELU belongs to the class of exponential partitions of unity (PU) \citep[Definition 4.2]{guhring21}, and by applying \citet[Theorem 4.9]{guhring21}. The argument is made explicit in Appendix~\ref{app:universal-continuous-time-approximation}.

\subsubsection{Continuous-time approximation}
\label{sec:velocity-learning-3-2}

A central question of practical relevance is whether, at inference time, the learned neural ODE \eqref{eq:neural-ode} can be simulated over arbitrary time discretizations, including ones finer than that employed during training, while retaining a prescribed accuracy in terms of trajectory error. Although neural ODEs rest on a continuous-time formulation, their training relies on discrete-time observations. Consequently, the learned system may overfit the training discretization, potentially compromising its ability to capture the underlying continuous-time dynamics and generalize to finer temporal discretizations.
This property, referred to as \textit{continuous-time approximation} and interpreted as a form of time super-resolution \citetext{\citealp[Definition~1]{krishnapriyan23}; \citealp[Definition~1]{farenga25}}, can be understood through the decomposition of the trajectory error, over a discrete grid $\{t_k\}_{k=0}^{\Nt}$, into two contributions, 
\begin{equation}
\sup_{0\leq k\leq \Nt} |x(t_k, x_0) - \hat{x}_k(x_0)| \leq \|x(\cdot, x_0) - \hat{x}(\cdot, x_0)\|_{L^\infty([0,T]; \R^n)} + C(\hat{f},T)\dt^p \label{eq:error-decomposition},
\end{equation}
namely the $L^\infty$-error between the true and learned continuous-time trajectories, and the numerical error arising from the integration of the learned system. This decomposition shows that the learned dynamics' continuous-time approximation properties are governed by two sources of error. The former is independent of the numerical scheme, and is controlled by the universal approximation property of flow maps established in Theorem \ref{thm:universal-approximation-flow}; whereas the latter is governed by the convergence order of the numerical integration scheme employed at inference time, yielding an $O(\dt^p)$ error for a one-step RK scheme of order $p\geq 1$, provided that the learned velocity field $\hat{f}$ is sufficiently regular for such an order to be attained \citep[Section II.3]{hairer08}.
The following result formalizes such an argument.

\begin{emphbox}
\begin{restatable}[Continuous-time approximation]{theorem}{continuoustimeapproximation}
\label{thm:continuous-time-approximation}
Let $f$ satisfy Assumption \ref{ass:vl-regularity-1}, and $\varrho$ satisfy Assumption \ref{ass:vl-regularity-2}, with in addition $\varrho \in C^p(\R)$ for some $p\geq 1$. Then, for every $\epsilon > 0$ there exists a deep neural network $\hat{f} \in \mathcal{N}_\varrho(\R^{n+1}; \R^n)$ and a step size $\dt^* = \dt^*(\epsilon, T, \hat{f}) > 0$ such that the numerical solution $\hat{x}_k(x_0)$ of \eqref{eq:neural-ode}, obtained with a one-step RK scheme of order $p$ over a uniform discretization $\{t_k\}_{k=0}^{\Nt}$ of $[0,T]$ with step size $\dt=T/\Nt$, satisfies
\begin{equation}
\sup_{x_0\in\X_0}\sup_{0\leq k\leq \Nt} |x(t_k, x_0) - \hat{x}_k(x_0)| \leq \epsilon, \qquad \forall \, \dt \leq \dt^*.
\end{equation}  
\end{restatable}
\end{emphbox}

The proof is provided in Appendix~\ref{app:universal-continuous-time-approximation}. The result establishes that the learned neural ODE can be simulated at inference time with arbitrary time step size $\dt \leq \dt^*$, while retaining a prescribed accuracy in terms of trajectory error. The associated threshold $\dt^*$ is determined by the desired accuracy $\epsilon$, through the decomposition in \eqref{eq:error-decomposition}, and by the learned velocity field $\hat{f}$ in two ways. First, the threshold depends on the constant arising from the $O(\dt^p)$ term in \eqref{eq:error-decomposition}, associated with the RK scheme and commonly depending on the Lipschitz constant of $\hat{f}$ and the time horizon $T$. Second, the threshold depends on the order of convergence $p$ of the numerical scheme, which is guaranteed to be attained only if $\hat{f}$ is sufficiently regular, hence the requirement on the activation function $\varrho$ to be $C^p(\R)$.
\section{Velocity matching}
\label{sec:velocity-matching}

The stability estimates provided in Section~\ref{sec:velocity-learning} justify replacing the trajectory-based objective with a velocity-based objective, a quadratic functional whose unique minimizer is the true velocity field $f$. While avoiding simulation of the learned dynamics, this objective is, however, not directly accessible, as it requires knowledge of the true velocity field along the reference trajectories, not available in practice. Any empirical surrogate of the theoretical velocity objective must therefore rely on velocity estimates as supervision signal from discrete observations of the system state. In this view, this is a price to be paid for removing simulation from training. 
Here we introduce a principled velocity-based objective, which we call \textit{velocity matching}. Rather than supervising the learned velocity field $\hat{f}$ at the discrete observation points, as in commonly adopted teacher forcing (TF) approaches, VM evaluates $\hat{f}$ along the piecewise linear interpolant of the observed trajectories, at times sampled uniformly within each interval, and supervises it with the velocity obtained by differentiating the interpolant.

In the following, in Section~\ref{sec:velocity-matching-1} we introduce the VM objective and describe its algorithmic implementation, framing it as an instance of an interpolant-based surrogate for $\Lf$ and relating it to prior work. Section~\ref{sec:velocity-matching-2} provides a bias-variance decomposition of the VM loss, and relates the VM bias term to the residual of an implicit Runge-Kutta method, highlighting the implicit nature of the resulting update. Section~\ref{sec:velocity-matching-3} interprets VM loss as a stochastic quadrature rule, establishing $O(\dt^2)$ order of approximation of the loss associated with the theoretical velocity objective $\Lf$. Section~\ref{sec:velocity-matching-4} casts VM as a regularized velocity regression whose penalty implicitly matches the action of the Jacobian of the learned velocity field to that of the true velocity field.

\subsection{Empirical objective and algorithmic aspects}
\label{sec:velocity-matching-1}

We seek an empirical counterpart of the theoretical velocity objective $\Lf$, defined in \eqref{eq:theoretical-velocity-objective}, overcoming the limitation of requiring knowledge of the true velocity field along the reference trajectories. A natural approach, dating back to the early developments of system identification \citep{bellman71, varah82, ramsay07}, consists in relying on \textit{two-stage} estimation, where a prior interpolation step produces a continuous approximation $\tilde{x}(t,x_0)$ of the observed trajectories, from which velocities can be estimated by differentiating such an interpolant.
Then, the learned vector field $\hat{f}$ is matched with the estimated velocities along the interpolated trajectories, yielding a surrogate objective for $\Lf$, which we denote by $\tLf$, reading as
\begin{equation}
    \tLf[\hat{f}] = \frac{1}{N}\sum_{i=1}^N\tlf(\hat{f};x_0^{(i)}), \quad \tlf(\hat{f};x_0) = \frac{1}{T}\| \partial_t\tilde{x}(\cdot, x_0) - \hat{f}(\cdot, \tilde{x}(\cdot, x_0))\|^2_{L^2([0,T];\R^n)}.
    \label{eq:surrogate-velocity-objective}
\end{equation}
Although built from the finite set of $\Nt+1$ observations, the surrogate objective is expressed in continuous-time form through the $L^2$-norm over $[0,T]$, since the interpolant $\tilde{x}(\cdot,x_0)$ provides a, possibly almost everywhere, continuous approximation of the observed trajectories and velocities, whose regularity is set by the chosen interpolation strategy. While prior approaches have favored globally regular interpolants, as cubic \citep{bellman71} or quintic splines \citep{hou26}, we adopt piecewise linear interpolation for the following reasons. First, the $L^2$ objective is insensitive to the regularity at the knots $\{t_k\}_{k=0}^{\Nt}$, a set of null measure, so that almost everywhere differentiability of the interpolant suffices to ensure well-posedness of the objective. Second, piecewise linear interpolation carries the computational cost of a first-order finite difference, as opposed to the cost of fitting higher-order splines. Although the resulting supervision signal is now piecewise constant, with discontinuities at the knots, the spectral bias \citep{rahaman19, xu19}  of neural network parameterizations of $\hat{f}$ allows smoothness to be recovered at the level of the approximator rather than on the target.

This choice yields the \textbf{velocity matching} (VM) objective, denoted by $\LVM$ with per-sample loss $\lVM$, as the specific instance of the empirical surrogate objective \eqref{eq:surrogate-velocity-objective} obtained from piecewise linear interpolation of the observed trajectories, with
\begin{equation}
\tilde{x}(t,x_0) :=  x(t_{k}, x_0) + (t-t_k)\, v_k(x_0), \qquad
\partial_t \tilde{x}(t,x_0) = v_k(x_0):=\dfrac{x(t_{k+1}, x_0) - x(t_k, x_0)}{t_{k+1}-t_k},
\label{eq:piecewise-linear-interpolant}
\end{equation}
for $t\in[t_k, t_{k+1})$ and $k=0,\dots, \Nt-1$. 
Under uniform temporal discretizations, the $L^2$-norm in \eqref{eq:surrogate-velocity-objective} decomposes over subintervals, and sampling the integration variable uniformly within each of them recasts the loss as a stochastic perturbation of the time and space arguments in $\hat{f}$. Accordingly, we define the per-sample VM loss as
\begin{eqbox}
\begin{equation}
    \lVM(\hat{f};x_0) = \frac{1}{\Nt}\sum_{k=0}^{\Nt-1}\E_{\tau\sim \mathcal{U}[0, \dt]}\big|v_k(x_0) - \hat{f}(t_k +\tau, x(t_k,x_0) + \tau v_k(x_0))\big|^2,
    \label{eq:velocity-matching-objective-loss}
\end{equation}
\end{eqbox}
which we adopt in practice. Its consistency with the interpolant-based objective is stated below and proved in Appendix \ref{app:velocity-matching-1}.

\begin{emphbox}
\begin{restatable}{proposition}{vmlossformulation}
    \label{prop:vm-loss-formulation}
    The loss \eqref{eq:velocity-matching-objective-loss} coincides with  the surrogate objective \eqref{eq:surrogate-velocity-objective} under piecewise linear interpolation \eqref{eq:piecewise-linear-interpolant}, on a uniform temporal discretization of step size $\dt$.
\end{restatable}
\end{emphbox}

Formulation \eqref{eq:velocity-matching-objective-loss} unfolds the structure of the VM objective in terms of a stochastic perturbation along the interpolant. Analytically, sampling the offset $\tau$ to evaluate the learned vector field $\hat{f}$ along the continuum of each subinterval, rather than at its left endpoint alone as in TF, yields: (i) a higher order of approximation of the theoretical velocity objective, and (ii) an implicit regularization effect on the Jacobian of the learned vector field, as shown in Section~\ref{sec:velocity-matching-3} and Section~\ref{sec:velocity-matching-4}, respectively.
Computationally, it reduces to a sampling procedure mirroring that of flow matching approaches \citep{lipman23, hou26}, as discussed in Remark \ref{rem:connection-flow-matching}. We detail the batched implementation in Algorithm \ref{alg:velocity-matching}.
In particular, the triple $(i_j, k_j, \tau_j)$ denotes the indices of the $j$-th sample in the batch, of size $B$, with $i_j$ and $k_j$ denoting the trajectory and temporal indices, respectively, and $\tau_j$ the offset within the selected subinterval. The velocity target $v_j$ is computed as the first-order finite difference approximation of the velocity along the selected trajectory and temporal index, while the collocation point $(\tilde{t}_j, \tilde{x}_j)$ is obtained by advancing along the chord connecting the two consecutive observations, with offset $\tau_j$. The learned vector field $\hat{f}$ is then evaluated at the collocation point and matched with the velocity target $v_j$. Thus, as each step evaluates the learned vector field at a single collocation point per sample, and constructs the velocity target as a first-order finite difference, the computational cost of VM is comparable to that of an explicit Euler TF update. 
Since the offset $\tau_j$ is sampled independently across the batch, the batch size $B$ controls the variance of the resulting gradient estimate. 

\begin{algorithm}[!th]
\DontPrintSemicolon
\SetKwInOut{KwReturn}{Return}
\caption{Velocity Matching - Batch Training}
\label{alg:velocity-matching}

\KwIn{Trajectories $\{x(t_k;x_0^{(i)})\}_{k=0}^{\Nt}$ with initial conditions $\{x_0^{(i)}\}_{i=1}^{N}$, temporal grid $\{t_k\}_{k=0}^{\Nt}$ of step size $\dt$, learnable velocity field $\hat{f}_\theta : \R^{n+1} \to \R^n$ of parameters $\theta$, batch size $B$, number of iterations $N_\mathrm{iter}$, learning rate $\eta$}
\KwOut{Updated parameters $\theta$}

\BlankLine

\begin{spacing}{1.3}

Initialize $\theta$\;

\For{$m = 1, \ldots, N_\mathrm{iter}$}{
  \vspace{.25em}
  Draw $B$ samples $(i_j, k_j, \tau_j) \sim \mathcal{U}\{1, N\} \times \mathcal{U}\{0, \Nt-1\} \times \mathcal{U}[0, \dt]$, for $j = 1, \ldots, B$\;
  
  Construct velocity targets $v_j = \dfrac{x(t_{k_j+1};x_0^{(i_j)}) - x(t_{k_j};x_0^{(i_j)})}{\dt}$, for $j = 1, \ldots, B$\;

  Construct temporal collocation points $\tilde{t}_j = t_{k_j} + \tau_j$, for $j = 1, \ldots, B$\;
  \vspace{.25em}
  Construct spatial collocation points $\tilde{x}_j = x(t_{k_j};x_0^{(i_j)}) + \tau_j v_j$, for $j = 1, \ldots, B$\;

  Compute loss $\mathcal{L}^{\mathrm{VM}}_{f,\text{batch}}(\theta) = \dfrac{1}{B} \sum_{j=1}^{B} \bigl| v_j - \hat{f}_{\theta}(\tilde{t}_j, \tilde{x}_j) \bigr|^2$\;

  Gradient step $\theta \leftarrow \theta - \eta \nabla_\theta \mathcal{L}^{\mathrm{VM}}_{f,\text{batch}}(\theta)$\;
}

\KwReturn{$\theta$}

\end{spacing}
\end{algorithm}

\begin{remark}[Connection to flow matching]
\label{rem:connection-flow-matching}
The VM construction is formally a \textit{flow matching} (FM) objective~\citep{lipman23}, where a learned velocity field is regressed onto the time derivative of an interpolant joining two endpoint samples, with the velocity field being evaluated along the interpolant itself. The connection with operator learning, in a continuous-time setting, was made explicit by \citet{hou26}, within the stochastic interpolant framework~\citep{albergo23, albergo25}.
However, VM differs from FM in the coupling between endpoints and, as a result, in the object being learned. Denote by $\rho(t)$ the law of $x(t, x_0)$ with $x_0 \sim \rho_0$, and by $\nu_k$ the joint law of $(x(t_k, x_0), x(t_{k+1}, x_0))$. Then, $\nu_k = (\operatorname{Id}, \phi_\dt)_\# \rho(t_k)$, where $\phi_\dt$ denotes the flow map of~\eqref{eq:ode} over a step $\dt$. Hence, the conditional law of $x(t_{k+1}, x_0)$ given $x(t_k, x_0)$ is the Dirac measure at $\phi_\dt(x(t_k, x_0))$, and the coupling is \textit{deterministic}.\footnote{Under the simplifying assumption of noise-free observations.} In FM, by contrast, the coupling is typically independent or chosen for computational convenience, and the associated velocity field transporting $\rho(t_k)$ to $\rho(t_{k+1})$ does not necessarily coincide with the physical dynamics.
In the present setting, the assumption of observing trajectories already yields a coupling between adjacent time steps, so that in VM each interpolant segment is anchored to an observed trajectory, and a single velocity field $\hat{f}$ is shared across all intervals, effectively approximating the underlying $f$. The choice of coupling, and of the resulting probability path, becomes instead essential when only the marginals $\{\rho(t_k)\}_{k=0}^{\Nt}$ are observed, as in stochastic or population dynamics settings~\citep{neklyudov23, lavenant24, berman24, blickhan25}, where trajectories need to be inferred.
\end{remark}

\subsection{Objective decomposition and implicit integration}
\label{sec:velocity-matching-2}

Here we address the connection between VM and implicit Runge-Kutta (IRK) schemes, and discuss its implications. We first isolate the two contributions governing the VM objective, through a bias-variance decomposition of the loss \eqref{eq:velocity-matching-objective-loss}, 
induced by its stochastic formulation with respect to the random offset $\tau$. 
To establish the subsequent results, we impose further regularity requirements on both the target and learnable velocity fields, requiring Lipschitz continuity in time, in addition to the Lipschitz continuity in space already assumed in Section~\ref{sec:velocity-learning}. 

\begin{assumption}
\label{ass:vm-regularity-1}
Denoting by $K$ the closed $\delta$-neighborhood of the convex hull of the reachable set $\X$ defined in \eqref{eq:reachable-set}, both target and learnable velocity fields, $f$ and $\hat{f}$, belong to the function class
\begin{equation}
        \mathcal{F} = \left \{g \in W^{1,\infty}_{\rm loc}(\R^{n+1}; \R^n) :  \|g\|_{W^{1,\infty}([0,T]\times K; \R^n)} \leq M\right \},
        \label{eq:hypothesis-class}
\end{equation}
for some $\delta>0$ and $M>0$.
\end{assumption}

This assumption ensures that the stability constants introduced in Section~\ref{sec:velocity-learning}, as well as interpolation error bounds, can be controlled uniformly.
The following proposition establishes the bias-variance decomposition of the VM loss, which is proved in Appendix \ref{app:velocity-matching-2}. 

\begin{emphbox}
\begin{restatable}[Bias-variance decomposition]{proposition}{vmlossdecomposition}
    \label{prop:vm-loss-decomposition}
    Under Assumption \ref{ass:vm-regularity-1}, denoting by $\hat{f}_k(\tau;x_0) := \hat{f}(t_k +\tau, x(t_k,x_0) + \tau v_k(x_0))$, the per-sample VM loss \eqref{eq:velocity-matching-objective-loss} admits the decomposition
    \begin{equation}
        \lVM(\hat{f};x_0) = \frac{1}{\Nt} \sum_{k=0}^{\Nt-1}\left ( \big|v_k(x_0) - \E_\tau [\hat{f}_k(\tau;x_0)] \big|^2 + \var_\tau [\hat{f}_k(\tau;x_0)] \right), \quad \tau \sim \mathcal{U}[0, \dt],
    \label{eq:vm-loss-decomposition}
    \end{equation}
    for every $x_0 \in \X_0$.
\end{restatable}
\end{emphbox}

The decomposition \eqref{eq:vm-loss-decomposition} holds with equality, with the two terms carrying distinct roles. The variance term accounts for the deviation of $\hat{f}$ along the interpolant, and is the contribution of the stochastic offset $\tau$ to the loss, unavailable through any deterministic collocation rule; its regularizing effect on the Jacobian of the learned velocity field is discussed in Section~\ref{sec:velocity-matching-4}. The bias term reads as a continuous collocation condition averaged over the subinterval $[t_k, t_{k+1}]$, namely
\begin{equation}
    \big |v_k(x_0) - \E_\tau [\hat{f}_k(\tau;x_0)] \big | =  \left |\frac{1}{\dt}\int_{t_k}^{t_{k+1}} \left ( f(t,x(t,x_0)) - \hat{f}(t,\tilde{x}(t,x_0)) \right ) dt \right |,
    \label{eq:vm-bias-term}
\end{equation}
having used the identity $v_k(x_0) = \dt^{-1}\int_{t_k}^{t_{k+1}} \partial_t x(t,x_0) dt$, by the fundamental theorem of calculus.
Enforcing the ODE residual at a single node $t_k + \gamma \dt$, with $\gamma \in [0,1]$, on a polynomial of degree 1, is equivalent to a one-stage method with tableau $a_{11} = c_1 = \gamma$ and $b_1 = 1$, implicit for $\gamma>0$, by the equivalence between collocation methods and IRK schemes \citep[Theorem 7.7]{hairer08}. Given such a correspondence, the VM bias term can be interpreted as a TF update, where the single-step update is performed via an IRK method, with the collocation point $\gamma$ sampled uniformly within the subinterval.
Proposition \ref{prop:vm-csrk} establishes such an equivalence, where the VM bias term is identified with the residual of a continuous-stage Runge-Kutta (csRK) method, which generalizes IRK methods to a continuum of stages. The resulting csRK scheme associated with VM bias recovers the \textit{average vector field} method \citep{quispel08, hairer10}, of order 2, A-stable and, in the case of Hamiltonian systems, energy preserving \citep{celledoni09}. 

\begin{emphbox}
    \begin{restatable}[VM as implicit TF]{proposition}{vmcsrk}
        \label{prop:vm-csrk}
        Under Assumption \ref{ass:vm-regularity-1}, the bias term in \eqref{eq:vm-loss-decomposition} reads as the residual obtained by inserting the observed pair $(x(t_k,x_0), x(t_{k+1},x_0))$ into the update of a csRK method with continuous-stage tableau
        \begin{equation}
            a(\gamma, \sigma) = \gamma, \quad b(\gamma) = 1, \quad c(\gamma) = \gamma, \qquad \gamma, \sigma \in [0,1].
            \label{eq:vm-csrk-tableau}
        \end{equation}
        Moreover, if $\dt M < 2$, the corresponding TF update, starting from $x(t_k,x_0)$, admits a unique solution $\hat{x}_{k+1}$, and its residual $\varepsilon_{k+1}^{\rm TF}(x_0) = x(t_{k+1},x_0) - \hat{x}_{k+1}$ satisfies 
        \begin{equation}
            \frac{\dt |v_k(x_0) - \E_\tau [\hat{f}_k(\tau;x_0)]|}{1+\frac{1}{2}\dt M} \leq |\varepsilon_{k+1}^{\rm TF}(x_0)| \leq \frac{\dt |v_k(x_0) - \E_\tau [\hat{f}_k(\tau;x_0)]|}{1-\frac{1}{2}\dt M},
            \label{eq:vm-csrk-residual-bound}
        \end{equation}
        for $k=0,\dots,N_t-1$.
    \end{restatable}
\end{emphbox}
The proof, along with details on csRK schemes, is provided in Appendix \ref{app:velocity-matching-2}. Denoting by $F$ the increment function associated with the csRK method, performing a TF update with an implicit scheme requires solving $\hat{x}_{k+1} = x(t_k, x_0) + \dt F(t_k, x(t_k,x_0), \hat{x}_{k+1}, \dt; \hat{f})$ for the predicted state $\hat{x}_{k+1}$, so that the residual $\varepsilon_{k+1}^{\rm TF}(x_0)$ can be formed. Thus, although TF is simulation-free, training would still require differentiating through the nonlinear solver, at a computational cost that scales with the number of iterations required for convergence. The VM bias, instead, avoids the implicit solve by directly evaluating the residual at the observed pair $(x(t_k,x_0), x(t_{k+1},x_0))$. In this sense, the VM objective carries the implicit method inside the optimization process, rather than in the forward pass, with the nonlinear system that a rollout or TF objective would require to solve at each step being replaced by a residual that the learned velocity field has to satisfy along the interpolant. 
Such a correspondence is made quantitative by the bounds \eqref{eq:vm-csrk-residual-bound}, which relate the two residuals up to multiplicative factors, showing that minimizing the VM bias term is equivalent to minimizing the residual of a TF update coupled with an IRK method.
In terms of stability properties, when considering the Dahlquist test equation $\dot{x} = \lambda x$, with $\lambda \in \C$, as in Section~\ref{sec:trajectory-learning}, the stability function associated with the \textit{average vector field} method reads as $R(z) = \frac{1+z/2}{1-z/2}$, with $z = \lambda \dt$, so that $|R(z)| \leq 1$ for $\Re(z) \leq 0$, yielding an A-stable method, namely a method whose stability region covers the entire left half of the complex plane. In particular, $R$ coincides with the stability function of the implicit midpoint method, which is a second-order A-stable method. This property, inherited by the VM bias term, agrees with the findings of \citet{lu25}, where the implicit midpoint rule is shown to be a principled choice for learning conservative or dissipative dynamics, preserving structural properties of the underlying system.

\subsection{Discretization bias, order of convergence and consistency}
\label{sec:velocity-matching-3}

The VM objective is an empirical surrogate of the theoretical velocity objective \eqref{eq:theoretical-velocity-objective}, and is therefore biased. We identify two sources of bias, both stemming from the discrete nature of observations over time.
First, the velocity target is estimated by a first-order finite difference $v_k(x_0)$, being the derivative of the piecewise linear interpolant. Second, the $L^2$-norm in time is approximated by uniformly sampling the offset $\tau$, and by evaluating the learned velocity field across the continuum of the interpolant, framing VM as a stochastic quadrature rule. 
The first source of bias is common to other empirical velocity-based objectives, as, for instance, in TF coupled with a first-order explicit Euler update \eqref{eq:teacher-forcing-velocity-formulation}. In this view, the distinguishing feature of VM lies in its stochastic collocation strategy, which, as shown below, yields an improved order of convergence in $\dt$ in the approximation of the theoretical velocity loss $\lf$, as compared to what TF attains.

\begin{emphbox}
\begin{restatable}[Order of convergence, VM loss]{theorem}{vmorder}
\label{thm:vm-order}
Under Assumption \ref{ass:vm-regularity-1}, for every $x_0 \in \X_0$, it holds that
    \begin{equation}
        |\lVM(\hat{f}; x_0) - \ell_f(\hat{f}; x_0) | \leq C^{\rm VM}_1 \dt^2 + C^{\rm VM}_2\dt^4,
        \label{eq:vm-order}
    \end{equation}
with $C^{\rm VM}_1 = 4M^3(1+M) + \frac{1}{3}M^2(1+M)^2$, and $C^{\rm VM}_2 = M^4(1+M)^2$.
\end{restatable}
\end{emphbox}    

The proof is provided in Appendix \ref{app:velocity-matching-3}, alongside a similar result for TF, which is shown to be of order $O(\dt)$ rather than $O(\dt^2)$, thus highlighting the improved order of convergence of VM. The proof relies on the decomposition \eqref{eq:vm-loss-decomposition}, whose terms are compared with the corresponding terms in the decomposition of the theoretical velocity loss $\lf$, and separately bounded.
The bias terms differ through the velocity target, and through the argument at which the learned velocity field is evaluated, namely the interpolant in the case of $\lVM$. Chord estimates, provided in Lemma~\ref{lemma:chord-estimates}, bound this difference by $O(\dt^2)$. The variance terms are separately $O(\dt^2)$, both integrands being Lipschitz in $\tau$, with constant controlled by $M$.
The estimate is a statement about the objective at fixed $\hat{f}$, and is uniform over $\mathcal{F}$, and it therefore transfers to the minimizers of the two objectives. In particular, combined with the previously established stability estimates, the following corollary closes the chain of estimates begun in Section~\ref{sec:velocity-learning}, bounding the trajectory loss $\lx$ in terms of the VM loss $\lVM$, and accounting for the discretization bias in $\dt$.

\begin{emphbox}
    \begin{restatable}[Loss control]{corollary}{vmcontrol}
        \label{cor:vm-control}
        Under Assumption \ref{ass:vm-regularity-1}, if
        \begin{equation}
            C_{\infty,2}(M)\|f(\cdot,x(\cdot,x_0)) - \hat{f}(\cdot, x(\cdot,x_0))\|_{L^2([0,T];\R^n)} < \delta,
            \label{eq:vm-control-condition}
        \end{equation}
        for every $x_0 \in \X_0$, then it holds that    
        \begin{equation}
            \lx(\hat{f}; x_0) \leq C_{2,2}^2(M)\left ( \lVM(\hat{f}; x_0) + C^{\rm VM}_1 \dt^2 + C^{\rm VM}_2\dt^4 \right ).
            \label{eq:vm-control}
        \end{equation}
    \end{restatable}
\end{emphbox}

The proof is provided in Appendix \ref{app:velocity-matching-3}.
The right hand side of \eqref{eq:vm-control} is computable, and requires no simulation of the learned dynamics, as it only depends on the training loss $\lVM$ plus an explicit discretization bias, with constants depending on the hypothesis class $\mathcal{F}$ alone. \hl{Neither term requires solving \eqref{eq:neural-ode}, so that the bound is \textbf{simulation-free}, overcoming the need for backpropagation through an ODE solver, as instead required by the trajectory loss $\lx$ on the left hand side.} The smallness condition \eqref{eq:vm-control-condition} is a technical requirement, as it yields confinement of the learned flow within $K$, under the $\delta$-neighborhood of Assumption \ref{ass:vm-regularity-1}, without which the stability estimates of Lemma~\ref{lemma:continuous-time-stability-estimates} would not hold. A similar argument has been employed in the proof of Theorem \ref{thm:universal-approximation-flow}. 

As a consequence of the previous result, we establish the consistency of the minimizer $\hat{f}_\ast$ of the VM objective $\LVM$ with the true velocity field $f$, in the $L^2$ sense, as $\dt \to 0$.

\begin{emphbox}
\begin{restatable}[Consistency of VM minimizer]{corollary}{vmconsistency}
    \label{cor:vm-consistency}
    Under Assumption \ref{ass:vm-regularity-1}, denote by $\mu_N$ the empirical occupation measure of the observed trajectories $\{x(t, x_0^{(i)})\}_{i=1}^N$, with $x_0^{(i)} \sim \rho_0$, and let $\hat{f}_\ast\in\arg\min_{\hat{f} \in \mathcal{F}} \LVM[\hat{f}]$. Then
    \begin{equation}
        \|\hat{f}_\ast - f\|_{L^2(\mu_N; \R^n)}^2 \leq 2(C^{\rm VM}_1 \dt^2 + C^{\rm VM}_2\dt^4).
    \end{equation}
\end{restatable}    
\end{emphbox}

The proof is provided in Appendix \ref{app:velocity-matching-3}. The result is stated in terms of the empirical occupation measure $\mu_N$, for a finite set of observed trajectories, associated to sample of initial conditions of size $N$ from the prescribed distribution $\rho_0$, relying on the notation of Section~\ref{sec:velocity-learning-1}.

\subsection{Implicit regularization and Jacobian matching}
\label{sec:velocity-matching-4}

The analysis carried out in the previous sections relies on a standing bound on the Lipschitz constant of the learned velocity field $\hat{f}$, which, for a continuously differentiable field on a convex domain, is equivalent to a bound on the norm of its Jacobian $\partial_x \hat{f}$. Such a bound enters the stability estimates of Lemmas~\ref{lemma:continuous-time-stability-estimates} and \ref{lemma:discrete-time-stability-estimate}, through the Gronwall constants depending on $\lip{\hat{f}}$, and is required by the approximation results of Section~\ref{sec:velocity-matching-3}. Likewise, Assumption~\ref{ass:vm-regularity-1}, on which the subsequent analysis of the VM objective relies, enforces a uniform bound of a similar nature.
In practice, such a bound may be enforced through a penalty on the norm of $\partial_x \hat{f}$. However, the proof of Theorem \ref{thm:universal-approximation-flow}, carried out in $W^{1,\infty}$-norm, makes clear that controlling the norm of the learned Jacobian alone is not sufficient, as what should be controlled is its distance from the true Jacobian $\partial_x f$. Indeed, arbitrary penalties on the Jacobian norm of the learned velocity field may steer the learning towards trivial solutions with arbitrarily small Jacobian norm, yielding sub-optimal approximation of the true velocity field $f$; whereas, directly enforcing \textit{Jacobian matching}, which has been shown to yield statistically accurate learning of chaotic dynamics \citep{park24}, is infeasible in the present setting, as $\partial_x f$ is not accessible, and its computation would require $O(n^2)$ complexity. 

In this section, we show that, relying on the bias-variance decomposition \eqref{eq:vm-loss-decomposition}, VM implicitly provides a form of Jacobian control, and that the role of the perturbation $\tau$ along the interpolant decouples. Specifically, the bias term gives rise to a penalty on the mismatch between the Jacobians' action on the true velocity, yielding a property which we refer to as \textit{implicit Jacobian matching}. The variance term penalizes the action of the learned Jacobian on the true velocity, hence acting as a regularizer.
As the following argument relies on a Taylor expansion of the bias and variance terms, we require further regularity, which is enforced by the following assumption.

\begin{assumption}
\label{ass:vm-regularity-2}
Let $K$ be as in Assumption \ref{ass:vm-regularity-1} and $M>0$. Both target and learnable velocity fields, $f$ and $\hat{f}$, are autonomous and belong to the function class
\begin{equation}
        \mathcal{F} = \left \{g \in W^{2,\infty}_{\rm loc}(\R^{n}; \R^n) :  \|g\|_{W^{2,\infty}( K; \R^n)} \leq M\right \}.
        \label{eq:hypothesis-class-2}
\end{equation}
\end{assumption}

The following result formalizes the above observations.

\begin{emphbox}
    \begin{restatable}[Implicit regularization]{proposition}{implicitregularization}
        \label{prop:vm-implicit-regularization}
        Under Assumption \ref{ass:vm-regularity-2}, for every $x_0 \in \X_0$ and $k = 0, \ldots, \Nt-1$, the bias term in \eqref{eq:vm-loss-decomposition} of the per-sample VM loss $\lVM$ satisfies the expansion
        \begin{equation}
            \big|v_k(x_0) - \E_\tau [\hat{f}_k(\tau;x_0)] \big|^2 = \frac{\dt^2}{4} \left | \left  (\partial_x f(x(t_k,x_0)) - \partial_x \hat{f}(x(t_k,x_0)) \right ) f (x(t_k,x_0))  \right |^2 + O(\dt^3),
            \label{eq:bias-expansion}
        \end{equation}
        provided $|f(x(t_k,x_0)) - \hat{f}(x(t_k,x_0))| \leq C \dt^2$, for some $C>0$ independent of $\dt$. 
        Analogously, the variance term in \eqref{eq:vm-loss-decomposition} satisfies the expansion
        \begin{equation}
            \var_\tau[\hat{f}_k(\tau;x_0)] = \frac{\dt^2}{12}\left |\partial_x \hat{f}(x(t_k,x_0)) f(x(t_k,x_0))\right |^2 + O(\dt^3).
            \label{eq:variance-expansion}
        \end{equation}
    \end{restatable}
\end{emphbox}

The proof is provided in Appendix \ref{app:velocity-matching-4}. The first claim holds in the regime where the learned velocity field is $O(\dt^2)$-close to the true one at the observed points. There, the remaining terms in the bias expansion are of higher order, and the leading order term reduces to \eqref{eq:bias-expansion}, namely a directional Jacobian matching condition. As a consequence, in this regime, the VM objective implicitly enforces a penalty on the mismatch between the Jacobians' action on the true velocity, which, while weaker than full Jacobian matching, emerges naturally from the stochastic collocation strategy employed by VM, as shown empirically in Section~\ref{sec:numerical-results}. 
The second claim holds without the closeness condition, and shows that the variance term induces a penalty on the action of the Jacobian of the learned velocity field on the true velocity, weighted by the variance $\var_\tau[\tau] = \dt^2/12$ of the perturbation $\tau\sim \mathcal{U}[0, \dt]$. By contrast with an arbitrary penalty on the Jacobian norm, it constrains only the variation of the learned velocity field along the flow, leaving the directions transverse to the flow unconstrained. This is a direct contribution of the random perturbation along the interpolant, as a deterministic collocation rule with a fixed offset $\tau$ would yield a vanishing variance term, so that an equivalent control would have to be enforced explicitly.

To understand the effect of such a penalty on the learned flow, we may proceed formally. If the learned velocity field is sufficiently close to the true one, replacing the true trajectory and velocity with their learned counterparts in~\eqref{eq:variance-expansion} yields
\begin{equation*}
\partial_x \hat{f}(x(t_k,x_0)) f(x(t_k,x_0)) \approx \partial_x \hat{f}(\hat{x}(t_k,x_0)) \hat{f}(\hat{x}(t_k,x_0)) = \partial_t^2 \hat{x}(t_k,x_0),
\end{equation*}
so that the variance term acts as a penalty on the acceleration of the learned system. In this sense, its role is akin to that of cubic spline interpolation, where the natural interpolant minimizes the acceleration among sufficiently smooth interpolants. 
This analogy is only qualitative, since VM does not enforce interpolation of the learned trajectory at the observed points, its main aim being to reframe the learning problem as velocity regression rather than trajectory fitting. Nevertheless, it provides an intuitive interpretation of the regularization effect induced by the stochastic collocation strategy on the learned flow.
Finally, both penalties are of order $\dt^2$, so that the implicit regularization does not degrade the order of convergence of the VM loss to the theoretical velocity loss established in Theorem~\ref{thm:vm-order}.
\section{Numerical results}
\label{sec:numerical-results}

Having characterized the velocity matching (VM) objective analytically, we now assess its behavior empirically. In 
Section~\ref{sec:velocity-learning}, we showed that the trajectory error is controlled by a simulation-free quadratic objective over velocities; in Section~\ref{sec:velocity-matching}, we showed that VM approximates such an objective to second order in the temporal step size of the observations, while implicitly matching the action of the Jacobians of the learned and true velocity fields along the flow.
In this section, we investigate these properties numerically, by comparing VM with trajectory-based strategies reviewed in Section~\ref{sec:learning-strategies}, on a range of nonlinear ordinary differential equations (ODEs) and spatially semi-discretized time-dependent partial differential equations (PDEs) benchmarks. In particular, Section~\ref{sec:numerical-results-1} concerns a comparison in terms of accuracy attained by the different training strategies as a function of their training cost. 
Section~\ref{sec:numerical-results-2} assesses the continuous-time approximation properties of the learned models, namely whether they can be simulated with a time step different from the one used during training, or with arbitrary Runge--Kutta schemes, without degrading the approximation error.
Finally, Section~\ref{sec:numerical-results-4} investigates properties specific to Hamiltonian and chaotic systems, namely the preservation of invariants and the reproduction of chaotic attractors.

\paragraph{Benchmark systems.} 
As ODEs, we consider the Euler equations of a free rigid body, namely a Lie--Poisson system whose trajectories lie on the unit sphere, and which conserves both its kinetic energy and a Casimir invariant; the Stuart--Landau oscillator, which is a normal form of a supercritical Hopf bifurcation whose trajectories are attracted by a stable limit cycle; an active particle in the M\"uller--Brown potential characterized by sustained oscillations within the potential wells; and the Lorenz system in its chaotic regime. As PDEs, we consider one-dimensional viscous Burgers, Fisher--KPP and Korteweg--de Vries (KdV) equations, and the Kuramoto--Sivashinsky (KS) equation in its chaotic regime. Their spatial semi-discretization yields a high-dimensional ODE system, so that the learning problem falls within the setting of Section~\ref{sec:trajectory-learning}, with the learned velocity field acting on the discretized state. Further details are provided in Appendix~\ref{app:experimental-details}.

\paragraph{Experimental setup.} A common setup is adopted for all numerical experiments, in which models trained by minimizing the VM loss~\eqref{eq:velocity-matching-objective-loss}, according to Algorithm~\ref{alg:velocity-matching}, are compared with models trained on the multiple-shooting rollout loss~\eqref{eq:rollout-loss-multiple-shooting} in five configurations, spanning the single-step and multi-step settings of Section~\ref{sec:learning-strategies}: teacher forcing (TF), namely $L=1$; fixed shooting lengths $L=5$ and $L=10$; a randomized shooting length drawn uniformly in $\{1,\dots,20\}$ at each iteration; and a curriculum strategy, in which the shooting length grows linearly from $L=1$ to $L=50$ along training. Simulation of the learned dynamics in rollout training relies on explicit Euler updates, the typical scheme adopted for neural surrogates. For a given benchmark system, all training strategies share the same data, architecture, and optimization hyperparameters, so that the resulting models differ only in the objective being minimized.
The accuracy of the learned dynamics is measured by the relative $L^2$ error  between reference and learned trajectories over the time interval $[0,T]$, with $T$ denoting either the validation or test horizon,
\begin{align}
\mathcal{E}_{\mathrm{traj}} &:= \E_{x_0\sim\rho_0}\frac{\|x(\cdot, x_0) - \hat{x}(\cdot, x_0)\|_{L^2([0,T];\R^n)}}{\|x(\cdot, x_0)\|_{L^2([0,T];\R^n)}}, && \text{(Trajectory error)}
\label{eq:traj-error}
\end{align}
where $x(t,x_0)$ and $\hat{x}(t,x_0)$ denote the solutions of the target and learned systems, respectively. 
In particular, training and validation share the same temporal horizon, $T_\mathrm{valid} = T_\mathrm{train}$, whereas testing is performed over a horizon twice as long, $T_\mathrm{test} = 2T_\mathrm{train}$, so that the test set assesses both generalization to unseen initial conditions and time extrapolation.
For the ODE benchmarks, the true velocity field $f$ and its Jacobian $\partial_x f$ are available, so that the accuracy of the learned field itself can also be measured, through the following relative errors
\begin{align}
\mathcal{E}_{\mathrm{vel}} &:= \E_{x_0\sim\rho_0}\frac{\|f(x(\cdot, x_0)) - \hat{f}(x(\cdot, x_0))\|_{L^2([0,T];\R^n)}}{\|f(x(\cdot, x_0))\|_{L^2([0,T];\R^n)}},&& \text{(Velocity error)} \label{eq:vel-error}\\
\mathcal{E}_{\mathrm{Jac}} &:= \E_{x_0\sim\rho_0}\frac{\|\partial_x f(x(\cdot, x_0)) - \partial_x\hat{f}(x(\cdot, x_0))\|_{L^2([0,T];\R^{n\times n})}}{\|\partial_xf(x(\cdot, x_0))\|_{L^2([0,T];\R^{n\times n})}}, && \text{(Jacobian error)}\label{eq:jac-error}
\end{align}
with $\R^{n\times n}$ equipped with the Frobenius norm. Note that both quantities are evaluated along the reference trajectories $x(t,x_0)$, rather than the learned ones, as they measure the discrepancy between the velocity fields on the support of the data, independently of the error accumulated by integrating the learned dynamics. Together, the velocity and Jacobian errors provide an $L^2$ proxy, along the reference trajectories, for the $W^{1,\infty}$ error of $\hat{f}$ on which our approximation results rely. 
In practice, we report the empirical counterparts of~\eqref{eq:traj-error}--\eqref{eq:jac-error}, in which the expectation over initial conditions and the time integrals are replaced by sample averages.
Each training objective was run with three independent seeds, and, for each run, the model attaining the lowest validation trajectory error \eqref{eq:traj-error} was retained for testing. 
Further details are given in Appendix~\ref{app:experimental-details}.

\subsection{Accuracy, fast convergence and implicit regularization}
\label{sec:numerical-results-1}

\begin{figure}[!t]
    \centering
    \includegraphics[width=1\textwidth]{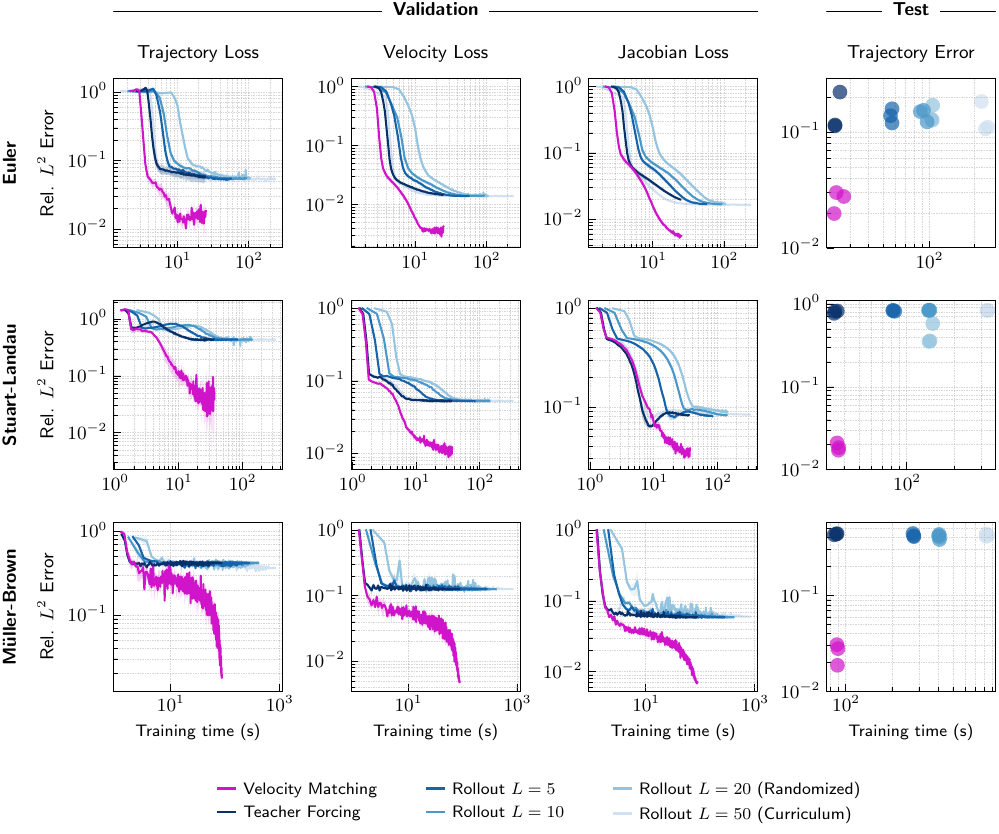}
    \caption{\textbf{Validation and test errors - ODEs}. (\emph{Left}) Validation trajectory, velocity and Jacobian errors as functions of wall-clock training time, for Euler, Stuart--Landau and active M\"uller--Brown systems. Solid lines denote the mean, and shaded areas the standard deviation, over three independent runs. (\emph{Right}) Test trajectory error against the total training time, for each independent run. Trajectory-based strategies saturate at a common error level, whereas VM attains errors up to one order of magnitude lower, at a training cost comparable with that of TF.}
    \label{fig:validation-losses-test-error-ode}
\end{figure}

We first compare the proposed VM objective against TF and rollout training strategies in terms of accuracy and of the computational cost required to attain it, under a common iteration budget, on both ODE and PDE benchmarks. For the ODE benchmarks, we then assess how the learned velocity fields and their Jacobians compare with the true ones.

\paragraph{Accuracy and training efficiency.} 
Figure~\ref{fig:validation-losses-test-error-ode} reports, for the Euler, Stuart--Landau and active M\"uller--Brown systems, the evolution of the three validation errors~(\ref{eq:traj-error}-\ref{eq:jac-error}) against elapsed training time, alongside the test trajectory error attained by each seed, plotted against the total training time of the corresponding run. The former shows how fast each objective reduces the validation errors during training, whereas the latter relates the final test accuracy to the overall computational cost under the common iteration budget.
The corresponding validation and test errors are collected in Table~\ref{table:results-ode} in Appendix~\ref{app:experimental-details}. Two observations hold across all three test cases. First, VM and TF have a comparable cost per iteration, since both evaluate $\hat{f}$ once per sample, as discussed in Section~\ref{sec:velocity-matching-1}, while, by contrast, the cost of the rollout-based strategies grows with the shooting length $L$, as their gradients are obtained by backpropagating through $L$ explicit Euler steps. Second, all trajectory-based strategies saturate at a common error level, in all three metrics and irrespective of $L$, whereas the validation errors of VM keep decreasing well below such a level, within a training time comparable with that of TF.  
On the test set, VM reduces the trajectory error by up to an order of magnitude relative to rollout strategies, consistently across test cases and seeds, whereas the latter yield test errors of $10^{-1}$ or larger. A comparable reduction is observed in the velocity and Jacobian errors.
These observations are consistent with the analysis carried out in the previous sections. Although VM never simulates the learned dynamics during training, it attains the lowest trajectory error, as expected from Corollary~\ref{cor:vm-control}, by which the trajectory loss is controlled by the VM loss up to a discretization bias of order $\dt^2$. By contrast, the TF objective approximates the theoretical velocity objective only to first order in $\dt$ (Theorem~\ref{thm:tf-order}), with its minimizers inheriting the corresponding bias. Longer shooting horizons do not remove this bias, since the same explicit Euler update is adopted, but rather reshape the loss landscape, as discussed in Section~\ref{sec:degeneracies-trajectory-objective}.

\begin{figure}[!t]
    \centering
    \includegraphics[width=1\textwidth]{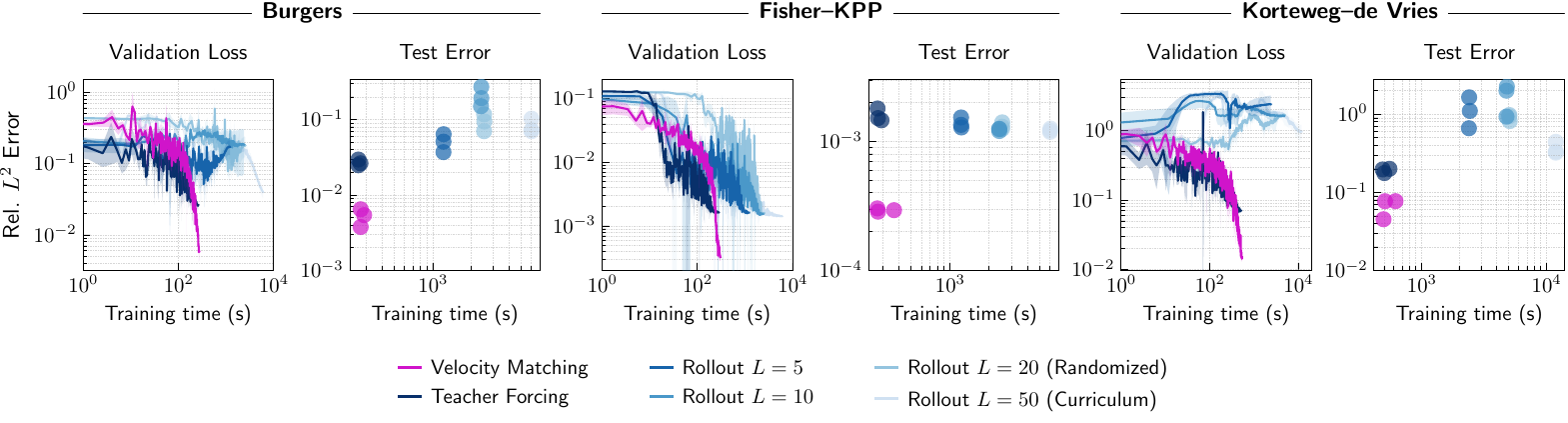}
    \caption{\textbf{Validation and test errors - PDEs}.  Validation trajectory error against elapsed training time (\emph{left}), and test trajectory error against total training time (\emph{right}), for the Burgers, Fisher--KPP, and KdV equations, respectively. Solid lines and shaded areas denote the mean and standard deviation over three seeds; each marker in the right panels corresponds to a single seed.}
    \label{fig:validation-losses-test-error-pde}
\end{figure}

Analogously, Figure~\ref{fig:validation-losses-test-error-pde} reports, for the Burgers, Fisher--KPP, and KdV equations, the evolution of the validation trajectory error against elapsed training time, alongside the test trajectory error attained by each seed against the total training time of the corresponding run. In this case, the spatial discretization of the PDE solution $u(x, t)$\footnote{In the PDE setting, we denote by $u$ the solution and by $x$ the spatial coordinate, the latter not to be confused with the state of~\eqref{eq:ode}, which here corresponds to the discretized solution $u_h$, with $h$ the spatial discretization step.}
yields a $1024$-dimensional time-dependent state $u_h(t)$, whose evolution is governed by a system of the form~\eqref{eq:ode}, in which the vector field $f$ corresponds to the spatially discretized differential operator on the right-hand side of the PDE, approximated by the learned one $\hat{f}$.
Validation and test trajectory errors for each equation and training strategy, averaged across seeds, are collected in Table~\ref{table:results-pde} of Appendix~\ref{app:experimental-details}, while Figures~\ref{fig:burgers-errors}--\ref{fig:kdv-errors} report the pointwise absolute errors on four test trajectories per equation.
Unlike in the ODE test cases, the dependence of the validation and test errors on the shooting length $L$ is strongly problem-dependent. For the Fisher--KPP equation, both errors improve slightly as $L$ increases, whereas for the Burgers and KdV equations rollout training exhibits instabilities that become more pronounced for larger $L$. 
By contrast, VM -- using the same architecture and optimization hyperparameters -- yields stable optimization procedure on all three test cases with the validation trajectory error steadily decreasing throughout training without the instabilities observed for rollout objectives. The retained models attain the lowest test errors across equations and seeds. This is consistent with the analysis of Section~\ref{sec:degeneracies-trajectory-objective}, which shows that the smoothness of trajectory-based objectives deteriorates as the rollout length $L$ increases, with gradients that vanish or explode exponentially in $L$, thus hindering the convergence of gradient-based optimization.

\begin{figure}[!t]
    \centering
\includegraphics[width=1\textwidth]{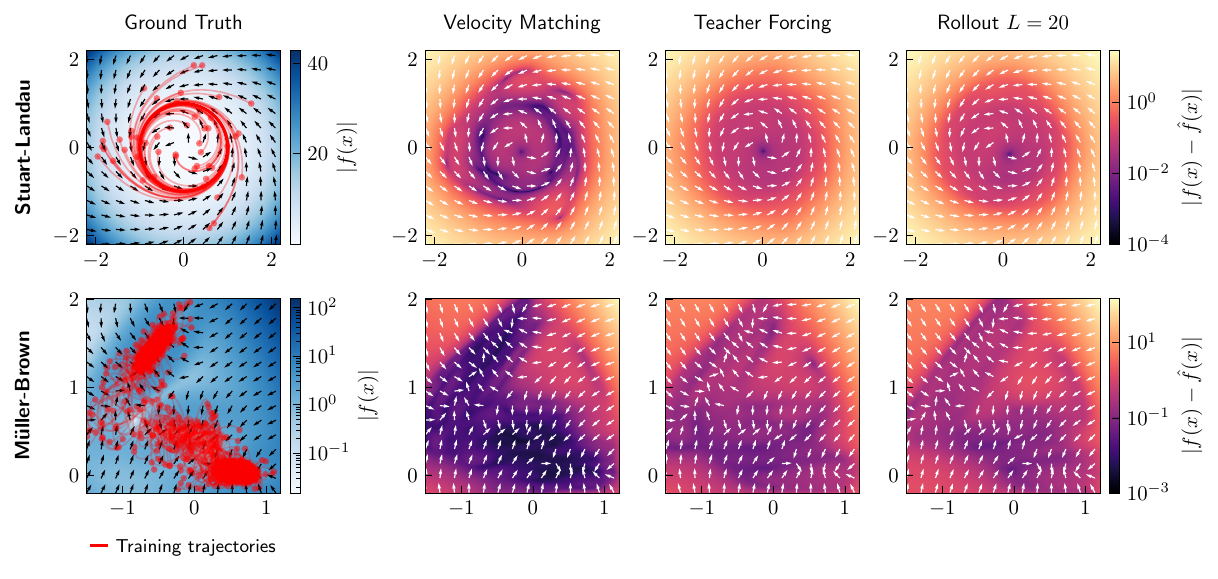}
    \caption{\textbf{Learned velocity fields.} Stuart--Landau (top) and active M\"uller--Brown (bottom) systems. (\emph{Left})~Magnitude and direction of the true velocity field, together with the training trajectories. (\emph{Right})~Pointwise error $|f(x) - \hat{f}(x)|$ of the velocity fields learned by VM, teacher forcing (TF), and randomized rollout with $L = 20$, over the two-dimensional state space $(x_1, x_2)$; arrows denote the normalized learned velocity field. For the active M\"uller--Brown system, the learned acceleration field is shown over the position plane $(x_1, x_2)$, averaged over velocity directions of magnitude $|\dot{x}| = v_0$, at which the active friction vanishes and the true acceleration reduces to the force $-\kappa \nabla V(x)$; the error is averaged over the same directions. The error of VM is lowest on the support of the training data, where it is one to two orders of magnitude below that of the other strategies.}
    \label{fig:vector-fields}
\end{figure}

\paragraph{Velocity approximation.} The velocity error~\eqref{eq:vel-error} measures the accuracy of the learned field along the reference trajectories in an averaged $L^2$ sense. Figure~\ref{fig:vector-fields} complements it by showing the pointwise error over the state space of the Stuart--Landau and active M\"uller--Brown systems. The error of VM is lowest over the region spanned by the training trajectories, namely around the limit cycle and along the transients spiraling towards it for the former, and within the potential wells for the latter, and grows away from this region.
This agrees with the variational characterization of Proposition~\ref{prop:variational-characterization}, by which the minimizer of the velocity objective is determined only on the support of the occupation measure~\eqref{eq:occupation-measure}, and is left unconstrained elsewhere. Corollary~\ref{cor:vm-consistency} transfers this characterization to the empirical VM objective, whose minimizer is $O(\dt^2)$-consistent with $f$ on the same support. Although TF also admits a velocity regression reading, its validation velocity error plateaus at the same level as that of rollout training, as shown in Figure~\ref{fig:validation-losses-test-error-ode}, suggesting that its $O(\dt)$ bias, compounded by the spectral constraints induced by the explicit Euler step, limits the attainable accuracy. The gap between the pointwise errors of VM and rollout training in Figure~\ref{fig:vector-fields} reflects a structural difference between the two objectives, as rollout training does not recover the true velocity field $f$, but rather the field that, when numerically simulated with the scheme and time step used during training, best reproduces the observed trajectories. This aspect will be further addressed in Section~\ref{sec:numerical-results-2}, since continuous-time approximation hinges on the capability of recovering $f$ itself.

\paragraph{Implicit Jacobian matching.} As shown in Proposition~\ref{prop:vm-implicit-regularization}, the bias term of the VM loss aligns the action of the learned Jacobian with that of the true Jacobian along the flow, without having access to the latter, while its variance term penalizes the variation of the learned field along the flow. Neither mechanism is available to TF, which evaluates $\hat{f}$ only at the observed states, or to rollout training, which relies on the same explicit Euler update. The Jacobian errors~\eqref{eq:jac-error} in Figure~\ref{fig:validation-losses-test-error-ode} are consistent with this picture. Indeed, although no information about the Jacobians is provided, the Jacobian error of VM decreases steadily throughout training, and, as reported in Table~\ref{table:results-ode}, its test value is lower than that of the trajectory-based strategies by a factor of 3--9 across the three test cases.
We stress that Jacobian accuracy matters beyond velocity accuracy, since the Jacobian of the vector field governs the local behavior of the dynamics, determining the stability of equilibria, as well as the rates at which nearby trajectories contract or separate. An accurate Jacobian thus preserves the qualitative structure of the target dynamics, and keeps the stability constants entering Lemma~\ref{lemma:continuous-time-stability-estimates} close to those of the true system.

\subsection{Continuous-time approximation}
\label{sec:numerical-results-2}

\begin{figure}[!t]
    \centering
\includegraphics[width=1\textwidth]{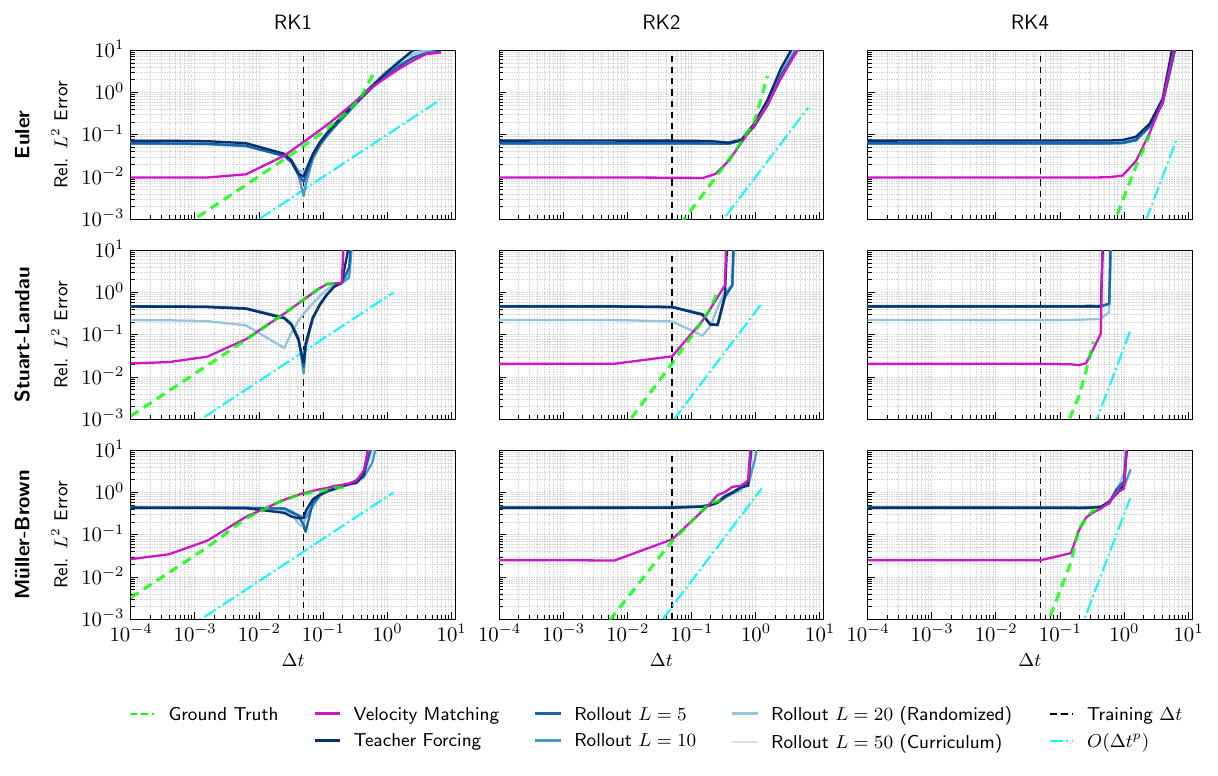}
    \caption{\textbf{Continuous-time approximation}. Test trajectory error \eqref{eq:traj-error} as a function of the inference step size $\dt$, when the learned dynamics are simulated with the explicit Euler (RK1), Heun (RK2) and classical fourth-order Runge--Kutta (RK4) schemes (columns), for the Euler, Stuart--Landau and active M\"uller--Brown systems (rows). The dashed vertical line marks the training step size, and the green dashed curve labeled \emph{Ground Truth} is the error obtained by integrating the true velocity field $f$ with the same scheme, while the dash-dotted line indicates the rate $\dt^p$ of a scheme of order $p$. The error of VM follows the error behavior of the true field down to a plateau that does not depend on the scheme, whereas trajectory-based strategies are accurate only at the training step size and with explicit Euler.}
    \label{fig:continuous-time-approximation}
\end{figure}

Formulating a model in continuous time, as an ODE, does not by itself guarantee that, once trained on observations at a fixed step, it can be simulated with other step sizes and numerical schemes without loss of accuracy. Theorem~\ref{thm:continuous-time-approximation} guarantees this property for a learned velocity field that is close to $f$ in $W^{1,\infty}$, through the decomposition~\eqref{eq:error-decomposition} of the error into the continuous-time error of the learned flow and the discretization error of the scheme. We assess the continuous-time approximation capability of the models learned by VM, TF, and the selected rollout strategies by studying the test trajectory error as a function of the step size $\dt$ used at inference. Figure~\ref{fig:continuous-time-approximation} reports this error against $\dt$ for the Euler, Stuart--Landau, and active M\"uller--Brown systems, with $\dt$ ranging from $10^{-4}$ to well beyond the training step size, and with each column corresponding to an explicit Runge--Kutta scheme of order $p \in \{1, 2, 4\}$. In all cases, the reference trajectories are computed with step size $10^{-4}$ and interpolated onto the coarser grids. Beyond assessing whether the learned model can be simulated on temporal grids other than the training one without retraining, a critical computational aspect for neural surrogates, this experiment also serves as a proxy for the closeness of $\hat{f}$ to $f$. By~\eqref{eq:error-decomposition}, the error of a model close to $f$ should follow that of $f$ itself for large $\dt$, where the discretization error dominates, and saturate at the continuous-time error of the learned flow as $\dt \to 0$.
We therefore restrict this experiment to the ODE benchmarks, where $f$ is available in closed form, so that the error attained by integrating $f$ with the same schemes and step sizes (dashed green lines) is used as a reference.
Note that, since Theorem~\ref{thm:continuous-time-approximation} is an approximation result, as it requires $\hat{f}$ to be close to $f$ on a region containing the trajectories of interest, continuous-time approximation would naturally be assessed on the training trajectories. Here, we instead evaluate the error on an additional test set, since such a property is of practical interest only if it holds for unseen initial conditions, thus yielding a more challenging setting that combines approximation and generalization errors.

The error associated with VM exactly follows the decomposition~\eqref{eq:error-decomposition}. For large step sizes, it coincides with the error obtained by integrating the true velocity field with the same scheme, and decreases at rate $\dt^p$, since the discretization error dominates. Following such a decomposition, this formally indicates that a field $\hat{f}$ sufficiently close to $f$ in $W^{1,\infty}$ has been learned, even though everything here is measured in $L^2$ norms. As the step size decreases, it saturates at a plateau that does not depend on the scheme, but rather on the continuous-time approximation error $\|x - \hat{x}\|_{L^\infty}$ of the learned flow, and is in all cases smaller than or equal to the error attained at the training discretization, so that the accuracy is not affected, thus yielding a model that can handle finer discretizations. Consistently, the plateau is reached at larger step sizes as the order $p$ increases, since the discretization error falls below the continuous-time error sooner for higher-order schemes.

The errors associated with rollout strategies behave differently. When integrated with explicit Euler (RK1), the scheme used for training, their error attains a sharp minimum at the training step size, and increases as the step size is refined, until it saturates at a plateau well above both their minimum and the VM plateau, which instead retains the training accuracy. Moreover, higher-order schemes do not improve their accuracy at any step size. Thus, the velocity field learned via rollouts is accurate only when combined with the scheme and step size used for training. So, the picture outlined in the previous section repeats, since, rather than approximating $f$, the model learns an Euler update that compensates for the discretization error of the scheme, thus overfitting the training discretization.
This is the empirical counterpart of the decomposition~\eqref{eq:tf-residual-decomposition}, which shows that TF minimizes the residual of the numerical scheme rather than the mismatch between velocity fields, and is in agreement with the observations of~\citet{krishnapriyan23}. By contrast, the VM objective approximates the theoretical velocity objective $\Lf$ to second order in $\dt$ without reference to any integrator (Theorem~\ref{thm:vm-order}), yielding a learned velocity field that can be simulated with arbitrary schemes and step sizes at inference.

\subsection{Learning Hamiltonian and chaotic dynamics}
\label{sec:numerical-results-4}

\begin{figure}[t]
    \centering
    \includegraphics[width=1\textwidth]{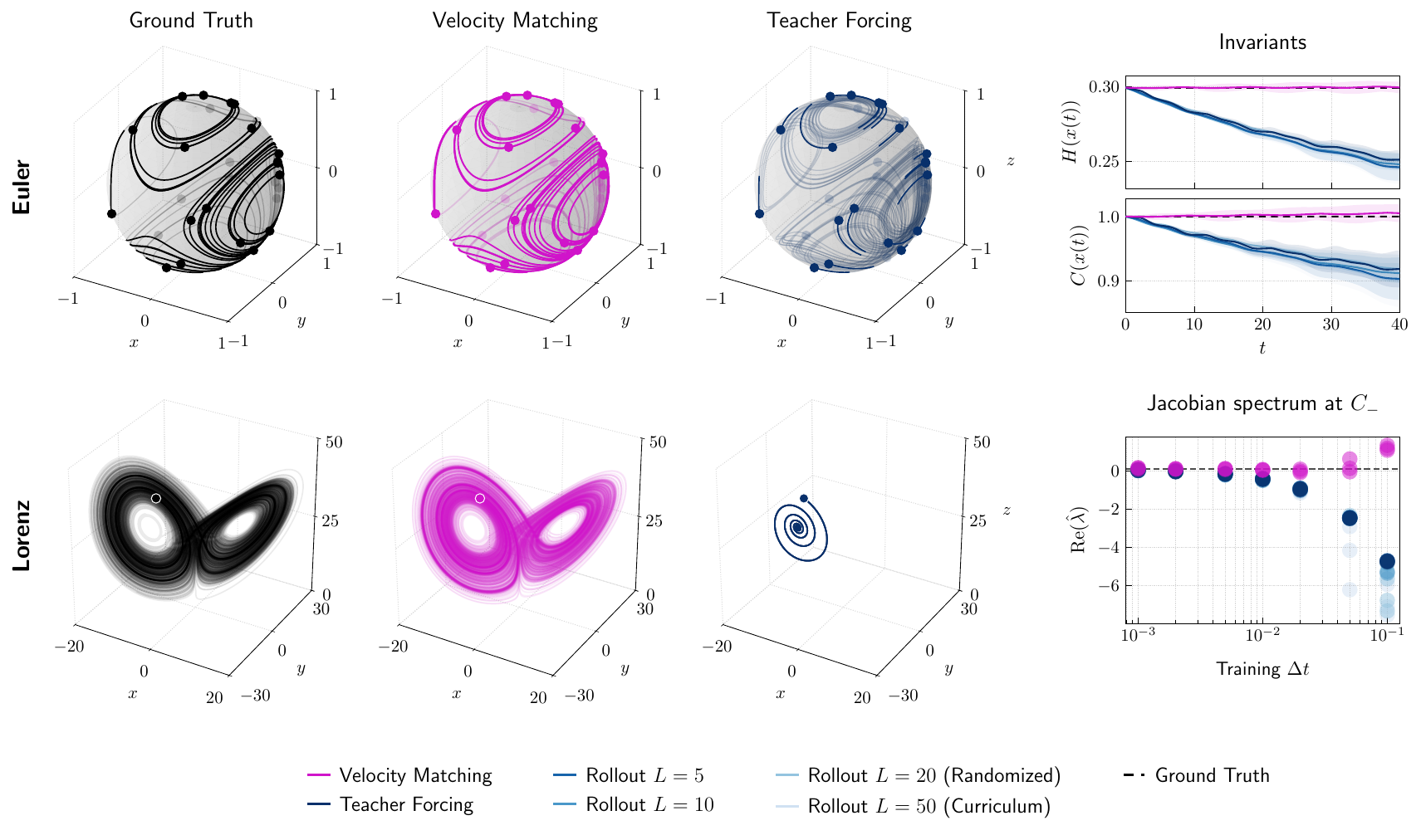}
    \caption{\textbf{Hamiltonian and chaotic dynamics.} (\emph{Top}) Euler equations for the free rigid body. Trajectories of the ground truth and of the dynamics learned by VM and TF on the unit sphere, from the same test initial conditions (markers), and evolution of the energy $H$ and of the Casimir $C$ along the learned trajectories, averaged over seeds, with shaded areas denoting the standard deviation. VM preserves both invariants, whereas TF and rollout strategies dissipate them, so that trajectories drift off the sphere. (\emph{Bottom}) Lorenz system. Orbits of the ground truth and of the dynamics learned by VM and TF with training step size $\dt = 2\cdot 10^{-2}$, from the same initial condition (marker), and real part of the complex-conjugate eigenvalues of the learned Jacobian at the critical point $C_-$ against the training step size. As $\dt$ increases, TF and rollout strategies learn increasingly negative real parts, turning $C_-$ into a stable focus onto which trajectories collapse, whereas VM retains a nonnegative real part and reproduces the attractor. Dashed lines denote the true values.}
\label{fig:euler-lorenz}
\end{figure}

Finally, we consider systems characterized by relevant qualitative features, possibly emerging over long times, namely the preservation of invariants and the structure of chaotic attractors. Since none of the tested objectives enforces these features explicitly, they provide an indirect test of the accuracy of the learned velocity field and its Jacobian.

\paragraph{Euler.} The Euler equations of a free rigid body conserve the kinetic energy, namely the Hamiltonian $H$, and the Casimir invariant $C(x)=|x|^2$, so that trajectories lie on the unit sphere. As shown in Figure~\ref{fig:euler-lorenz}, the trajectories learned by VM conserve both invariants up to small fluctuations, over a test horizon twice as long as the training one. By contrast, those learned by TF and by the rollout-based strategies drift monotonically, and lose approximately $15\%$ of $H$ and $10\%$ of $C$ by the end of the horizon, which reveals a systematic dissipation in the learned dynamics. Such dissipation is consistent with Proposition \ref{prop:jacobian-spectral-constraint}, by which training through explicit Euler updates constrains the spectrum of the learned Jacobian to lie within the stability region of the scheme, and hence towards contractive dynamics. Note that, for conservative dynamics such as the rigid body, the linearization about a stationary rotation around the major or minor axis~\citep[Theorem~15.9.1]{marsden99} has purely imaginary eigenvalues~\citep[Eq.~(15.9.12)]{marsden99} when restricted to the tangent space at the equilibrium of the unit sphere in $\R^3$, which therefore lie outside the stability region of explicit Euler for every step size. By contrast, the VM objective does not push the spectrum of the learned Jacobian off the imaginary axis, since its bias term is the residual of the average vector field method (Proposition~\ref{prop:vm-csrk}), whose stability function satisfies $|R(z)| = 1$ on the imaginary axis. As a result, the learned system preserves the structural properties of the reference dynamics.

\paragraph{Lorenz.} 
In the chaotic regime, where pointwise trajectory errors become uninformative beyond a few Lyapunov times, the relevant question is whether the learned dynamics reproduces the attractor. We consider the case of the Lorenz system \citep{lorenz63}, with training trajectories starting from initial conditions sampled from a concentrated distribution $\rho_0\approx \delta_{x_0}$. We vary the training step size $\dt$ from $10^{-3}$ to $10^{-1}$, and report the orbits of the learned systems for $\dt = 2\cdot 10^{-2}$ in Figure~\ref{fig:euler-lorenz}, alongside the spectrum of the Jacobian of the learned dynamics at the critical point $C_-$, for the considered range of training step sizes $\dt$.
The latter explains the collapse of TF trajectories onto $C_-$ in Figure~\ref{fig:euler-lorenz}, and analogously for the rollout strategies. For the true dynamics, the two critical points $C_\pm$, other than the origin, are unstable \citep{sparrow82}. Their complex-conjugate eigenvalues have a small, but positive, real part, which makes trajectories escape from their neighborhood and switch between the two regions of the attractor. 
As the training step size increases, TF and the rollout-based strategies learn complex eigenvalues with an increasingly negative real part, so that $C_-$ becomes a stable equilibrium, inducing the collapse of trajectories onto it. This stabilization, a consequence of the spectral constraint of Proposition~\ref{prop:jacobian-spectral-constraint}, prevents the chaotic dynamics and its attractor from being learned. By contrast, VM recovers the real part of these eigenvalues accurately over most of the range of step sizes, in line with the implicit Jacobian matching mechanism of Proposition~\ref{prop:vm-implicit-regularization}. 
We stress that, although formal, the connection between VM and Jacobian matching strategies, established through the implicit regularization result of Proposition~\ref{prop:vm-implicit-regularization}, becomes relevant when learning statistically accurate chaotic dynamics.
Indeed, \citet{park24} showed, through an ergodic-theoretic analysis based on shadowing, that small errors in the vector field alone do not guarantee the recovery of the invariant measure, whereas closeness of both the learned dynamics and its Jacobian does, proposing, consistently, to augment the regression loss with a Jacobian matching term. 
This approach has recently been extended by \citet{kang26}, who match Jacobians at randomly perturbed states to implicitly enforce second-order consistency. 
Both strategies, however, require access to the true Jacobian at the training states, and involve forming Jacobian matrices, at a cost of $O(n)$ derivative evaluations and $O(n^2)$ memory per training state.
By contrast, VM only matches the action of the Jacobian along the flow, which stochastic collocation extracts from trajectory data alone at the cost of a single evaluation of $\hat{f}$ per collocation point, without forming any Jacobian.

\begin{figure}[t]
    \centering
    \includegraphics[width=1\textwidth]{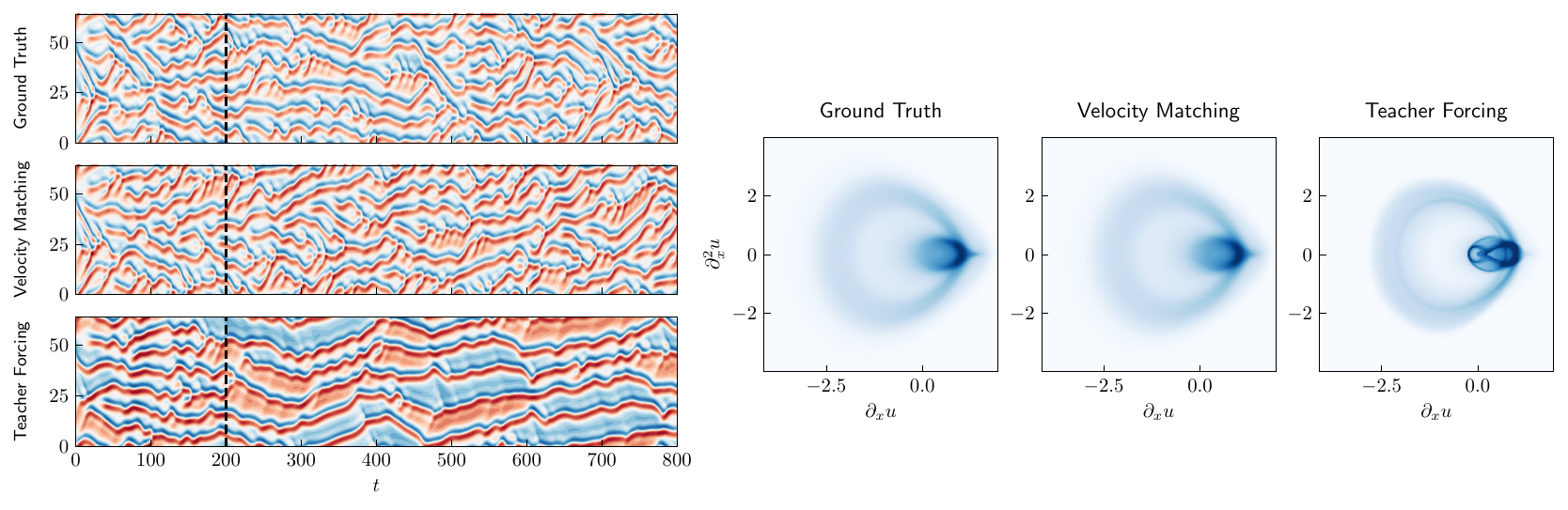}
    \caption{\textbf{Kuramoto--Sivashinsky.} (\emph{Left})~Evolution of the discretized solution $u_h(t)$ for the ground truth and for the dynamics learned by VM and TF, from the same test initial condition, over the test temporal interval $[0, 800]$; the dashed line marks the end of the training horizon, $T = 200$. (\emph{Right})~Joint probability density of the first two spatial derivatives $(\partial_x u, \partial_x^2 u)$, estimated over the test trajectories.}
\label{fig:kuramoto-sivashinsky}
\end{figure}

\paragraph{Kuramoto--Sivashinsky.} We conclude with the KS equation on a periodic domain of length $64$, discretized on $256$ grid points, whose dynamics exhibits spatiotemporal chaos. Although the comparison between VM and TF is only qualitative, this test case extends the previous discussion to a high-dimensional setting. While the fourth-order term makes the semi-discretized system stiff \citep{kassam05}, the associated modes may decay on time scales much shorter than the sampling step $\dt$, so that neither objective can identify them from the data. The two objectives differ instead in their treatment of the advective term, which transports the solution without dissipating it, and whose linearization thus has eigenvalues close to the imaginary axis \citep{leveque07}. As for the rigid body and the Lorenz system, TF compensates for the explicit Euler update by damping these modes, which acts as an additional diffusion on the learned dynamics.
The training trajectories cover the time interval $[0, 200]$, while the test trajectories extend to $T = 800$. Figure~\ref{fig:kuramoto-sivashinsky} provides a qualitative comparison of the space-time evolution of the learned dynamics and of their statistics on the attractor, through the joint probability density of the first two spatial derivatives $(\partial_x u, \partial_x^2 u)$~\citep{linot20}. Although individual trajectories decorrelate from the reference after a few Lyapunov times, as expected for chaotic dynamics, the dynamics learned by VM generates spatiotemporal chaos qualitatively similar to the true one, both within and well beyond the training horizon, and its joint density closely matches the reference. By contrast, the dynamics learned by TF evolves towards a more regular regime, dominated by low-wavenumber patterns, and its joint density concentrates around the origin, consistently with the spectral constraints induced by the explicit Euler bias, which damps the high-wavenumber content of the solution.

\section{Conclusions}
\label{sec:conclusions}

This work introduces Velocity Matching (VM), a simulation-free framework for learning continuous-time dynamical systems formulated as ordinary differential equations (ODEs), a task commonly formulated as the minimization of a trajectory-based objective. Beyond the computational cost associated with simulation-based training, we show that trajectory-based objectives suffer from structural degeneracies, which hinder gradient-based optimization and bias the learned dynamics, thus motivating the transition towards simulation-free velocity-based learning.
To this end, we develop a theoretical framework for velocity-based learning, in which the velocity field of the underlying ODE is characterized as the minimizer of a quadratic functional, and relate this objective to its trajectory-based counterpart through the introduction of suitable $L^p$-$L^q$ stability estimates, which can be read as a Lax equivalence result in a learning setting. These estimates justify simulation-free training based on velocity supervision and enable us to establish a universal approximation result for the flow of the true dynamics by neural ODEs, whose velocity field is parameterized by deep neural networks. This result in turn allows us to characterize the continuous-time approximation property of the learned dynamics. 
Relying on this theoretical framework, we characterize the properties of the VM objective. In particular, through a bias-variance decomposition, we relate the VM loss to implicit integration and show that VM provides a higher-order approximation of the velocity population risk. Combined with the stability estimates, this result yields a simulation-free bound on the trajectory loss, alongside the consistency of the VM minimizer as the step size vanishes. Moreover, the same decomposition reveals an implicit regularization of the Jacobian of the learned field along the observed trajectories, connecting VM to Jacobian matching approaches, without requiring access to the true Jacobian.
Numerical results on ODEs and semi-discretized time-dependent PDEs, featuring limit cycles as well as conservative and chaotic dynamics, support these findings, showing that VM improves accuracy over trajectory-based strategies at a lower training cost.

Several questions remain open. The effect of the stochastic collocation of VM on specific classes of nonlinear dynamics, possibly with noisy observations, remains largely unexplored, and the analysis of Section~\ref{sec:velocity-matching} is only a first step in this direction. It would be valuable to clarify for which classes of systems, whether oscillatory, conservative, dissipative, or multiscale, stochastic collocation is most beneficial, as this could guide the design of collocation distributions that are adapted to specific classes of systems. Moreover, the constants appearing in the stability estimates of Section~\ref{sec:velocity-learning-2}, on which the subsequent results rely, grow exponentially with the time horizon and the Lipschitz constant, as is common for Gr\"onwall-type estimates, and may thus become uninformative. Specializing these estimates to specific classes of dynamics, by exploiting their structure, could yield sharper bounds and more informative guarantees for learning. Finally, although we have extended the stability estimates to a Wasserstein estimate, adopting an ergodic perspective could connect the implicit Jacobian regularization of VM to statistical guarantees on the invariant measures of chaotic dynamics.

\subsection*{Acknowledgements}

NF and AM acknowledge the Project “Reduced Order Modeling and Deep Learning for the real-time approximation
of PDEs (DREAM)” (Starting Grant No. FIS00003154), funded by the Italian Science Fund (FIS) - Ministero
dell’Università e della Ricerca. AM acknowledges the project “Dipartimento di Eccellenza” 2023-2027 funded by
MUR.

\bibliography{refs}

\newpage
\appendix
\section{Additional results and proofs of Section \ref{sec:trajectory-learning}}
\label{app:trajectory-learning}

In this section we shall extend the discussion on the challenges of trajectory-based learning, presented in Section \ref{sec:degeneracies-trajectory-objective}, by providing detailed arguments for the issues related to the optimization of the trajectory-based learning objective, in both continuous- and discrete-time settings. 

\subsection{Additional results of Section \ref{sec:loss-degeneracy-continuous-time}}
\label{app:continuous-time-loss-landscape-analysis}

To address the analysis of the trajectory-based learning objective a simplified setting is considered in which the target system is linear, and the learning problem is cast as a parameter identification one. This allows to derive closed-form expressions and to characterize the criticalities of the trajectory-based learning objective, alongside their dependence on the regime of the target and learned dynamics, as well as on the temporal horizon. Our main interest is in the limiting behavior of the loss landscape, and its higher-order derivatives, in the tails of the parameter space, as well as in the infinite-horizon limit, being the main sources of degeneracy in the loss landscape. Asymptotic expressions are derived, aiming at characterizing the rates at which such degeneracies occur, and the regimes in which they are more pronounced. 
Hence, we proceed by deriving explicit expressions for the energy-normalized trajectory objective, its gradient and Hessian, by considering the problem of learning a 1-dimensional linear system, with dynamics given by $\dot{x} = \lambda x$, for some $\lambda \in \mathbb{R}$. We cast such a problem into a parameter identification one, by considering a parametric family of linear systems, given by $\dot{\hat{x}} = \hat{\lambda} \hat{x}$, with $\hat{\lambda} \in \mathbb{R}$.
We assume the initial condition $x(0) = \hat{x}(0) = x_0$ to be fixed, and $\lambda, \hat{\lambda} \neq 0$.
The trajectory-based learning objective, for a single fixed initial condition $x_0$, normalized by the average trajectory energy, is defined as
\begin{equation}
    \bar{\ell}_x(\hat{\lambda}; x_0) := \frac{\|x(\cdot, x_0) - \hat{x}(\cdot, x_0)\|_{L^2([0,T];\R)}^2}{\|x(\cdot, x_0)\|_{L^2([0,T];\R)}^2},
    \label{eq:energy-normalized-trajectory-objective}
\end{equation}
which, considering the explicit solution of the linear systems, can be expressed as
\begin{equation*}
    \bar{\ell}_x(\hat{\lambda}) = \frac{\int_0^T (e^{\lambda t} - e^{\hat{\lambda} t})^2 dt}{\int_0^T e^{2\lambda t} dt},
\end{equation*}
thus depending only on the parameters $\lambda, \hat{\lambda}$ and the time horizon $T$. The normalization is introduced to ignore the scaling effects of the solution norm, initial condition and time horizon on the loss landscape, which would otherwise dominate in certain regimes.
To perform explicit computations, we define the auxiliary  moment function $M_k(\xi) = \int_0^T  t^k e^{\xi t} dt$, for $\xi \in \mathbb{R}$ and $k \in \mathbb{N}$, allowing to express the above objective, alongside its derivatives, in closed-form.
In particular, $k$-th order moments can be computed by means of integration by parts, yielding the following recursive formula
\begin{equation*}
    M_k(\xi) = \frac{T^k e^{\xi T}}{\xi} - \frac{k}{\xi} M_{k-1}(\xi), \quad k \in \mathbb{N}_{\geq 1}, \quad \xi \in \mathbb{R}, \ \xi \neq 0,
\end{equation*}
from which closed-form expressions for orders $k = 0,1,2,3$ can be derived, resulting in
\begin{equation*}
    M_0(\xi) = \frac{e^{\xi T} - 1}{\xi}, \qquad
    M_1(\xi) = \frac{(\xi T - 1)e^{\xi T} + 1}{\xi^2},
\end{equation*}
\begin{equation*}
    M_2(\xi) = \frac{(\xi^2 T^2 - 2\xi T + 2)e^{\xi T} - 2}{\xi^3}, \qquad M_3(\xi) = \frac{(\xi^3 T^3 - 3\xi^2 T^2 + 6 \xi T - 6)e^{\xi T} +
    6}{\xi^4}.
\end{equation*}
We recall the following identities for higher-order derivatives of the moment function, verifiable by direct differentiation,
\begin{align}
\begin{split}
    \partial_\xi^k M_0(\xi) &= M_k(\xi), \quad k \in \mathbb{N}, \\
    \partial_\xi M_0(h(\xi)) &= h'(\xi) M_1(h(\xi)), \\
    \partial_\xi^2 M_0(h(\xi)) &= h''(\xi) M_1(h(\xi)) + (h'(\xi))^2 M_2(h(\xi)), \\
    \partial_\xi^3 M_0(h(\xi)) &= h'''(\xi) M_1(h(\xi)) + 3 h'(\xi) h''(\xi) M_2(h(\xi)) + (h'(\xi))^3 M_3(h(\xi)),
\end{split}
\label{eq:moment-identities}
\end{align}
with $h$ a sufficiently regular function of $\xi$. 
Additionally, we recall the asymptotic behavior of the moment function, which is useful for characterizing the tail behavior of the loss landscape. For fixed $T>0$ and $\xi \to \pm\infty$,
\begin{equation}
    \lim_{\xi \to -\infty} \xi M_0(\xi) = - 1, \qquad 
    \lim_{\xi \to -\infty} \xi^2 M_1(\xi) = 1, \qquad 
    M_k(\xi) \sim \frac{T^k e^{\xi T}}{\xi}, \quad \xi \to +\infty,
    \label{eq:moment-tails}
\end{equation}
while, in the infinite-horizon limit $T\to+\infty$, for fixed $\xi\neq0$, we have
\begin{equation}
    M_k(\xi) \to \frac{k!}{|\xi|^{k+1}}, \quad \xi < 0, 
    \qquad 
    M_k(\xi) \sim \frac{T^k e^{\xi T}}{\xi},  \quad \xi > 0.
    \label{eq:moment-infinite-horizon}
\end{equation}
The behavior of the moment is what drives the asymmetric behavior and regime dependence of the loss landscape studied in the following, being dominated by the exponential growth of the moment for positive rates, and instead by saturation for negative rates.
\paragraph{Loss landscape analysis.}
We proceed to derive closed-form expressions for the loss, gradient and Hessian of the energy-normalized trajectory objective \eqref{eq:energy-normalized-trajectory-objective}.
Relying on the explicit solution of the linear systems, with $x(t, x_0) = e^{\lambda t} x_0$ and $\hat{x}(t, x_0) = e^{\hat{\lambda} t} x_0$,
and the definition of the zero-th order moment, the loss can be expressed as
\begin{align}
    \bar{\ell}_x(\hat{\lambda}) &= 1 + \frac{M_0(2\hat{\lambda})}{M_0(2\lambda)} - 2\frac{M_0(\hat{\lambda}+\lambda)}{M_0(2\lambda)}, \label{eq:normalized-loss-closed-form}\\ 
    &= 1 + \frac{\lambda}{\hat{\lambda}} \frac{e^{2\hat{\lambda}T} - 1}{e^{2\lambda T} - 1} - \frac{4\lambda}{\lambda+\hat{\lambda}} \frac{e^{(\lambda+\hat{\lambda})T} - 1}{e^{2\lambda T} - 1}. \nonumber
\end{align}
Differentiating the above with respect to $\hat{\lambda}$, and using the moment identities \eqref{eq:moment-identities}, the closed-form expression for the gradient results in
\begin{align}
    \partial_{\hat{\lambda}} \bar{\ell}_x(\hat{\lambda}) &= \frac{2}{M_0(2\lambda)} \left( M_1(2\hat{\lambda}) - M_1(\lambda + \hat{\lambda}) \right), \label{eq:normalized-gradient-moment-form}\\
    &= \frac{\lambda}{\hat{\lambda}^2} \frac{(2\hat{\lambda}T - 1)e^{2\hat{\lambda}T} + 1}{e^{2\lambda T} - 1} - \frac{4\lambda}{(\lambda + \hat{\lambda})^2} \frac{((\lambda + \hat{\lambda})T - 1)e^{(\lambda + \hat{\lambda})T} + 1}{e^{2\lambda T} - 1}. \nonumber
\end{align}
Finally, applying again the moment identities \eqref{eq:moment-identities} with $h(\hat{\lambda}) = 2\hat{\lambda}$ and $h(\hat{\lambda}) = \lambda + \hat{\lambda}$, the Hessian of \eqref{eq:energy-normalized-trajectory-objective} with respect to $\hat{\lambda}$ is given by
\begin{align}
    \partial^2_{\hat{\lambda}} \bar{\ell}_x(\hat{\lambda}) &= \frac{2}{M_0(2\lambda)} \left( 2 M_2(2\hat{\lambda}) - M_2(\lambda + \hat{\lambda}) \right), \label{eq:normalized-hessian-moment-form}\\
    & = \frac{\lambda}{\hat{\lambda}^3} \frac{(4\hat{\lambda}^2T^2 - 4\hat{\lambda}T + 2)e^{2\hat{\lambda}T} - 2}{e^{2\lambda T} - 1} - \frac{4\lambda}{(\lambda + \hat{\lambda})^3} \frac{((\lambda + \hat{\lambda})^2T^2 - 2(\lambda + \hat{\lambda})T + 2)e^{(\lambda + \hat{\lambda})T} - 2}{e^{2\lambda T} - 1}. \nonumber
\end{align}

Given the closed-form expressions above, we characterize the landscape of $\bar{\ell}_x(\hat{\lambda})$ through two results reading against the requirements of gradient-based optimization, namely degeneracy of the gradient and of the Hessian. In particular, we show that the loss landscape exhibits a regime-dependent behavior, with the presence of vanishing or exploding gradients, alongside ill-conditioning, and how the issues are made worse, at fast rates, in the infinite-horizon limit $T \to +\infty$. Regarding the loss, it is immediate to verify that, for every fixed $T>0$, the minimizer is attained at $\hat{\lambda} = \lambda$, with $\bar{\ell}_x(\lambda) = 0$. By the asymptotic behavior of the moment function \eqref{eq:moment-tails}, we have that the loss flattens out for $\hat{\lambda} \to -\infty$, with $\bar{\ell}_x(\hat{\lambda}) \to 1$, while 
it inherits exponential growth for $\hat{\lambda} \to +\infty$, with $\bar{\ell}_x(\hat{\lambda}) \sim M_0(2\hat{\lambda})/M_0(2\lambda)$.
Given that the loss is smooth, the presence of the horizontal asymptote, having a positive value, alongside with the presence of the minimizer at which the loss vanishes, implies a non-convex landscape.
Considering the infinite-horizon limit $T \to +\infty$, the loss landscape exhibits a transition between the two regimes, with the following limiting behavior
\begin{equation}
    \lim_{T \to +\infty} \bar{\ell}_x(\hat{\lambda}) =
    \begin{cases}
        \dfrac{(\lambda - \hat{\lambda})^2}{\hat{\lambda}(\lambda + \hat{\lambda})}, & \lambda < 0, \ \hat{\lambda} < 0, \\[10pt]
        +\infty, & \lambda<0, \ \hat{\lambda}>0,
    \end{cases}
    \qquad
    \lim_{T \to +\infty} \bar{\ell}_x(\hat{\lambda}) =
    \begin{cases}
        1, & \lambda>0, \ \hat{\lambda}<\lambda, \\
        0, & \lambda>0, \ \hat{\lambda}=\lambda, \\
        +\infty, & \lambda>0, \ \hat{\lambda}>\lambda.
    \end{cases}
    \label{eq:infinite-horizon-landscape}
\end{equation}
Hence, the loss landscape converges to a smooth, non-convex profile in the dissipative regime $\lambda < 0$, while it converges to a discontinuous function in the unstable regime $\lambda > 0$, with a blow-up at $\hat{\lambda} > \lambda$ happening in both cases.
This is consistent with the intuition that learning in the unstable regime is a challenging task, and that increasing the time horizon may not be beneficial, with the collapse to a discontinuous landscape happening exponentially fast in $T$.

It is legitimate to expect that, the flattening of the loss landscape in the dissipative regime, and the blow-up in the unstable one, induce vanishing and exploding gradients, respectively. In particular, the exponential nature of moments, and their asymptotic behavior, suggests that the rates at which such degeneracies occur are exponential in $T$. 
We proceed to characterize the gradient of the loss landscape, being the main quantity of interest for gradient-based optimization. The following proposition characterizes the asymptotic behavior of the gradient, both in the tails $(\hat{\lambda} \to \pm\infty)$ and in the infinite-horizon limit $(T \to +\infty)$, highlighting the presence of vanishing or exploding gradients, depending on the regime of the target dynamics.

\begin{emphbox}
\begin{proposition}[Gradient behavior] 
\label{prop:gradient-behavior}        
Let $T>0$ and $\lambda \neq 0$ be fixed. The asymptotic behavior of the gradient of the energy-normalized trajectory objective \eqref{eq:energy-normalized-trajectory-objective} is given by
\begin{equation}
    \partial_{\hat{\lambda}} \bar{\ell}_x(\hat{\lambda}) \sim
    \begin{cases}
        -\dfrac{1}{M_0(2\lambda)}\dfrac{3}{2\hat{\lambda}^2}, & \hat{\lambda} \to -\infty,\\[15pt]
        \dfrac{1}{M_0(2\lambda)}\dfrac{T e^{2\hat{\lambda}T}}{\hat{\lambda}}, & \hat{\lambda} \to +\infty.
    \end{cases}
    \label{eq:gradient-tail-behavior}
\end{equation}
Considering the infinite-horizon limit, the asymptotic behavior of the gradient, in the unstable regime $\lambda > 0$, is given by
\begin{equation}
    \partial_{\hat{\lambda}} \bar{\ell}_x(\hat{\lambda}) \sim
    \begin{cases}
        \dfrac{1}{e^{2\lambda T}} \left ( \dfrac{\lambda}{\hat{\lambda}^2} - \dfrac{4\lambda}{(\lambda + \hat{\lambda})^2}\right ), &\hat{\lambda} < -\lambda,\\[15pt]
        -\dfrac{4\lambda T}{\lambda + \hat{\lambda}}e^{(\hat{\lambda} - \lambda) T}, &|\hat{\lambda}| < \lambda,\\[15pt]
        \dfrac{2\lambda T}{\hat{\lambda}}e^{2(\hat{\lambda} - \lambda) T}, &\hat{\lambda} > \lambda,\\
    \end{cases} \qquad T \to +\infty,
    \label{eq:gradient-infinite-horizon-unstable}
\end{equation}
while in the stable regime $\lambda < 0$, it holds that
\begin{equation}
    \lim_{T \to +\infty} \partial_{\hat{\lambda}} \bar{\ell}_x(\hat{\lambda}) = \frac{4\lambda}{(\lambda + \hat{\lambda})^2} - \frac{\lambda}{\hat{\lambda}^2}, \quad \hat{\lambda} < 0, \qquad \partial_{\hat{\lambda}} \bar{\ell}_x(\hat{\lambda})  \sim -\dfrac{2\lambda T}{\hat{\lambda}}e^{2\hat{\lambda} T}, \quad \hat{\lambda} > 0.
    \label{eq:gradient-infinite-horizon-stable}
\end{equation}
\end{proposition}
\end{emphbox}

\begin{proof}
The asymptotic behavior of the gradient in \eqref{eq:gradient-tail-behavior},
for $\hat{\lambda} \to \pm\infty$, is obtained by applying \eqref{eq:moment-tails} to the closed-form expression of the gradient \eqref{eq:normalized-gradient-moment-form}.
Considering the infinite-horizon behavior, in the unstable regime $\lambda > 0$ we distinguish between three cases:
(i) for $\hat{\lambda} < -\lambda$, both $M_1(2\hat{\lambda})$ and $M_1(\lambda + \hat{\lambda})$ converge to finite values
by \eqref{eq:moment-infinite-horizon}, while $M_0(2\lambda)$ at the denominator diverges exponentially fast in $T$, hence governing the asymptotic behavior of the gradient towards zero; (ii) for $|\hat{\lambda}| < \lambda$, both terms $M_1(2\hat{\lambda})/M_0(2\lambda)$ and $M_1(\lambda + \hat{\lambda})/M_0(2\lambda)$ converge exponentially fast to zero, with the second term being slower, hence dominating; (iii) for $\hat{\lambda} > \lambda$, both terms diverge exponentially fast in $T$, with the first term, $M_1(2\hat{\lambda})/M_0(2\lambda)$, dominating over the second, $M_1(\lambda + \hat{\lambda})/M_0(2\lambda)$, thus yielding \eqref{eq:gradient-infinite-horizon-unstable}.
In the stable regime $\lambda < 0$, it is enough to consider the following two cases: (i) when $\hat{\lambda}<0$ both terms in \eqref{eq:normalized-gradient-moment-form} converge to finite values, yielding the first limit in \eqref{eq:gradient-infinite-horizon-stable}; (ii) while, in the case $\hat{\lambda}>0$, the first term diverges exponentially fast in $T$, both in the case in which $\hat{\lambda} < -\lambda$, when the second term converges to a finite value, and in the case in which $\hat{\lambda} > -\lambda$, when the second term diverges exponentially fast in $T$, but at a slower rate than the first term, thus yielding the asymptotic in \eqref{eq:gradient-infinite-horizon-stable}.

\end{proof}

The results of Proposition \ref{prop:gradient-behavior} highlight the presence of vanishing or exploding gradients, depending on the regime of the target dynamics, and how fast such degeneracies occur in the infinite-horizon limit. The behavior of the gradient in the tails of the parameter space, reported in \eqref{eq:gradient-tail-behavior}, is consistent with flattening of the loss landscape for $\hat{\lambda} \to -\infty$, and with the blow-up of the loss landscape for $\hat{\lambda} \to +\infty$. In particular, the gradient vanishes polynomially fast for $\hat{\lambda} \to -\infty$, while it explodes exponentially fast for $\hat{\lambda} \to +\infty$, trivially inheriting the exponential growth of the loss landscape in that regime. In the infinite-horizon limit, the gradient presents a regime-dependent behavior, visualized in
Figure \ref{fig:trajectory-loss-landscape} over the $(\hat{\lambda}, \lambda)$-plane, and summarized as follows. In the unstable regime $\lambda > 0$, the gradient vanishes as $O(e^{-2\lambda T})$ for $\hat{\lambda} < -\lambda$, and as $O(Te^{(\hat{\lambda} - \lambda) T})$ for $|\hat{\lambda}| < \lambda$; while, for $\hat{\lambda} > \lambda$, it explodes at $O(Te^{2(\hat{\lambda} - \lambda) T})$-rate. Instead, in the stable regime $\lambda < 0$, the gradient converges to a smooth profile for $\hat{\lambda} < 0$, while it explodes at $O(Te^{2\hat{\lambda} T})$-rate for $\hat{\lambda} > 0$.

Now, we consider the Hessian of the loss landscape. In particular, being the loss not globally convex, the perspective of local convergence for gradient-based optimization is adopted, thus requiring a characterization of the Hessian at the minimizer $\hat{\lambda} = \lambda$.
Besides regularity of the loss, as minimum requirement for local convergence of gradient descent methods, it can be verified that the Hessian \eqref{eq:normalized-hessian-moment-form} is positive definite at the minimizer $\hat{\lambda} = \lambda$, with
\begin{equation}
    \partial^2_{\hat{\lambda}} \bar{\ell}_x(\lambda) = \frac{2 M_2(2\lambda)}{M_0(2\lambda)} \in (0, 2T^2).
    \label{eq:hessian-at-minimizer}
\end{equation}
The following proposition characterizes the behavior of the Hessian in the infinite-horizon limit, highlighting the presence of ill-conditioning in the unstable regime.

\begin{emphbox}
\begin{proposition}[Hessian behavior] 
\label{prop:hessian-behavior}    
Let $\lambda \neq 0$ be fixed. Considering the Hessian of the energy-normalized trajectory objective \eqref{eq:energy-normalized-trajectory-objective} at the minimizer $\hat{\lambda} = \lambda$, it holds that
\begin{equation*}
    \lim_{T \to +\infty} \partial^2_{\hat{\lambda}} \bar{\ell}_x(\lambda) =
    \begin{cases}
        \dfrac{1}{\lambda^2}, & \lambda < 0, \\
        +\infty, & \lambda > 0,
    \end{cases}
\end{equation*}
where the divergence in the unstable regime $\lambda > 0$ happens at a quadratic rate in $T$, with $\partial^2_{\hat{\lambda}} \bar{\ell}_x(\lambda) \sim 2T^2$.
\end{proposition}
\end{emphbox}

\begin{proof}
The behavior of the Hessian in the infinite-horizon limit is obtained by applying \eqref{eq:moment-infinite-horizon} to the Hessian at the minimizer \eqref{eq:hessian-at-minimizer}. The diverging behavior in the unstable regime $\lambda > 0$ at a quadratic rate in $T$ is a direct consequence of the asymptotic behavior of the second-order moment $M_2(2\lambda) \sim T^2 e^{2\lambda T}/(2\lambda)$.
\end{proof}

\paragraph{Implications for gradient-based optimization.}
We now turn to the perspective of local convergence of gradient descent, to study the practical implications of the above results on the optimization of the trajectory-based learning objective via first-order methods.
Proposition \ref{prop:hessian-behavior} hints at the increasing sharpness of the loss landscape at the minimizer, with the Hessian diverging at a quadratic rate in $T$, thus suggesting that the learning rate and the radius of guaranteed local convergence given by \citet[Theorem 1.2.4]{nesterov18} shrink with $T$. 
The following arguments formalize such an intuition, quantifying the rates at which such degeneracies develop.
Following \citet[Section 1.2]{nesterov18}, the gradient descent update for the energy-normalized trajectory objective \eqref{eq:energy-normalized-trajectory-objective} reads as
\begin{equation*}
    \hat{\lambda}^{(k+1)} = \hat{\lambda}^{(k)} - \eta_k \partial_{\hat{\lambda}} \bar{\ell}_x(\hat{\lambda}^{(k)}),
\end{equation*}
with $\eta_k$ the learning rate at iteration $k$. Under suitable regularity assumptions on the loss (e.g. $C^2$ with Lipschitz-continuous Hessian), local convergence of gradient descent is generally tied to the spectral properties of the Hessian at the minimizer. In particular, letting $r_k := |\hat{\lambda}^{(k)} - \lambda|$ be the distance of the $k$-th iterate from the minimizer, we have that requiring convergence of the gradient descent update yields the following inequality
\begin{equation*}
    0 < r_k < \bar{r} := \frac{2 \partial_{\hat{\lambda}}^2 \bar{\ell}_x(\lambda)}{\beta},
\end{equation*}
with $\beta$ the Lipschitz\footnote{Here, Lipschitz continuity is intended in a local sense, hence holding in a neighborhood of the minimizer, which is sufficient for the study of local convergence.} constant of the Hessian $\partial_{\hat{\lambda}}^2 \bar{\ell}_x(\hat{\lambda})$, and $\bar{r}$ the radius of the ball of attraction of the minimizer. By Proposition \ref{prop:hessian-behavior}, we have that $\partial_{\hat{\lambda}}^2 \bar{\ell}_x(\lambda) \sim 2T^2$ in the unstable regime, while $\beta$ can be bounded below by the value of the third derivative of the loss at the minimizer, leading to 
\begin{equation*}
    \bar{r} \leq \frac{4T^2}{\partial_{\hat{\lambda}}^3 \bar{\ell}_x(\lambda)} \sim \frac{4T^2}{6T^3} = \frac{2}{3T}, \quad T \to +\infty,
\end{equation*}
showing that the guaranteed radius shrinks polynomially in $T$. In particular, the asymptotic behavior of the third derivative of the loss at the minimizer can be derived by applying \eqref{eq:moment-infinite-horizon} to the closed-form expression of the third derivative, which reads as
\begin{equation*}
    \partial_{\hat{\lambda}}^3 \bar{\ell}_x(\lambda) = \frac{2}{M_0(2\lambda)} \left( 4 M_3(2\hat{\lambda}) - M_3(\lambda + \hat{\lambda}) \right)\bigg|_{\hat{\lambda} = \lambda} = \frac{6 M_3(2\lambda)}{M_0(2\lambda)} \sim 6T^3, \quad T \to +\infty.
\end{equation*}
Analogously, relying on the optimal choice of the learning rate, we have
\begin{equation*}
    \eta_k = \frac{1}{\partial_{\hat{\lambda}}^2 \bar{\ell}_x(\lambda)} \sim \frac{1}{2T^2}, \quad T \to +\infty, 
\end{equation*}
thus showing that the learning rate shrinks polynomially in $T$, at a quadratic rate, in the unstable regime.
The above considerations highlight the role of the time horizon on the smoothness of the loss landscape, as already observed by \cite{ribeiro20}, and its implications on the convergence of gradient-based optimization. The analysis, while limited to a one-dimensional linear system, sheds light on the possible challenges that may arise in the context of high-dimensional, nonlinear systems learning via trajectory-based objectives,  where the presence of unstable modes may dominate the loss landscape. In particular, the exponential rates at which gradients become uninformative, and the polynomial rates at which the learning rate and the basin of attraction shrink with $T$, suggest that casting the learning problem as a minimization of a trajectory-based objective to be addressed via gradient-based optimization may be ill-suited.

\subsection{Proof of Proposition \ref{prop:loss-degeneracy-discrete-time}, \ref{prop:jacobian-spectral-constraint}}
\label{app:loss-degeneracy-discrete-time}
\lossdegeneracydiscretetime*

\begin{remark}
For a Runge-Kutta method with tableau $(A,b)$, the stability function $R(\xi)$ is rational, with denominator $\det(I - \xi A)$ \citep[Proposition 3.1]{hairer91}, so in our case, $\det(I - \hat{\lambda}\dt A)$ is a polynomial in $\hat{\lambda}$ admitting finitely many roots, at which the recursion $\hat{x}_{k+1} = R(\hat{\lambda}\dt)\hat{x}_k$, and hence the loss \eqref{eq:rollout-loss}, is undefined. Since $|R(\xi)| \to +\infty$ as $\xi$ approaches such roots, each admits a punctured neighborhood where $|R(\hat{\lambda}\dt)| > 1$. Setting $\ell_{x,\infty}(\hat{\lambda};x_0)=+\infty$ at such points is therefore a continuous extension of the loss, under which the result of Proposition \ref{prop:loss-degeneracy-discrete-time} holds for every $\hat{\lambda} \in \C$.
\end{remark}

\begin{proof} Let us denote by $\ell_{x,\infty}(\hat{\lambda};x_0)$ the infinite-horizon limit of the rollout loss \eqref{eq:rollout-loss}, which, being the dynamics linear, can be expressed as
\begin{equation*}
    \ell_{x,\infty}(\hat{\lambda};x_0) = \lim_{\Nt \to +\infty} \frac{|x_0|^2}{\Nt} \sum_{k=1}^{\Nt} \underbrace{| e^{\lambda k \dt} - R(\hat{\lambda} \dt)^k|^2}_{a_k},
\end{equation*}
having used the fact that $x(t_k, x_0) = e^{\lambda k \dt} x_0$, alongside the identity $\hat{x}_k = R(\hat{\lambda} \dt)^k x_0$ for one-step Runge-Kutta methods. We denote partial sums by $S_{\Nt} = \sum_{k=1}^{\Nt} a_k$, and by $S_\infty$ the associated infinite series.
Hence, we proceed separately for the three different cases. 
\begin{enumerate}
    \item First, collapse of the loss for $|R(\hat{\lambda} \dt)| < 1$ is addressed. We notice that, $a_k$ can be expressed as
    \begin{equation}
        a_k = |e^{\lambda \dt}|^{2k} + |R(\hat{\lambda} \dt)|^{2k} - 2\Re\left (e^{\lambda k \dt} \overline{R(\hat{\lambda} \dt)^k}\right ),
        \label{eq:ak}
    \end{equation}
    so that we can consider the three geometric series associated to the different terms in the above expression, up to a shift in the index $k$. Since $|e^{\lambda \dt}| < 1$, $|R(\hat{\lambda} \dt)| < 1$, and $|e^{\lambda \dt} \overline{R(\hat{\lambda} \dt)}| < 1$, then the series converge absolutely, yielding
    \begin{equation*}
        S_\infty := \sum_{k=1}^{\infty} a_k = \frac{|e^{\lambda\dt}|^2}{1-|e^{\lambda\dt}|^2} + \frac{|R(\hat\lambda\dt)|^2}{1-|R(\hat\lambda\dt)|^2} - 2\Re\left(\frac{e^{\lambda\dt}\overline{R(\hat\lambda\dt)}}{1-e^{\lambda\dt}\overline{R(\hat\lambda\dt)}}\right) < +\infty.
    \end{equation*}
    Hence, being $a_k \geq 0$, it holds that the partial sums $S_{\Nt}$ of the series are bounded by $S_\infty$ for every $\Nt \in \N$, thus
    \begin{equation*}
        0 \leq \ell_{x,\infty}(\hat{\lambda};x_0) = \lim_{\Nt \to +\infty} \frac{|x_0|^2}{\Nt} S_{\Nt} \leq \lim_{\Nt \to +\infty} \frac{|x_0|^2}{\Nt} S_\infty = 0,
    \end{equation*}
yielding $\ell_{x,\Nt}(\hat{\lambda};x_0)= O(\Nt^{-1})$ as $\Nt \to +\infty$, for every fixed $\hat\lambda$.
\item When $|R(\hat{\lambda} \dt)| = 1$, by \eqref{eq:ak}, $a_k$ reads as
\begin{equation*}
    a_k = |e^{\lambda \dt}|^{2k} + 1 - 2\Re\left (e^{\lambda k \dt} \overline{R(\hat{\lambda} \dt)^k}\right ),
\end{equation*}
so that, being $|e^{\lambda \dt}| < 1$, and
\begin{equation*}
     \left|\Re \left (e^{\lambda k \dt} \overline{R(\hat{\lambda} \dt)^k}\right ) \right|\leq |e^{\lambda \dt}|^k \to 0, \quad \text{as} \ k \to +\infty,
\end{equation*}
then $a_k \to 1$, as $k \to +\infty$. Hence, by Cesàro's theorem, $\ell_{x,\infty}(\hat{\lambda};x_0) = |x_0|^2$ follows.

\item Finally, considering the case $|R(\hat{\lambda} \dt)| > 1$, the reverse triangle inequality yields the following lower bound on $a_k$
\begin{equation*}
 a_k \geq \big||e^{\lambda \dt}|^k - |R(\hat{\lambda} \dt)|^k \big|^2  =  |R(\hat{\lambda} \dt)|^{2k}\Bigg(\frac{|e^{\lambda \dt}|^k}{|R(\hat{\lambda} \dt)|^k} - 1\Bigg)^2.
\end{equation*}
Hence, by nonnegativity of $a_k$, and by the previous lower bound, it holds that
\begin{equation*}
\frac{1}{\Nt} S_{\Nt} \geq \frac{1}{\Nt}a_{\Nt} \geq \frac{1}{\Nt}|R(\hat{\lambda} \dt)|^{2\Nt}\Bigg(\frac{|e^{\lambda \dt}|^{\Nt}}{|R(\hat{\lambda} \dt)|^{\Nt}} - 1\Bigg)^2 \to +\infty, \quad \text{as} \ \Nt \to +\infty,
\end{equation*}
so, passing to the limit, it holds that $\ell_{x,\infty}(\hat{\lambda};x_0) = +\infty$, concluding the proof.
\end{enumerate}
\end{proof}

\jacobianspectralconstraint*

\begin{proof} We proceed in three steps, showing that $\bar{x}$ is an equilibrium for the learned dynamics; that $\Phi_\dt$, being the one-step map associated to the numerical integration of the learned dynamics, is a local diffeomorphism; finally, that the Jacobian of $\Phi_\dt$ at $\bar{x}$ satisfies the spectral constraint.
    \begin{enumerate}
        \item \textit{$\bar{x}$ is an equilibrium for the learned dynamics.} Denote by $a_k(x_0) := |x(t_k, x_0) - \hat{x}_k(x_0)|^2$. The series being convergent by \eqref{eq:bounded-rollout-loss}, and $a_k(x_0) \geq 0$, then it holds that $a_k(x_0) \to 0$ as $k \to +\infty$. By triangle inequality, it follows that
        \begin{equation*}
            |\hat{x}_k(x_0) - \bar{x}| \leq |x(t_k, x_0) - \bar{x}| + |x(t_k, x_0) - \hat{x}_k(x_0)|,
        \end{equation*}
        thus yielding $\hat{x}_k(x_0) \to \bar{x}$ as $k \to +\infty$, for every $x_0 \in B(\bar{x}, \delta)$. Hence, considering that $\hat{x}_{k+1}(x_0) = \Phi_\dt(\hat{x}_k(x_0))$, and passing to the limit $k \to +\infty$, we have that $\bar{x} = \bar{x} + \dt \hat{f}(\bar{x})$, by continuity of $\hat{f}$, concluding that: 
        (i) $\bar{x}$ is a fixed point of the one-step map $\Phi_\dt$, and (ii) $\hat{f}(\bar{x}) = 0$, that is, $\bar{x}$ is an equilibrium for the learned dynamics.
        
        \item \textit{$\Phi_\dt$ is a local diffeomorphism.} The Jacobian of $\Phi_\dt$ reads as $\partial_x \Phi_\dt(x) = I + \dt \partial_x \hat{f}(x)$. 
        With a slight abuse of notation, we denote by $|\cdot|$ also the operator norm induced by the Euclidean norm, so that $|\partial_x \hat{f}(x)| = \sup_{|v|=1} |\partial_x \hat{f}(x)v|$. With this notation, we have $|\partial_x \hat{f}(x)| \le \lip{\hat{f}}$ at every $x\in\R^n$, hence the assumption $\dt\,\lip{\hat{f}} < 1$ gives $|\dt \partial_x \hat{f}(x)| < 1$, implying that $\partial_x \Phi_\dt(x)$ is invertible by the Neumann series. By the inverse function theorem, $\Phi_\dt$ is then a local $C^1$ diffeomorphism at every $x \in \R^n$.

        \item \textit{Jacobian spectral constraint.} We proceed by contradiction, supposing that
        $\rho(I + \dt \partial_x \hat{f}(\bar{x})) > 1$, so that the unstable subspace of
        $\partial_x \Phi_\dt(\bar{x})$ is nontrivial. 
        Being $\Phi_\dt$ a local $C^1$ diffeomorphism, and $\bar{x}$ a fixed point of $\Phi_\dt$, the center-stable manifold theorem
        \citep[Theorem III.7]{shub87} applies, so that there exist a neighborhood $U$ of $\bar{x}$ and a $C^1$
        embedded manifold $\mathcal{M}^{\rm cs}$ with $\dim \mathcal{M}^{\rm cs} < n$, such that, if 
        $\{\Phi_\dt^{k}(y)\}_{k \ge 0}\subset U$, namely, the forward orbit starting from $y\in U$ remains in $U$, then
        $y \in \mathcal{M}^{\rm cs}$.
        Fix $x_0 \in B(\bar{x},\delta)$ and write $\hat{x}_k = \Phi_\dt^{k}(x_0)$. By step 1, we have that
        $\hat{x}_k \to \bar{x}$, so there is some time index $k_* = k_*(x_0) \in \N$ with $\hat{x}_k \in U$ for every
        $k \ge k_*$, namely, the forward orbit of $\hat{x}_{k_*}$ is such that $\{\hat{x}_{k_*+j}\}_{j \ge 0}\subset U$, hence $\hat{x}_{k_*} \in \mathcal{M}^{\rm cs}$, that is
        $x_0 \in \Phi_\dt^{-{k_*}}(\mathcal{M}^{\rm cs})$. Note that the initial condition $x_0$ itself need not belong
        to $\mathcal{M}^{\rm cs}$, its orbit being unconstrained before time $k_*$. Since $k_*$ depends on $x_0$, this
        yields
        \begin{equation*}
            B(\bar{x},\delta) \ \subseteq\ \bigcup_{k \in \N} \Phi_\dt^{-k}\big(\mathcal{M}^{\rm cs}\big).
        \end{equation*}
        Since $\mathcal{M}^{\rm cs}$ has positive codimension $< n$, it has null Lebesgue measure in $\R^n$, and being $\Phi_\dt$ a local diffeomorphism by step 2, each preimage
        $\Phi_\dt^{-k}(\mathcal{M}^{\rm cs})$ has null measure as well. So, the countable union of such sets has thus null Lebesgue measure, contradicting the fact that $B(\bar{x},\delta)$ has positive measure. Therefore
        $\rho(I + \dt \partial_x \hat{f}(\bar{x})) \le 1$, and since
        $\sigma(I + \dt \partial_x \hat{f}(\bar{x})) = \{1 + \dt\hat{\nu} : \hat{\nu} \in
        \sigma(\partial_x \hat{f}(\bar{x}))\}$ by spectral mapping, this reads as
        $\sigma(\partial_x \hat{f}(\bar{x})) \subseteq \overline{\mathcal{A}}_\dt$.
\end{enumerate}
\end{proof}

\begin{remark} 
While $\bar{x}$ is asymptotically stable for the target dynamics by assumption, and hence Lyapunov stable, the first step of the proof only establishes that it is an attracting fixed point of the one-step map $\Phi_\dt$. 
Lyapunov stability of the equilibrium $\bar{x}$ for $\Phi_\dt$ would yield the result at once by a linearization argument, but it would require a stronger assumption than \eqref{eq:bounded-rollout-loss}, namely that $\sum_{k\ge1}a_k(x_0)$ vanish as $x_0 \to \bar{x}$, rather than merely be finite for each $x_0$. 
Attraction only constrains the asymptotic behavior of the orbit $\hat{x}_k(x_0)$, which may leave $B(\bar{x},\delta)$ during the transient, before converging.
The center-stable manifold theorem circumvents this, being applied to the tail of each converging orbit. Then, local invertibility of the one-step map $\Phi_\dt$ ensures that, whenever $\rho(\partial_x\Phi_\dt(\bar{x})) > 1$, the set of initial conditions attracted to $\bar{x}$ has null measure, yielding a contradiction with the fact that it contains the ball $B(\bar{x},\delta)$, of positive measure.
\end{remark}
\section{Proofs of Section \ref{sec:velocity-learning}}
\label{app:velocity-learning}

\subsection{Proof of Proposition \ref{prop:velocity-regression-occupation-measure}, \ref{prop:variational-characterization}}
\label{app:velocity-regression}

\velocityregressionoccupationmeasure*

\begin{proof}
We first establish joint measurability of the flow map, then show that $\mu$ is a well-defined probability measure, and finally derive the equivalence between the two formulations of the objective $\Lf$. We denote by $\mathcal{B}(\X_0)$ and $\mathcal{B}([0,T])$ the Borel $\sigma$-algebras over $\X_0\subset \R^n$ and $[0,T]$, respectively, and by $\mathcal{B}([0,T])\otimes \mathcal{B}(\X_0)$ the product $\sigma$-algebra over $[0,T]\times \X_0$, with $\X_0\subset \R^n$ being the support of the initial distribution $\rho_0$,
as defined in Section \ref{sec:trajectory-learning}. We denote by $\leb{[0,T]}$ the Lebesgue measure over $[0,T]$. 
\begin{enumerate}
    \item \textit{Joint measurability.} Under the assumption of a globally well-defined flow $x(t,x_0)$, jointly continuous in $(t,x_0)\in[0,T]\times \X_0$, the mapping $(t,x_0)\mapsto x(t,x_0)$ is $\mathcal{B}([0,T])\otimes\mathcal{B}(\X_0)$-measurable. In addition, for any Borel measurable $g:\R^n\to \R_+$, the composition $(t,x_0)\mapsto g(x(t,x_0))$ is jointly measurable as well, as a composition of measurable maps.
    \item \textit{Definition of the occupation measure.} Let us denote by $\rho_t := x(t,\cdot)_\#\rho_0$ the push-forward of the initial distribution $\rho_0$ under the flow map $x(t,\cdot)$, for $t\in[0,T]$, reading as
    \begin{equation*}
        \rho_t(A) = \rho_0(x(t,\cdot)^{-1}(A)) = \int_{\R^n} 1_A(x(t,x_0)) \rho_0(dx_0),
    \end{equation*}
    with $x(t,\cdot)^{-1}(A) := \{x_0\in\X_0: x(t,x_0)\in A\}$ denoting the preimage of $A$ under the flow map.
    Since the mapping $(t,x_0)\mapsto 1_A(x(t,x_0))$ is jointly measurable by the previous step, by Tonelli's theorem \citep[Theorem 1.7.15]{tao11}, the map $t\mapsto \rho_t(A)$ is measurable for any Borel set $A\subset \R^n$.
    Let $\pi$ denote the product measure $\frac{1}{T}\leb{[0,T]} \otimes \rho_0$. For any Borel set $A\subset \R^n$, by the definition of push-forward it holds that
     \begin{align}
        (x(\cdot,\cdot)_\#\pi)(A) &= \pi(x(\cdot, \cdot)^{-1}(A)), \nonumber\\
        &= \int_{[0,T]\times \R^n} 1_{x(\cdot, \cdot)^{-1}(A)}(t,x_0) \, \pi(dt, dx_0), \nonumber\\
        &= \frac{1}{T}\int_0^T \int_{\R^n} 1_A(x(t,x_0)) \rho_0(dx_0) dt, \nonumber\\
        &= \frac{1}{T}\int_0^T \rho_t(A) dt  = \mu(A),
        \label{eq:push-forward-occupation-measure}
     \end{align}
    where the third step follows from the definition of the product measure $\pi$ and by Tonelli's theorem, having denoted by $x(\cdot, \cdot)^{-1}(A) := \{(t,x_0)\in[0,T]\times \X_0: x(t,x_0)\in A\}$.
    Thus, $\mu$ is a probability measure on $\R^n$, as the push-forward of the probability measure $\pi$ under the flow map $x(\cdot,\cdot)$.
     \item \textit{Equivalence of formulations.} Considering the definition of $\Lf$, the equivalence follows from
    \begin{align*}
        \Lf[\hat{f}]
        &:= \frac{1}{T} \int_{\R^n} \int_0^T |f(x(t,x_0)) - \hat{f}(x(t,x_0))|^2 dt \, \rho_0(dx_0),\\
        &=  \int_{[0,T]\times\R^n} |f(x(t,x_0)) - \hat{f}(x(t,x_0))|^2 \pi(dt, dx_0),\\
        &=  \int_{\R^n} |f(x) - \hat{f}(x)|^2 (x(\cdot,\cdot)_\#\pi)(dx),\\
        &\overset{\eqref{eq:push-forward-occupation-measure}}{=} \int_{\R^n} |f(x) - \hat{f}(x)|^2 \mu(dx),
        \end{align*}
where the second step follows from Tonelli's theorem, by measurability and non-negativity of the map $(t,x_0)\mapsto |f(x(t,x_0)) - \hat{f}(x(t,x_0))|^2$; the third step follows from the change of variables formula for push-forward measures \citep[Proposition 1.7]{ambrosio21}.
\end{enumerate}
\end{proof}

\variationalcharacterization*

\begin{proof}
Under the standing assumptions, $\operatorname{supp}\mu\subseteq \X$ is compact as the continuous image of a compact set. Being $f\in W^{1,\infty}_{\rm loc}(\R^n; \R^n)$, $f$ is continuous and hence bounded on $\X$, so that $f\in L^2(\mu;\R^n)$ and $\Lf[\hat f]<\infty$ for every $\hat f\in L^2(\mu;\R^n)$. By Proposition~\ref{prop:velocity-regression-occupation-measure},
\begin{equation*}
    \Lf[\hat f] = \int_{\R^n}|f(x)-\hat f(x)|^2\,\mu(dx) = \|f-\hat f\|^2_{L^2(\mu)}.
\end{equation*} 
Since the integrand is nonnegative, $\Lf[\hat f]=0$ if and only if $|f-\hat f|=0$ $\mu$-a.e., namely, $\hat f=f$ $\mu$-a.e. In particular, since $\Lf$ is convex, non-negative, with $\Lf[f]=0$, then $f$ is the unique minimizer of $\Lf$ over $L^2(\mu;\R^n)$, i.e., every minimizer coincides with $f$ $\mu$-a.e.
\end{proof}

\subsection{Proof of Lemma \ref{lemma:continuous-time-stability-estimates}, \ref{lemma:discrete-time-stability-estimate}}
\label{app:stability-estimates}

\continuoustimestabilityestimates*

\begin{proof}
We proceed by first identifying a compact set within which the solutions lie, and then applying Gronwall's lemma to obtain the desired estimates.
\begin{enumerate}
\item \textit{Confinement.} By assumption, both flow maps are defined on $[0,T] \times \X_0$. Since $f$ and $\hat{f}$ are locally Lipschitz in the second argument, the flow maps $(t,x_0)\mapsto x(t,x_0)$ and $(t,x_0)\mapsto\hat{x}(t,x_0)$ are jointly continuous there \citep[Theorem~2.9]{teschl12}. As continuous images of the compact set $[0,T]\times\X_0$, the reachable sets $\X := x([0,T]\times\X_0)$ and $\widehat{\X} := \hat{x}([0,T]\times\X_0)$ are compact, and so is $K:=\X\cup\widehat{\X}$. Finally, each $\hat{f}(t,\cdot) \in W^{1,\infty}_{\rm loc}(\R^n;\R^n)$ is Lipschitz on the compact $K$, by local Lipschitz continuity, and the map $t\mapsto \lip{\hat{f}(t,\cdot)|_{K}}$ is bounded on $[0,T]$ by continuity, so that $\Lambda < +\infty$.

\item \textit{Gronwall's bound.}
Fix $x_0\in\X_0$ and abbreviate $x(t) := x(t,x_0)$, $\hat{x}(t) := \hat{x}(t,x_0)$. Since
$x(0)=\hat{x}(0)$, by triangle inequality we have
\begin{align}
    |x(t) - \hat{x}(t)|
    &\leq \int_0^t \left ( |f(s,x(s)) - \hat{f}(s,x(s))| + |\hat{f}(s,x(s)) - \hat{f}(s,\hat{x}(s))| \right ) ds, \nonumber\\
    &\leq \int_0^t \left ( |f(s,x(s)) - \hat{f}(s,x(s))| + \Lambda |x(s) - \hat{x}(s)| \right ) ds, \nonumber\\
    &= \alpha(t) + \Lambda \int_0^t |x(s) - \hat{x}(s)| ds,
\end{align}
where $\alpha(t) := \int_0^t\xi(s)ds$, and $\xi(s) := |f(s,x(s)) - \hat{f}(s,x(s))|$; the latter inequality follows from the fact that $x(t), \hat{x}(t)\in K$ for all $t\in[0,T]$, by the previous step, with $\Lambda = \sup_{t\in[0,T]}\lip{\hat{f}(t,\cdot)|_{K}}$. Hence, by Gronwall's lemma \citep[Lemma~2.7]{teschl12} we have
\begin{align}
    |x(t) - \hat{x}(t)| &\leq \alpha(t) + \Lambda \int_0^t e^{\Lambda (t-s)} \alpha(s) ds, \nonumber\\
    &= \alpha(t) + \left [ - e^{\Lambda (t-s)} \alpha(s)\right]_0^t + \int_0^t e^{\Lambda (t-s)} \xi(s) ds, \nonumber\\
    &= \int_0^t e^{\Lambda (t-s)} \xi(s) ds, \label{eq:gronwall-bound}
\end{align}
where the first equality follows from integration by parts.

\item \textit{Estimates.} The estimates follow by applying H\"older's inequality to \eqref{eq:gronwall-bound}, noting that the integrand is nonnegative. In particular, for the $L^\infty$-$L^\infty$ estimate, we have
\begin{equation*}
    \int_0^t e^{\Lambda (t-s)} \xi(s) ds \leq  \left ( \int_0^t e^{\Lambda (t-s)} ds \right ) \|\xi\|_{L^\infty([0,T])} = \frac{e^{\Lambda t} - 1}{\Lambda} \|\xi\|_{L^\infty([0,T])},
\end{equation*}
from which the final estimate follows by taking the supremum over $t\in[0,T]$, hence holding with $C = (e^{\Lambda T} - 1)/\Lambda$.
For the $L^\infty$-$L^2$ and $L^2$-$L^2$ estimates, again by H\"older's inequality we have
\begin{equation}
    \int_0^t e^{\Lambda (t-s)} \xi(s) ds \leq \left( \int_0^t e^{2\Lambda (t-s)} ds \right)^{\frac12} \left( \int_0^t \xi(s)^2 ds \right)^{\frac12} = \left( \frac{e^{2\Lambda t} - 1}{2\Lambda} \right)^{\frac12} \|\xi\|_{L^2([0,t])}. 
    \label{eq:holder-l2}
\end{equation}
Hence, the $L^\infty$-$L^2$ estimate follows by taking the supremum over $t\in[0,T]$, so that it holds with $C = \sqrt{(e^{2\Lambda T} - 1)/(2\Lambda)}$. While, for the $L^2$-$L^2$ estimate, we take the square of both sides and then integrate over $t\in[0,T]$, to obtain
\begin{equation*}
    \int_0^T |x(t) - \hat{x}(t)|^2 dt \leq \left( \int_0^T  \frac{e^{2\Lambda t} - 1}{2\Lambda} dt \right)\|\xi\|_{L^2([0,T])}^2,
\end{equation*}
from which the final estimate follows with $C = \sqrt{e^{2\Lambda T} - 2\Lambda T - 1}/(2\Lambda)$.
\end{enumerate}
\end{proof}

\discretetimestabilityestimate*

\begin{proof}
We proceed by first identifying a compact set enclosing the solutions, and then applying discrete Gronwall's lemma to obtain the desired estimate.
\begin{enumerate}
   \item \textit{Confinement.} By similar arguments as in the proof of Lemma~\ref{lemma:continuous-time-stability-estimates}, the continuous-time flow map $x(t,x_0)$ is jointly continuous, yielding a compact reachable set $\X := x([0,T]\times\X_0)$. Considering the numerical solution of the learned system $\hat{x}_k$, it is obtained by iteratively applying the recursive relation $\hat{x}_{k+1} = \hat{x}_k + \dt F(t_k, \hat{x}_k, \dt; \hat{f})$, with $\hat{x}_0 = x_0$. For RK methods, the increment function $F$ is a linear combination of evaluations of $\hat{f}$ at the RK nodes, hence continuous in the second argument, so that the mapping $x_0\mapsto \hat{x}_k$ is continuous for any $k=0,\ldots, \Nt$. Hence, the set $\widehat{\X} := \{\hat{x}_k(x_0): k=0,\ldots, \Nt, x_0\in\X_0\}$ is compact as the continuous image of a compact set. We take $R>0$ with $\X\cup\widehat{\X}\subset B_R(0)$ and set $K := \overline{B_R(0)}$, compact and convex.
   
   \item \textit{Discrete Gronwall's bound.}
   We proceed by first decomposing the error between the true trajectory and the numerical approximation of the learned one at time $t_k$ into two terms, relying on the intermediate one-step solution $\hat{x}_k'$, defined as
   \begin{equation*}
    \hat{x}_{k+1}' = x(t_k) + \dt \, F(t_k, x(t_k), \dt; \hat{f}), \quad k=0,\ldots, \Nt-1,
   \end{equation*}
   where the dependence on the initial condition $x_0$ is omitted for ease of notation. 
   Hence, denoting by $\varepsilon_{k+1}:=x(t_{k+1}) - \hat{x}_{k+1}'$ the residual defined in \eqref{eq:tf-residual}, we have
    \begin{align*}
    x(t_{k+1}) - \hat{x}_{k+1} &= x(t_{k+1}) - \hat{x}_{k+1}' + \hat{x}_{k+1}' - \hat{x}_{k+1}\\
    &= \varepsilon_{k+1} + x(t_k) + \dt F(t_k, x(t_k), \dt; \hat{f}) - \hat{x}_k - \dt F(t_k, \hat{x}_k, \dt; \hat{f}),\\
    &= \varepsilon_{k+1} + \dt \big( F(t_k, x(t_k), \dt; \hat{f}) - F(t_k, \hat{x}_k, \dt; \hat{f}) \big) + x(t_k) - \hat{x}_k,
    \end{align*}
    from which, by recursively applying the above relation, it follows that
    \begin{equation*}
    x(t_{k+1}) - \hat{x}_{k+1} = \sum_{j=1}^{k+1} \varepsilon_j + \dt \sum_{j=1}^{k} \big( F(t_j, x(t_j), \dt; \hat{f}) - F(t_j, \hat{x}_j, \dt; \hat{f}) \big),
    \end{equation*}
    noticing that $x(t_0) = \hat{x}_0$. Taking the norm of both sides and applying the triangle inequality, we obtain
    \begin{align*}
    |x(t_{k+1}) - \hat{x}_{k+1}| &\leq \sum_{j=1}^{k+1} |\varepsilon_{j}| + \dt \sum_{j=1}^{k} \big| F(t_j, x(t_j), \dt; \hat{f}) - F(t_j, \hat{x}_j, \dt; \hat{f}) \big|,\\
    &\leq \sum_{j=1}^{k+1} |\varepsilon_{j}| + \dt \, \Lambda \sum_{j=1}^{k} |x(t_j) - \hat{x}_j|,
    \end{align*}
    where the last inequality follows from the fact that $x(t_j), \hat{x}_j\in K$ for all $j=1,\ldots, k$, by the previous step, with $\Lambda = \sup_{t\in[0,T]}\lip{F(t,\cdot,\dt; \hat{f})|_{K}}$. Hence, applying discrete Gronwall's lemma \citep[Lemma~11.2]{quarteroni07}, we obtain
    \begin{equation}
        |x(t_{k+1}) - \hat{x}_{k+1}| \leq e^{k \dt \Lambda } \sum_{j=1}^{k+1} |\varepsilon_{j}|,
        \label{eq:discrete-gronwall-bound}
    \end{equation}
    from which, by taking the supremum over $k=0,\ldots, \Nt-1$, the final estimate follows.
   \end{enumerate}
\end{proof}

\subsection{Proof of Corollary \ref{cor:theoretical-objective-control}, \ref{cor:wasserstein-stability-estimate}, \ref{cor:empirical-objective-control}}
\label{app:objective-control}

\theoreticalobjectivecontrol*
\begin{proof} By the $L^2$-$L^2$ stability estimate in Lemma~\ref{lemma:continuous-time-stability-estimates}, it holds that
\begin{align*}
        \lx(\hat{f};x_0) &= \frac{1}{T}\|x(\cdot,x_0) - \hat{x}(\cdot,x_0)\|^2_{L^2([0,T];\R^n)}, \\
        &\leq C_{2,2}^2(\Lambda) \frac{1}{T} \|f(\cdot, x(\cdot, x_0)) - \hat{f}(\cdot, x(\cdot, x_0))\|^2_{L^2([0,T];\R^n)}, \\
        &= C_{2,2}^2(\Lambda) \lf(\hat{f};x_0),
\end{align*}
for any initial condition $x_0\in\X_0$. Taking the expectation with respect to $x_0\sim \rho_0$ on both sides, the result follows.
\end{proof}

For the sake of completeness, we recall the definition of $p$-Wasserstein distance, following \citet[Chapter 6]{villani09}.

\begin{definition} Let $(X,d)$ be a metric space, and $p\in[1,+\infty)$. For any two probability measures $\mu,\nu$ on $X$, the Wasserstein distance of order $p$ between $\mu$ and $\nu$ is defined as
\begin{equation}
    \mathcal{W}_p^p(\mu,\nu) := \inf_{\pi\in\Pi(\mu,\nu)} \int_{X\times X} d(x,y)^p \pi(dx,dy),
\end{equation}
with $\Pi$ denoting the set of all joint probability measures on $X\times X$ whose marginals are $\mu$ and $\nu$.
\end{definition}

\wassersteinstabilityestimate*
\begin{proof}
The bound follows by leveraging the diagonal coupling between $\rho_t$ and $\hat{\rho}_t$, defined as the push-forward of $\rho_0$ under the flow maps of the true and learned systems, namely $(x(t,\cdot), \hat{x}(t,\cdot))_\#\rho_0$. Thus, denoting by $\Pi(\rho_t, \hat{\rho}_t)$ the set of all couplings between $\rho_t$ and $\hat{\rho}_t$, we have
\begin{align*}
    \mathcal{W}_2^2(\rho_t, \hat{\rho}_t) &= \inf_{\pi \in \Pi(\rho_t, \hat{\rho}_t)} \int_{\R^n\times\R^n} |x - x'|^2 \, \pi(dx,dx'), \\
    &\leq \int_{\R^n\times \R^n} |x - x'|^2 \, (x(t,\cdot), \hat{x}(t,\cdot))_\#\rho_0(dx,dx'),\\
    &=\int_{\R^n} |x(t,x_0) - \hat{x}(t,x_0)|^2 \, \rho_0(dx_0),\\
    &\overset{\eqref{eq:holder-l2}}{\leq} \frac{e^{2\Lambda T} - 1}{2\Lambda}  \int_{\R^n} \|f(\cdot, x(\cdot, x_0)) - \hat{f}(\cdot, x(\cdot, x_0))\|_{L^2([0,T];\R^n)}^2 \, \rho_0(dx_0),\\
    &= T C_{\infty,2}^2(\Lambda) \, \E_{x_0\sim \rho_0}\lf(\hat{f};x_0),
\end{align*}
having applied the pointwise bound \eqref{eq:holder-l2} from the proof of Lemma \ref{lemma:continuous-time-stability-estimates} in the last inequality.
\end{proof}

\empiricalobjectivecontrol*

\begin{proof}
    Applying Cauchy-Schwarz to \eqref{eq:discrete-gronwall-bound}, it follows that
    \begin{align*}
        |x(t_k,x_0) - \hat{x}_k(x_0)|^2 \leq e^{2T\Lambda} \Nt \sum_{j=1}^{\Nt}|\varepsilon_j(x_0)|^2, 
    \end{align*}
    uniformly for any $k=0,\ldots, \Nt$. Hence, averaging over $k$, we obtain
    \begin{align*}
       \ell_{x,\Nt}(\hat{f};x_0) 
       &= \frac{1}{\Nt}\sum_{k=1}^{\Nt}|x(t_k,x_0) - \hat{x}_k(x_0)|^2, \\ 
       &\leq e^{2T\Lambda} \Nt \sum_{k=1}^{\Nt}|\varepsilon_k(x_0)|^2, \\
       &= e^{2T\Lambda} N_t^2 \, \ell_{x,\Nt,1}(\hat{f};x_0).
    \end{align*}
    Averaging over $x_0\sim \rho_0$ on both sides, the result follows.
\end{proof}

\subsection{Proof of Theorem \ref{thm:universal-approximation-flow}, \ref{thm:continuous-time-approximation}}
\label{app:universal-continuous-time-approximation}

We first provide a result establishing that the Gaussian error linear unit (GELU) activation function is an exponential $(1,1)$ partition of unity (PU), defined in \citet[Section 4.1]{guhring21} and required by \citet[Theorem 4.9]{guhring21} to establish approximation rates in Theorem \ref{thm:universal-approximation-flow}.

\begin{lemma}
\label{lemma:gelu-exponential-pu}
The GELU activation function $\varrho(x) = x \Phi(x)$, with $\Phi$ being the standard Gaussian cumulative distribution function, is an exponential $(1,1)$-PU-admissible in the sense of \citet[Definition 4.2]{guhring21}, with $A=0$, $B=1$, $C=\sqrt{2/\pi}$, $D=1/2$ and $R=1$.
\end{lemma}

\begin{proof} The proof follows by direct verification of the conditions (a)-(d) in \citet[Definition 4.2]{guhring21}. Denote by $\phi=\Phi'$ the standard Gaussian probability density function. Then note that
    \begin{equation}
    \varrho'(x) = \Phi(x) + x\phi(x), \qquad \varrho''(x) = (2-x^2)\phi(x).
    \end{equation}
    We proceed to verify the conditions (a)-(d) in \citet[Definition 4.2]{guhring21}.
    \begin{enumerate}[(a)]
        \item $\varrho$ is Lipschitz continuous, noting that $\varrho \in C^\infty(\R)$ and $\varrho'(x) \to 1$ as $x\to +\infty$ and $\varrho'(x) \to 0$ as $x\to -\infty$, and that $\varrho''$ vanishes only at $x = \pm \sqrt{2}$, so that $\varrho'$ attains its maximum at $x = \sqrt{2}$, with $\varrho'(\sqrt{2}) = \Phi(\sqrt{2}) + \sqrt{2}\phi(\sqrt{2})$, so that $\|\varrho'\|_{L^\infty(\R)} < \infty$.
        \item $\phi, \Phi \in C^\infty(\R)$, so $\varrho \in C^\infty(\R)\subset C^1(\R\setminus[-R,R])$, for every $R>0$.
        \item $\varrho'\in W^{0,\infty}(\R) = L^\infty(\R)$, by (a).
        \item Take $A=0$, $B=1$. Then, recalling that $1-\Phi(x) \leq \phi(x)/x$ for $x>0$, we have
        \begin{equation}
            |1-\varrho'(x)| = |1 - \Phi(x) - x\phi(x)| \leq (1-\Phi(x)) + x\phi(x) \leq \left (x + \frac{1}{x}\right )\phi(x),
            \label{eq:d1}
        \end{equation} 
        for $x>0$, while, by symmetry of the Gaussian distribution, for $x<0$ we have
        \begin{align}
            |\varrho'(x)| &= |\Phi(x) + x\phi(x)|  = |1 - \Phi(-x) + x\phi(x)|, \nonumber \\  &\leq (1-\Phi(-x)) + |x|\phi(x) \leq \left (|x| + \frac{1}{|x|}\right )\phi(x). \label{eq:d2}
        \end{align} 
        Both \eqref{eq:d1} and \eqref{eq:d2} yield the same bound, namely $g(x) := (|x| + 1/|x|)\phi(x)$ for $x\neq 0$. In particular, for $x>0$, we have that $g'(x) = -(x^2 + 1/x^2)\phi(x) < 0$, so that $g$ is strictly decreasing for $x>0$, while it is strictly increasing for $x<0$, by symmetry. Hence, setting $h(x) = g(x) e^{\frac{x}{2}}$, we have that $h$ is strictly decreasing for $x>0$, so that
        \begin{equation*}
            |1-\varrho'(x)| \leq g(x) = h(x) e^{-\frac{x}{2}} \leq h(1) e^{-\frac{x}{2}} = \sqrt{\frac{2}{\pi}} e^{-\frac{x}{2}}, \quad x>1.
        \end{equation*}
        Similarly, setting $h(x) = g(x) e^{-\frac{x}{2}}$, we have that $h$ is strictly increasing for $x<0$, so that
        \begin{equation*}
            |\varrho'(x)| \leq g(x) = h(x) e^{\frac{x}{2}} \leq h(-1) e^{\frac{x}{2}} = \sqrt{\frac{2}{\pi}} e^{\frac{x}{2}}, \quad x<-1.
        \end{equation*}
        Hence, exponential decay conditions (d.1) and (d.2) are satisfied by taking $C=\sqrt{2/\pi}$, $D=1/2$ and $R=1$, while (d.3) holds vacuously.
    \end{enumerate}
\end{proof}
\universalapproximationflow*

\begin{proof} The proof develops in four steps.
\begin{enumerate}
    \item \textit{Approximation of the velocity field and Lipschitz bound.} By Assumption \ref{ass:vl-regularity-1}, $f\in W^{2,\infty}([0,T]\times K; \R^n)$. Since $K$ is compact, a box $Q\subset \R^{n+1}$ can be found such that $[0,T]\times K\subseteq Q$. Composing with an affine bijection $\psi$ mapping $Q$ onto the unit cube, we may take $[0,1]^{n+1}$ as the domain of interest. Moreover, the composition $f\circ \psi^{-1}$ can be rescaled by the resulting Sobolev norm $\|f\circ \psi^{-1}\|_{W^{2,\infty}([0,1]^{n+1}; \R^n)}$, so that each component lies in the unit ball of $W^{2,\infty}([0,1]^{n+1})$. The affine reparameterization and rescaling can be respectively absorbed into the first and last layer of the neural network architecture, so that the resulting approximant belongs to the same class $\mathcal{N}_\varrho(\R^{n+1}; \R^n)$. Hence, given an accuracy $\tilde{\epsilon} \in (0,\frac12]$, applying \citet[Theorem 4.9]{guhring21} componentwise with accuracy $\tilde{\epsilon} / C_\psi$, where $C_\psi = \|f\circ \psi^{-1}\|_{W^{2,\infty}([0,1]^{n+1}; \R^n)}\max\{1,|D\psi|\}$, 
    yields a deep neural network $\hat{f}~\in~\mathcal{N}_\varrho(\R^{n+1}; \R^n)$ such that
    \begin{equation}
        \|f - \hat{f}\|_{W^{1,\infty}([0,T]\times K; \R^n)} \leq \tilde{\epsilon}.
        \label{eq:velocity-approximation}
    \end{equation}
    Such a reduction to the unit cube, and the subsequent rescaling, only affects constants in the approximation bound, but not rates. Then, by the definition of the $W^{1,\infty}$-norm and convexity of $K$, the bound above implies that
    \begin{align*}
        \tilde{\epsilon}\geq\|f - \hat{f}\|_{W^{1,\infty}([0,T]\times K; \R^n)} 
        &\geq \|\partial_x f - \partial_x \hat{f}\|_{L^\infty([0,T]\times K; \R^{n\times n})}, \\ 
        &\geq |\|\partial_x f\|_{L^\infty([0,T]\times K; \R^{n\times n})} - \|\partial_x \hat{f}\|_{L^\infty([0,T]\times K; \R^{n\times n})}|,\\ 
        &= |\lip{f|_K} - \lip{\hat{f}|_K}|,
    \end{align*}
    from which it follows that $\lip{\hat{f}|_K} \leq M+\frac12$, since $\lip{f|_K} \leq M$ by Assumption \ref{ass:vl-regularity-1}.
    Considering the $L^\infty$-$L^\infty$ stability estimate of Lemma \ref{lemma:continuous-time-stability-estimates}, by monotonicity of the constant $C_{\infty,\infty}(\Lambda)$ in $\Lambda$, then this implies that
    $C_{\infty,\infty}(\lip{\hat{f}|_K}) \leq C_{\infty,\infty}(M+\frac12)$.

    \item \textit{Existence and confinement of the learned flow.} Write $C:=C_{\infty,\infty}(M+\frac12)$. Taking $\tilde{\epsilon} < \delta / C$, 
    we claim that $\hat{x}(t,x_0)$ exists and remains in $K$ for all $t\in[0,T]$ and $x_0 \in \X_0$. 
    Since $\hat{f}$ is locally Lipschitz in the second argument, the flow map $\hat{x}(t,x_0)$ is well-defined on a maximal interval $[0,T_{\max})$, with $T_{\max}>0$ \citep[Theorem~2.13]{teschl12}. 
    Fix $x_0 \in \X_0$, denote $I:=[0,\min\{T,T_{\max}\})$, and set
    \begin{equation*}
        \xi(t):=C\|f(\cdot,x(\cdot,x_0)) - \hat{f}(\cdot, x(\cdot,x_0))\|_{L^\infty([0,t];\R^n)}, \qquad t \in [0,T],
    \end{equation*}
    which is continuous and non-decreasing in $t$, with $\xi(0) = 0$, and $\xi(T)<\delta$ under the choice of $\tilde{\epsilon}$,
    and is defined on all of $[0,T]$ since it depends only on the true trajectory $x(t,x_0)$, which is well-defined over $[0,T]$. 
    We shall proceed by contradiction, assuming that the solution leaves $K$ at some time, namely that there exists a first exit time $t^* \in (0,\min\{T,T_{\max}\})$ such that $\hat{x}(t^*,x_0) \in \partial K$, which is well defined since $t^*<T_{\max}$.
     Therefore, the distance from the convex hull of the reachable set $\conv(\X)$ is exactly $\delta$, i.e. $\dist(\hat{x}(t^*,x_0),\conv(\X)) = \delta$, and since $\X \subseteq \conv(\X)$, then
    \begin{align}
        \delta = \dist(\hat{x}(t^*,x_0),\conv(\X)) &\leq \dist(\hat{x}(t^*,x_0),\X), \nonumber \\ 
        &= \inf_{y \in \X}|\hat{x}(t^*,x_0) - y|, \nonumber\\
        &\leq |\hat{x}(t^*,x_0) - x(t^*,x_0)| \label{eq:dist-bound-1}.
    \end{align}
    On $[0,t^*]$ both trajectories lie in $K$, with $K$ convex, so that the $L^\infty$-$L^\infty$ stability estimate in Lemma~\ref{lemma:continuous-time-stability-estimates} applies on $[0,t^*]$ with $\Lambda \leq M +\frac12$, yielding
    \begin{equation*}
        |\hat{x}(t^*,x_0) - x(t^*,x_0)|\leq \xi(t^*) \leq \xi(T) < \delta,
    \end{equation*}
    by monotonicity of $\xi$, contradicting \eqref{eq:dist-bound-1}. Therefore, $\hat{x}(t,x_0) \in K$ for all $t \in I$. Now, if $T_{\max} \leq T$ then the solution must leave $K$ as $t\to T_{\max}^-$ \citep[Corollary~2.16]{teschl12}, contradicting the fact that $\hat{x}(t,x_0) \in K$ for all $t \in I$. Hence, $T_{\max} > T$, and the solution is defined on $t\in[0,T]$  with values in $K$, for all $x_0 \in \X_0$.

    \item \textit{Approximation of the flow map.} Given the confinement established in the previous step, the $L^\infty$-$L^\infty$ stability estimate of Lemma \ref{lemma:continuous-time-stability-estimates} applies, yielding
    \begin{align*}
        \|x - \hat{x}\|_{L^\infty([0,T]\times \X_0)}
        &\leq C_{\infty,\infty}(\Lambda) \sup_{x_0\in\X_0} \|f(\cdot, x(\cdot, x_0)) - \hat{f}(\cdot, x(\cdot, x_0))\|_{L^\infty([0,T];\R^n)},\\
        &\leq  C\|f - \hat{f}\|_{W^{1,\infty}([0,T]\times K;\R^n)},\\
        &\overset{\eqref{eq:velocity-approximation}}{\leq} C\tilde{\epsilon}.
    \end{align*}
    Without loss of generality we can assume that $\epsilon < \delta$. Setting $\tilde{\epsilon} = \min\left \{ \frac12, \frac{\epsilon}{C} \right \}$, then gives $C\tilde{\epsilon} \leq \epsilon < \delta$, so that the smallness condition in Step 2 is satisfied, and the final bound follows. 
    \item Taking $\varrho$ to be the GELU activation function, Lemma \ref{lemma:gelu-exponential-pu} establishes that it is an exponential $(1,1)$-PU. Moreover, it is three times continuously differentiable in a neighborhood of $x=0$, being $C^\infty(\R)$, and $\varrho''(0) = \sqrt{2/\pi} \neq 0$, so that the 
    third requirement of \citet[Theorem 4.9]{guhring21} is satisfied. Hence, the theorem applies, yielding the rate $O(n\epsilon^{-(n+1)})$.
\end{enumerate}
\end{proof}

\continuoustimeapproximation*

\begin{proof} Fix $x_0\in \X_0$.  We decompose the error between the true trajectory $x(t,x_0)$ and the numerical approximation of the learned one $\hat{x}_k(x_0)$, over the time grid $\{t_k\}_{k=0}^{\Nt}$, into two terms, by introducing the continuous-time flow map $\hat{x}(t,x_0)$ of the learned system, as
    \begin{align*}
    \sup_{0\leq k\leq \Nt} |x(t_k, x_0) - \hat{x}_k(x_0)| &\leq \sup_{0\leq k\leq \Nt}|x(t_k, x_0) - \hat{x}(t_k, x_0)| + \sup_{0\leq k\leq \Nt} |\hat{x}(t_k, x_0) - \hat{x}_k(x_0)|,\\
    &\leq \|x(\cdot, x_0) - \hat{x}(\cdot, x_0)\|_{L^\infty([0,T]; \R^n)} + C(\hat{f},T)\dt^p,
    \end{align*}
    where the bound over the second term follows from the order of convergence of the RK scheme, with $C(\hat{f},T)$ a constant depending on the learned velocity field $\hat{f}$ and the final time $T$.
    By Theorem \ref{thm:universal-approximation-flow}, there exists a deep neural network $\hat{f} \in \mathcal{N}_\varrho(\R^{n+1}; \R^n)$ such that $\sup_{x_0\in\X_0}\|x(\cdot, x_0) - \hat{x}(\cdot, x_0)\|_{L^\infty([0,T]; \R^n)} \leq \epsilon/2$. Then, choosing $\dt^* = \left (\frac{\epsilon}{2C(\hat{f},T)}\right )^{1/p}$, the second term is bounded by $\epsilon/2$ for all $\dt \leq \dt^*$, yielding the result.
\end{proof}

\section{Proofs of Section \ref{sec:velocity-matching}}
\label{app:velocity-matching}

\subsection{Proof of Proposition \ref{prop:vm-loss-formulation}}
\label{app:velocity-matching-1}

\vmlossformulation*

\begin{proof}
The result follows from additivity of the $L^2$-norm over the partition $\{t_k\}_{k=0}^{\Nt-1}$ of the temporal interval $[0,T]$, and a change of variable $t = t_k + \tau$ with $\tau \sim \mathcal{U}[0, \dt]$ over each subinterval $[t_k, t_{k+1}]$, namely
\begin{align*}
   \tlf(\hat{f};x_0) &= \frac{1}{T}\| \partial_t\tilde{x}(\cdot, x_0) - \hat{f}(\cdot, \tilde{x}(\cdot, x_0))\|^2_{L^2([0,T];\R^n)}, \\
   &= \frac{1}{T}\int_0^T |\partial_t\tilde{x}(t, x_0) - \hat{f}(t, \tilde{x}(t, x_0))|^2 dt, \\
   &\overset{\eqref{eq:piecewise-linear-interpolant}}{=} \frac{1}{T}\sum_{k=0}^{\Nt-1}\int_{t_k}^{t_{k+1}} |v_k(x_0) - \hat{f}(t, x(t_k,x_0) + (t-t_k) v_k(x_0))|^2 dt, \\
   &= \frac{1}{T}\sum_{k=0}^{\Nt-1}\int_{0}^{\dt} |v_k(x_0) - \hat{f}(t_k+\tau, x(t_k,x_0) + \tau v_k(x_0))|^2 d\tau, \\
   &= \frac{1}{\Nt}\sum_{k=0}^{\Nt-1}\E_{\tau\sim \mathcal{U}[0, \dt]}|v_k(x_0) - \hat{f}(t_k+\tau, x(t_k,x_0) + \tau v_k(x_0))|^2,\\
   &=\lVM(\hat{f};x_0).
\end{align*}
\end{proof}

\subsection{Proof of Proposition \ref{prop:vm-loss-decomposition}, \ref{prop:vm-csrk}}
\label{app:velocity-matching-2}

\vmlossdecomposition*

\begin{proof}
The decomposition follows from the identity $\E|y - Z|^2 = |y - \E[Z]|^2 + \var[Z]$ for any constant vector $y\in\R^n$ and random variable $Z\in\R^n$, of which we report the derivation for completeness:
\begin{align*}
    \E |y - Z|^2 &= \E |y - \E[Z] + \E[Z] - Z|^2, \\
    &= \E |y - \E[Z]|^2 + 2\E\ \langle y - \E[Z], \E[Z] - Z\rangle + \E |\E[Z] - Z|^2, \\
    &= |y - \E[Z]|^2 + \E |\E[Z] - Z|^2, \\
    &= |y - \E[Z]|^2 + \var[Z].
\end{align*}

The result then follows by applying the identity with $y = v_k(x_0)$ and $Z = \hat{f}_k(\tau;x_0)$, and averaging over $k=0,\dots, \Nt-1$, with $x_0 \in \X_0$ fixed.
The regularity of $\hat{f}\in\mathcal{F}$ gives joint continuity of $\hat{f}$ over $[0,T]\times K$, thus ensuring that $\E_\tau[\hat{f}_k(\tau;x_0)]$ and $\var_\tau[\hat{f}_k(\tau;x_0)]$ are well-defined. 
\end{proof}

Prior to proving Proposition \ref{prop:vm-csrk}, we recall the definition of continuous-stage Runge-Kutta (csRK) methods, which generalize $s$-stage Runge-Kutta schemes to a continuum of stages, following \citet{hairer10} and \citet{tang18}. 

\begin{definition}[Continuous-stage Runge-Kutta methods] Let $a(\gamma, \sigma)$, $b(\gamma)$, and $c(\gamma)$ be continuous functions of $\gamma, \sigma \in [0,1]$, with $c(\gamma) = \int_0^1 a(\gamma, \sigma) d\sigma$. The one-step method $x_k \mapsto x_{k+1}$, applied to \eqref{eq:ode}, defined by
    \begin{align}
            X(\gamma) &= x_k + \dt \int_0^1 a(\gamma, \sigma) f(t_k + c(\sigma)\dt, X(\sigma)) \, d\sigma, \label{eq:csrk-stage}\\
            x_{k+1} &= x_k + \dt \int_0^1 b(\gamma) f(t_k + c(\gamma)\dt, X(\gamma)) \, d\gamma, \label{eq:csrk-update}
    \end{align}
    is a continuous-stage Runge-Kutta (csRK) method, with continuous-stage variable $X(\gamma)$.
\end{definition}
Here, the continuous-stage tableau $(a(\gamma, \sigma), b(\gamma), c(\gamma))$ generalizes the coefficients $(a_{ij}, b_i, c_i)$ of a classical Runge-Kutta method.
In the autonomous setting, $X(\gamma)$ identifies with the collocation polynomial \citep{hairer10}, yielding the continuous approximation of the solution $X(\gamma) \approx x(t_k + \gamma \dt)$ over the subinterval $[t_k, t_{k+1}]$. Given the relation \eqref{eq:csrk-stage}, the resulting method is implicit.

\vmcsrk*

\begin{proof}
The first two steps of the proof establish the equality between the bias term in \eqref{eq:vm-loss-decomposition} and the residual of a csRK method with tableau \eqref{eq:vm-csrk-tableau}. The third step establishes the bounds \eqref{eq:vm-csrk-residual-bound} relating the bias term to the TF residual associated with the implicit csRK update.
\begin{enumerate}
    \item \textit{Bias reformulation.} Considering the expectation over the random offset $\tau\sim\mathcal{U}[0,\dt]$, by the change of variables $\tau = \gamma \dt$,
    \begin{align*}
        \E_\tau [\hat{f}_k(\tau;x_0)]
        &= \frac{1}{\dt}\int_{0}^{\dt} \hat{f}(t_k +\tau, x(t_k,x_0) + \tau v_k(x_0))\, d\tau,\\
        &= \int_{0}^{1} \hat{f}(t_k +\gamma\dt, x(t_k,x_0) + \gamma\dt v_k(x_0))\, d\gamma, \\
        &= \int_{0}^{1} \hat{f}(t_k +\gamma\dt, (1- \gamma )x(t_k,x_0) + \gamma x(t_{k+1},x_0)) \, d\gamma,
    \end{align*}
    the last equality following from $\dt v_k(x_0) = x(t_{k+1},x_0) - x(t_k,x_0)$ as in \eqref{eq:piecewise-linear-interpolant}.
    Inserting this expression into the bias term in \eqref{eq:vm-loss-decomposition}, we obtain
    \begin{multline}
        \big |v_k(x_0) - \E_\tau [\hat{f}_k(\tau;x_0)] \big|^2
        =\frac{1}{\dt^2} \bigg|x(t_{k+1},x_0)- x(t_k,x_0) \\ - \dt \int_{0}^{1} \hat{f}(t_k +\gamma\dt, (1- \gamma )x(t_k,x_0) + \gamma x(t_{k+1},x_0) ) d\gamma \bigg|^2,
        \label{eq:vm-bias-term-reformulation}
    \end{multline}

    \item \textit{Continuous-stage Runge-Kutta residual.} Let us consider the csRK method (\ref{eq:csrk-stage}-\ref{eq:csrk-update}) applied to the learned dynamics $\hat{f}$, namely
    \begin{align*}
            X(\gamma) &= \hat{x}_k + \dt \int_0^1 a(\gamma, \sigma) \hat{f}(t_k + c(\sigma)\dt, X(\sigma)) \, d\sigma, \\
            \hat{x}_{k+1} &= \hat{x}_k + \dt \int_0^1 b(\gamma) \hat{f}(t_k + c(\gamma)\dt, X(\gamma)) \, d\gamma.
    \end{align*}
    The specific choice of continuous-stage tableau $a(\gamma, \sigma) = \gamma$, $b(\gamma) = 1$, and $c(\gamma) = \gamma$, for $\gamma, \sigma \in [0,1]$, yields the one-step method
    \begin{align}
        \hat{x}_{k+1} &= \hat{x}_k + \dt \int_0^1  \hat{f}(t_k + \gamma\dt, (1-\gamma)\hat{x}_k + \gamma \hat{x}_{k+1}) \, d\gamma,\nonumber\\
        &:= \hat{x}_k + \dt F(t_k, \hat{x}_k, \hat{x}_{k+1}, \dt; \hat{f}),
        \label{eq:csrk-update-avf}
    \end{align}
    coinciding with the \textit{average vector field} method, with $F$ denoting the increment function associated with the scheme.
    Substituting the observed pair $(x(t_k,x_0), x(t_{k+1},x_0))$ into the csRK update \eqref{eq:csrk-update-avf}, yields the residual
    \begin{equation}
        \varepsilon_{k+1}^{\rm VM}(x_0) = x(t_{k+1},x_0) - x(t_k,x_0) - \dt F(t_k, x(t_k,x_0), x(t_{k+1},x_0), \dt; \hat{f})
        \label{eq:csrk-residual}
    \end{equation}
    and comparison with \eqref{eq:vm-bias-term-reformulation} gives
    \begin{equation}
        \big |v_k(x_0) - \E_\tau [\hat{f}_k(\tau;x_0)] \big|^2 = \frac{1}{\dt^2} |\varepsilon_{k+1}^{\rm VM}(x_0)|^2,
        \label{eq:vm-csrk-residual-equality}
    \end{equation}
    which proves the first assertion.

    \item \textit{Equivalence with implicit TF update.}
    In a teacher forcing setting, the learned dynamics $\hat{f}$ is advanced by a single step from the observed state $x(t_k,x_0)$, over the subinterval $[t_k, t_{k+1}]$, yielding the implicit update
    \begin{equation}
        \hat{x}_{k+1} = x(t_k,x_0) + \dt F(t_k, x(t_k,x_0), \hat{x}_{k+1}, \dt; \hat{f}),
        \label{eq:tf-csrk-update}
    \end{equation}
    whose solution defines the residual $\varepsilon_{k+1}^{\rm TF}(x_0) = x(t_{k+1},x_0) - \hat{x}_{k+1}$, having denoted by $F$ the increment function associated with the csRK method. Hence, TF first solves the implicit equation for $\hat{x}_{k+1}$, and then computes the residual, carrying the nonlinear solve in the forward pass. On the other hand, when addressing VM bias minimization, the residual \eqref{eq:csrk-residual} is computed directly given the observed pair, without solving the implicit equation for $\hat{x}_{k+1}$. We thus proceed to derive the bounds relating the two residuals, showing equivalence between the two approaches.
    First, $F$ is Lipschitz continuous with respect to its third argument, with Lipschitz constant at most $M/2$, since
    \begin{align*}
        |F_\dt(t, y, z; \hat{f}) - F_\dt(t, y, z'; \hat{f})| &\leq \int_0^1 | \hat{f}(t + \gamma\dt, (1-\gamma)y + \gamma z) - \hat{f}(t + \gamma\dt, (1-\gamma)y + \gamma z')| d\gamma, \\ 
        &\leq M \int_0^1 \gamma |z-z'| d\gamma = \frac{M}{2} |z-z'|.
    \end{align*}
    Hence, the map $z \mapsto x(t_k,x_0) + \dt F(t_k, x(t_k,x_0), z, \dt; \hat{f})$ is Lipschitz, and a contraction whenever $\dt M < 2$. Thus, by Banach fixed point theorem, the implicit equation~\eqref{eq:tf-csrk-update} admits a unique solution $\hat{x}_{k+1}\in \bar{B}(x(t_{k+1}),\delta)$, so the TF update is well defined.
    Then, subtracting the two residuals
    \begin{equation}
        \varepsilon_{k+1}^{\rm VM} - \varepsilon_{k+1}^{\rm TF} = \dt \left ( F(t_k, x(t_k), x(t_{k+1}), \dt; \hat{f}) - F(t_k, x(t_k), \hat{x}_{k+1}, \dt; \hat{f}) \right ),
        \label{eq:residual-difference}
    \end{equation}
    by the triangle inequality and Lipschitz continuity of $F$, the lower bound follows
    \begin{align*}
        | \varepsilon_{k+1}^{\rm VM}| &\leq |\varepsilon_{k+1}^{\rm TF}| + \dt \left |F(t_k, x(t_k), x(t_{k+1}), \dt; \hat{f}) - F(t_k, x(t_k), \hat{x}_{k+1}, \dt; \hat{f}) \right |,\\
        &\leq |\varepsilon_{k+1}^{\rm TF}| + \frac{\dt M}{2} |x(t_{k+1}) - \hat{x}_{k+1}|, \\
        &= |\varepsilon_{k+1}^{\rm TF}| + \frac{\dt M}{2} |\varepsilon_{k+1}^{\rm TF}|, \\
        &= \left(1 + \frac{\dt M}{2}\right) |\varepsilon_{k+1}^{\rm TF}|,
    \end{align*}
    yielding
    \begin{equation*}
        \frac{1}{1 + \frac{1}{2}\dt M} |\varepsilon_{k+1}^{\rm VM}| \leq |\varepsilon_{k+1}^{\rm TF}|,
    \end{equation*}
    where dependence on $x_0$ is omitted for notational convenience.
    As for the upper bound, reversing \eqref{eq:residual-difference}, we have
    \begin{align*}
        | \varepsilon_{k+1}^{\rm TF}| &\leq |\varepsilon_{k+1}^{\rm VM}| + \dt \left |F(t_k, x(t_k), x(t_{k+1}), \dt; \hat{f}) - F(t_k, x(t_k), \hat{x}_{k+1}, \dt; \hat{f}) \right |,\\
        &\leq |\varepsilon_{k+1}^{\rm VM}| + \frac{\dt M}{2} |x(t_{k+1}) - \hat{x}_{k+1}|, \\
        &= |\varepsilon_{k+1}^{\rm VM}| + \frac{\dt M}{2} |\varepsilon_{k+1}^{\rm TF}|,
    \end{align*}
    which gives 
    \begin{equation*}
        |\varepsilon_{k+1}^{\rm TF}| \leq \frac{1}{1 - \frac{1}{2}\dt M} |\varepsilon_{k+1}^{\rm VM}|.
    \end{equation*}
    Substituting $|\varepsilon_{k+1}^{\rm VM}| = \dt |v_k(x_0) - \E_\tau [\hat{f}_k(\tau;x_0)]|$ from \eqref{eq:vm-csrk-residual-equality} yields \eqref{eq:vm-csrk-residual-bound}, concluding the proof.
\end{enumerate}
\end{proof}

\subsection{Proof of Theorem~\ref{thm:vm-order}, Corollary \ref{cor:vm-control}, \ref{cor:vm-consistency}}
\label{app:velocity-matching-3}

First, we recall some basic estimates about piecewise linear interpolation, which will be used in the proof of Theorem~\ref{thm:vm-order}. 
\begin{lemma}[Chord estimates]
    \label{lemma:chord-estimates}
    Under assumption \ref{ass:vm-regularity-1}, for every $k=0,\dots, \Nt-1$, $t\in[t_k, t_{k+1}]$ and $x_0\in \X_0$, it holds that
    \begin{equation}
        |x(t,x_0) - \tilde{x}(t,x_0)| \leq M_1\dt^2, \qquad |v_k(x_0) - f(t,x(t,x_0))| \leq M_1\dt, \qquad |v_k(x_0)| \leq M,
        \label{eq:chord-estimates}
    \end{equation}
    with $M_1 := M(1+M)$.
\end{lemma}

\begin{proof} 
    We first show that, under Assumption~\ref{ass:vm-regularity-1}, the map $t\mapsto f(t,x(t,x_0))$ is Lipschitz continuous.
        \begin{align*}
            |f(t,x(t,x_0)) - f(s,x(s,x_0))| 
            &\leq \int_s^t \left | \frac{d}{dr} f(r,x(r,x_0)) \right | dr, \\
            &\leq (\|\partial_t f\|_{L^\infty} + \|\partial_x f\|_{L^\infty}\|f \|_{L^\infty}) |t-s|,\\
            &\leq M(1+M)|t-s|,
        \end{align*}
        so that $M(1+M)$ bounds its Lipschitz constant, with the norm $\|\cdot\|_{L^\infty}$ taken over $[0,T] \times K$ for $f$ and over $[0,T]$ for $x(\cdot, x_0)$.
    \begin{enumerate}
    \item The above estimate bounds the Lipschitz constant of $\partial_t x(\cdot, x_0)$, so that the error of the piecewise linear interpolant $\tilde{x}(\cdot, x_0)$ can be bounded as
    \begin{align*}
    |x(t,x_0) - \tilde{x}(t,x_0)| &=  \left|\int_{t_k}^t \left ( \partial_s x(s,x_0) - v_k(x_0)\right ) ds\right|, \\
    &= \left|\int_{t_k}^t \left ( \partial_s x(s,x_0) - \frac{1}{\dt}\int_{t_k}^{t_{k+1}} \partial_r x(r,x_0) dr\right ) ds\right|, \\
    &= \left|\int_{t_k}^t \frac{1}{\dt}\int_{t_k}^{t_{k+1}} \left ( \partial_s x(s,x_0) - \partial_r x(r,x_0)\right ) dr ds\right|, \\
    &\leq \frac{1}{\dt}\int_{t_k}^{t_{k+1}}\int_{t_k}^{t_{k+1}} |\partial_s x(s,x_0) - \partial_r x(r,x_0)| dr ds, \\
    &\leq \frac{1}{\dt}\int_{t_k}^{t_{k+1}}\int_{t_k}^{t_{k+1}} M(1+M) |s-r| dr ds, \\
    &\leq M_1\dt^2.
    \end{align*}
    \item The second estimate follows from the fundamental theorem of calculus, and by the previously established Lipschitz continuity of the map $t\mapsto f(t,x(t,x_0))$, with
    \begin{align*}
        |v_k(x_0) - f(t,x(t,x_0))|
        &= \left|\frac{1}{\dt}\int_{t_k}^{t_{k+1}} f(s,x(s,x_0)) ds - f(t,x(t,x_0))\right|, \\
        &\leq \frac{1}{\dt}\int_{t_k}^{t_{k+1}} |f(s,x(s,x_0)) - f(t,x(t,x_0))| ds, \\
        &\leq \frac{1}{\dt}\int_{t_k}^{t_{k+1}} M(1+M) |s-t| ds, \\ 
        &\leq M(1+M) \dt.
    \end{align*}
    \item The third estimate trivially follows from Assumption \ref{ass:vm-regularity-1}, since
    \begin{equation*}
        |v_k(x_0)|  \leq \frac{1}{\dt}\int_{t_k}^{t_{k+1}} |f(s,x(s,x_0))| ds \leq \|f\|_{L^\infty([0,T] \times K; \R^n)} \leq M.
    \end{equation*}
    \end{enumerate}
\end{proof}

\vmorder*

\begin{proof} We first establish the decomposition of the error, and then bound the two terms separately.
    \begin{enumerate}
        \item \textit{Decomposition of the error.}  By Proposition \ref{prop:vm-loss-decomposition}, we have
        \begin{align*}
            \lVM(\hat{f}; x_0) &= \frac{1}{\Nt} \sum_{k=0}^{\Nt-1}\left ( \big|v_k(x_0) - \E_\tau [\hat{f}_k(\tau;x_0)] \big|^2 + \var_\tau [\hat{f}_k(\tau;x_0)] \right ),
        \end{align*}
        with $\hat{f}_k(\tau;x_0) := \hat{f}(t_k +\tau, x(t_k,x_0) + \tau v_k(x_0))$.
        The theoretical loss \eqref{eq:theoretical-velocity-objective} admits an equivalent representation, by decomposing the integral over the subintervals $[t_k, t_{k+1}]$, as
        \begin{align*}
            \lf(\hat{f}; x_0) &= \frac{1}{T} \int_{0}^{T} |f(t,x(t,x_0)) - \hat{f}(t,x(t,x_0))|^2 dt,\\
            &= \frac{1}{T} \sum_{k=0}^{\Nt-1} \int_{t_k}^{t_{k+1}} |f(t,x(t,x_0)) - \hat{f}(t,x(t,x_0))|^2 dt,\\
            &= \frac{1}{T} \sum_{k=0}^{\Nt-1} \int_{0}^{\dt} |f(t_k + \tau,x(t_k + \tau,x_0)) - \hat{f}(t_k + \tau,x(t_k + \tau,x_0))|^2 d\tau,\\
            &= \frac{1}{\Nt} \sum_{k=0}^{\Nt-1} \E_{\tau} |f(t_k + \tau,x(t_k + \tau,x_0)) - \hat{f}(t_k + \tau,x(t_k + \tau,x_0))|^2,\\
            &=: \frac{1}{\Nt} \sum_{k=0}^{\Nt-1} \E_{\tau} |\zeta_k(\tau;x_0)|^2.
        \end{align*}
        Using $ \E_{\tau} |\zeta_k(\tau;x_0)|^2 = |\E_{\tau} [\zeta_k(\tau;x_0)]|^2 + \var_\tau [\zeta_k(\tau;x_0)]$, it follows that
        \begin{equation}
            \lf(\hat{f}; x_0) = \frac{1}{\Nt} \sum_{k=0}^{\Nt-1} \left( |\E_{\tau} [\zeta_k(\tau;x_0)]|^2 + \var_\tau [\zeta_k(\tau;x_0)] \right),
            \label{eq:theoretical-velocity-loss-decomposition}
        \end{equation}
        from which
        \begin{multline*}
            \lVM(\hat{f}; x_0) - \lf(\hat{f}; x_0) = \frac{1}{\Nt} \sum_{k=0}^{\Nt-1} \Big ( \underbrace{|v_k(x_0) - \E_\tau [\hat{f}_k(\tau;x_0)]|^2  - |\E_{\tau} [\zeta_k(\tau;x_0)]|^2}_{A_k}  \\ + \underbrace{\var_\tau [\hat{f}_k(\tau;x_0)] - \var_\tau [\zeta_k(\tau;x_0)]}_{B_k} \Big ).
        \end{multline*}
        We proceed to bound the two terms $A_k$ and $B_k$ separately, showing that both are of order $O(\dt^2)$.
        \item \textit{Bound on $A_k$.} The bias term can be rewritten as
        \begin{align*}
            v_k(x_0) - \E_\tau [\hat{f}_k(\tau;x_0)] &= \frac{1}{\dt}\int_{t_k}^{t_{k+1}} \left (f(t,x(t,x_0)) - \hat{f}(t,\tilde{x}(t,x_0)) \right ) dt,\\
            &= \frac{1}{\dt}\int_{t_k}^{t_{k+1}} \left (f(t,x(t,x_0)) - \hat{f}(t,x(t,x_0)) \right ) dt \\ &\hspace{5em} + \frac{1}{\dt}\int_{t_k}^{t_{k+1}} \left (\hat{f}(t,x(t,x_0)) - \hat{f}(t,\tilde{x}(t,x_0))\right ) dt,\\
            &= \E_\tau [\zeta_k(\tau;x_0)] + \E_\tau [\beta_k(\tau;x_0)],
        \end{align*}
        having adopted the change of variables $t = t_k + \tau$, with $\tau \sim \mathcal{U}[0, \dt]$, and $\beta_k$ indicating the argument of the second integral above. Hence, it can be bounded as
        \begin{align*}
            |A_k| &= \left | |\E_\tau [\zeta_k(\tau;x_0)] + \E_\tau [\beta_k(\tau;x_0)]|^2 - |\E_\tau [\zeta_k(\tau;x_0)]|^2 \right |, \\
            &= \left ||\E_\tau [\zeta_k(\tau;x_0)]|^2 + 2  \langle \E_\tau [\zeta_k(\tau;x_0)], \E_\tau [\beta_k(\tau;x_0)]  \rangle + |\E_\tau [\beta_k(\tau;x_0)]|^2 - |\E_\tau [\zeta_k(\tau;x_0)]|^2 \right |, \\
            &\leq 2   |\E_\tau [\zeta_k(\tau;x_0)]| | \E_\tau [\beta_k(\tau;x_0)]| + |\E_\tau [\beta_k(\tau;x_0)]|^2.
        \end{align*}
        By Lipschitz continuity of $\hat{f}$ and the chord estimate \eqref{eq:chord-estimates}, the second term can be bounded as
        \begin{align*}
            |\E_\tau [\beta_k(\tau;x_0)]| &\leq \frac{1}{\dt}\int_{t_k}^{t_{k+1}} |\hat{f}(t,x(t,x_0)) - \hat{f}(t,\tilde{x}(t,x_0))| dt, \\
            &\leq \frac{M}{\dt}\int_{t_k}^{t_{k+1}} |x(t,x_0) -  \tilde{x}(t,x_0)| dt, \\
            &\leq \frac{M}{\dt}\int_{t_k}^{t_{k+1}} M(1+M)\dt^2 dt = M^2(1+M) \dt^2,
        \end{align*}
        while
        \begin{align*}
            |\E_\tau [\zeta_k(\tau;x_0)]| &\leq \frac{1}{\dt}\int_{t_k}^{t_{k+1}} | f(t,x(t,x_0)) - \hat{f}(t,x(t,x_0)) | dt, \\
            &\leq \|f\|_{L^\infty([0,T]\times K; \R^n)} + \|\hat{f}\|_{L^\infty([0,T]\times K; \R^n)} \leq 2M.
        \end{align*}
        Therefore, independently of $k$, we have that
        \begin{equation}
            |A_k| \leq 4 M^3(1+M) \dt^2 + M^4(1+M)^2 \dt^4.
            \label{eq:bound-Ak}
        \end{equation}
        
        \item \textit{Bound on $B_k$.} Given a Lipschitz continuous function $g:[0,\dt] \to \R^n$, it holds that
        \begin{equation*}
            \var_\tau [g(\tau)] = \E_\tau |g(\tau) - \E_\tau [g(\tau)]|^2 \leq \E_\tau \left |g(\tau) - g\left (\frac{\dt}{2}\right )\right |^2 \leq \lip{g}^2\var_\tau[\tau] = \lip{g}^2 \frac{\dt^2}{12}.
        \end{equation*}
        Since both maps $\tau \mapsto \hat{f}_k(\tau;x_0)$ and $\tau \mapsto \zeta_k(\tau;x_0)$ are Lipschitz continuous, with Lipschitz constants respectively bounded by $M_1$ and $2M_1$, we have that
        \begin{align*}
            0 \leq \var_\tau[\hat{f}_k(\tau;x_0)] \leq M_1^2 \frac{\dt^2}{12}, \qquad 0 \leq \var_\tau[\zeta_k(\tau;x_0)] \leq M_1^2 \frac{\dt^2}{3},
        \end{align*}
        so that 
        \begin{equation}
            |B_k| = |\var_\tau [\hat{f}_k(\tau;x_0)] - \var_\tau [\zeta_k(\tau;x_0)]| \leq M_1^2\frac{\dt^2}{3}.
            \label{eq:bound-Bk}
        \end{equation}
    \end{enumerate}
    Combining the bounds \eqref{eq:bound-Ak} and \eqref{eq:bound-Bk} concludes the proof.
\end{proof}

For completeness, we also provide the following result, establishing $O(\dt)$ convergence of the teacher forcing (TF) loss \eqref{eq:teacher-forcing-velocity-formulation} to the theoretical loss.

\begin{theorem}[Order of convergence, TF loss]
\label{thm:tf-order}
Under assumption \ref{ass:vm-regularity-1}, for every $x_0 \in \X_0$, it holds that
    \begin{equation}
        |\lTF(\hat{f}; x_0) - \ell_f(\hat{f}; x_0) | \leq C^{\rm TF}_1 \dt + C^{\rm TF}_2\dt^2,
        \label{eq:tf-order}
    \end{equation}
with $C^{\rm TF}_1 = 4M^2(1+M)$, and $C^{\rm TF}_2 = \frac{4}{3}M^2(1+M)^2$.
\end{theorem}

\begin{proof}
Considering the $k$-th term in the TF loss \eqref{eq:teacher-forcing-velocity-formulation}, we can achieve a similar decomposition to that of $\lVM$ in the proof of Theorem~\ref{thm:vm-order}, as follows
\begin{equation*}
    v_k(x_0) - \hat{f}(t_k, x(t_k,x_0)) = \E_\tau [\zeta_k(\tau;x_0)] + \E_\tau [\beta_k(\tau;x_0)], 
\end{equation*}
with $\zeta_k$ defined as in the proof of Theorem~\ref{thm:vm-order}, namely
\begin{equation*}
    \zeta_k(\tau;x_0) := f(t_k + \tau,x(t_k + \tau,x_0)) - \hat{f}(t_k + \tau,x(t_k + \tau,x_0)),
\end{equation*}
while $\beta_k$ is defined as
\begin{equation*}
    \beta_k(\tau;x_0) := \hat{f}(t_k +\tau,x(t_k + \tau,x_0)) - \hat{f}(t_k,x(t_k,x_0)).
\end{equation*}
Thus, the TF loss can be expressed as
\begin{align*}
    | v_k(x_0) - \hat{f}(t_k, x(t_k,x_0))|^2 &= |\E_\tau [\zeta_k(\tau;x_0)]|^2 + |\E_\tau [\beta_k(\tau;x_0)]|^2 + 2 \langle \E_\tau [\zeta_k(\tau;x_0)], \E_\tau [\beta_k(\tau;x_0)] \rangle,
\end{align*}
which, combined with the theoretical loss decomposition \eqref{eq:theoretical-velocity-loss-decomposition}, yields
\begin{align*}
    | v_k(x_0) - \hat{f}(t_k, x(t_k,x_0))|^2 - \E_\tau |\zeta_k(\tau;x_0)|^2 &=  |\E_\tau [\beta_k(\tau;x_0)]|^2 \\ &+ 2 \langle \E_\tau [\zeta_k(\tau;x_0)], \E_\tau [\beta_k(\tau;x_0)] \rangle  - \var_\tau [\zeta_k(\tau;x_0)].
\end{align*}
Denoting by $M_1 := M(1+M)$, by Lipschitz continuity of the map $\tau \mapsto \hat{f}(t_k +\tau,x(t_k + \tau,x_0))$, then $|\E_\tau [\beta_k(\tau;x_0)]| \leq M_1 \dt$, while $|\E_\tau [\zeta_k(\tau;x_0)]| \leq 2M$. The variance term follows the same bound as in the proof of Theorem~\ref{thm:vm-order}, so that $|\var_\tau [\zeta_k(\tau;x_0)]| \leq \frac{1}{3}M_1^2 \dt^2$. Hence,
\begin{align*}
    | | v_k(x_0) - \hat{f}(t_k, x(t_k,x_0))|^2 - \E_\tau |\zeta_k(\tau;x_0)|^2 | &\leq M_1^2 \dt^2 + 4 M M_1 \dt + \frac{1}{3}M_1^2 \dt^2,\\
    &\leq 4M^2(1+M)\dt +  \frac{4}{3} M^2(1+M)^2 \dt^2, 
\end{align*}
\end{proof}

\vmcontrol*

\begin{proof}
    We first show that the learned trajectory remains confined in $K$, so that Lemma~\ref{lemma:continuous-time-stability-estimates} applies with $\Lambda \leq M$, and then combine the resulting estimate with Theorem~\ref{thm:vm-order}.
    \begin{enumerate}
        \item \textit{Confinement.} Fix $x_0 \in \X_0$, and set
        \begin{equation*}
            \xi(t):=C_{\infty,2}(M)\|f(\cdot,x(\cdot,x_0)) - \hat{f}(\cdot, x(\cdot,x_0))\|_{L^2([0,t];\R^n)}, \qquad t \in [0,T],
        \end{equation*}
        which is continuous and non-decreasing in $t$, with $\xi(0) = 0$, and $\xi(T)<\delta$ by assumption. We claim that $\hat{x}(t,x_0) \in K$ for all $t \in [0,T]$. Since $\hat{f}$ is locally Lipschitz in the second argument, the flow map $\hat{x}(t,x_0)$ is well-defined on a maximal interval $[0,T_{\max})$, with $T_{\max}>0$ \citep[Theorem~2.13]{teschl12}. 
        We proceed by contradiction, assuming that there exists a first exit time $t^* \in (0,\min\{T,T_{\max}\})$ such that $\hat{x}(t^*,x_0) \in \partial K$. Therefore, the distance from the convex hull of the reachable set $\conv(\X)$ is exactly $\delta$, i.e. $\dist(\hat{x}(t^*,x_0),\conv(\X)) = \delta$, and since $\X \subseteq \conv(\X)$, then
        \begin{align}
            \delta = \dist(\hat{x}(t^*,x_0),\conv(\X)) &\leq \dist(\hat{x}(t^*,x_0),\X), \nonumber \\ 
            &= \inf_{y \in \X}|\hat{x}(t^*,x_0) - y|, \nonumber\\
            &\leq |\hat{x}(t^*,x_0) - x(t^*,x_0)| \label{eq:dist-bound}.
        \end{align}
        On $[0,t^*]$ both trajectories lie in $K$, with $K$ convex, so that $L^\infty$-$L^2$ stability estimates in Lemma~\ref{lemma:continuous-time-stability-estimates} apply on $[0,t^*]$ with $\Lambda \leq M$, yielding
        \begin{equation*}
            |\hat{x}(t^*,x_0) - x(t^*,x_0)|\leq \xi(t^*) \leq \xi(T) < \delta,
        \end{equation*}
        by monotonicity of $\xi$, contradicting \eqref{eq:dist-bound}. Therefore, $\hat{x}(t,x_0) \in K$ for all $t \in [0,T]$, and Lemma \ref{lemma:continuous-time-stability-estimates} applies with $\Lambda \leq M$.

        \item \textit{Objective control.} By the first step and the $L^2$-$L^2$ stability estimate of Lemma~\ref{lemma:continuous-time-stability-estimates}, following the argument of Corollary \ref{cor:theoretical-objective-control}, then
        \begin{equation*}
            \lx(\hat{f}; x_0) \leq C_{2,2}^2(M) \lf(\hat{f};x_0) \leq C_{2,2}^2(M) \left ( \lVM(\hat{f}; x_0) + C^{\rm VM}_1 \dt^2 + C^{\rm VM}_2\dt^4 \right ),
        \end{equation*}
        the second inequality following from Theorem~\ref{thm:vm-order}. Since $x_0$ is arbitrary, the estimate holds for every $x_0 \in \X_0$.
    \end{enumerate}
\end{proof}

\vmconsistency*

\begin{proof}
    The empirical occupation measure $\mu_N$, is defined as $\mu_N := \frac{1}{NT}\sum_{i=1}^N \int_0^T x(t,\cdot)_\# \delta_{x_0^{(i)}} dt$. Since the estimate of Theorem~\ref{thm:vm-order} is uniform in $x_0 \in \X_0$, it is legitimate to average over $x_0 \sim \rho_0$, yielding
    \begin{equation}
    \left |\LVM[\hat{f}] - \|f - \hat{f}\|_{L^2(\mu_N; \R^n)}^2\right | \leq C^{\rm VM}_1 \dt^2 + C^{\rm VM}_2\dt^4.
    \label{eq:vm-order-empirical-measure}
    \end{equation}
    Applying \eqref{eq:vm-order-empirical-measure} at $\hat{f} = \hat{f}_\ast$ yields
    \begin{align*}
        \|f-\hat{f}_\ast\|_{L^2(\mu_N; \R^n)}^2 &\leq \LVM[\hat{f}_\ast] + C^{\rm VM}_1 \dt^2 + C^{\rm VM}_2\dt^4, \\
        &\leq \LVM[f] + C^{\rm VM}_1 \dt^2 + C^{\rm VM}_2\dt^4, \\
        &\leq \|f-f\|_{L^2(\mu_N; \R^n)}^2 + 2(C^{\rm VM}_1 \dt^2 + C^{\rm VM}_2\dt^4), \\
        & = 2( C^{\rm VM}_1 \dt^2 + C^{\rm VM}_2\dt^4 ),
    \end{align*}
    where the second inequality follows from the fact that $\hat{f}_\ast$ is a minimizer of $\LVM$, and the third inequality follows from \eqref{eq:vm-order-empirical-measure} applied a second time with $\hat{f} = f$, with the conclusion following trivially.
\end{proof}

\subsection{Proof of Proposition \ref{prop:vm-implicit-regularization}}
\label{app:velocity-matching-4}

\implicitregularization*

\begin{proof} 
    We proceed by Taylor expansion with integral remainder, similarly as in the proof of Theorem~\ref{thm:vm-order} in Appendix \ref{app:velocity-matching-3}, denoting by $\partial_x f$ the Jacobian of $f$, and similarly for $\hat{f}$. By \citet[Section~5.8.2, Theorem 4]{evans10} applied to $g$ and $\partial_x g$, any $g\in\mathcal{F}$ admits a representative in $C^1(K;\R^n)$, with $|g|$, $|\partial_x g|$, $\lip{g}$ and $\lip{\partial_x g}$ bounded by $M$ on $K$. Such bounds hold both along the trajectory and along the interpolant, as $K$ contains the convex hull of $\X$.
    \begin{enumerate}
    \item \textit{Bias term.} First consider the velocity target, which is a first-order finite difference. We notice that $\partial_t^2 x(t,x_0) = \partial_x f(x(t,x_0))f(x(t,x_0))$, and the map $t \mapsto \partial_t^2 x(t,x_0)$ is Lipschitz continuous with constant $2M^3$, namely, for every $t,s \in [0,T]$,
    \begin{align*}
        |\partial_t^2 x(t,x_0) - \partial_t^2 x(s,x_0)| &= |\partial_x f(x(t,x_0))f(x(t,x_0)) - \partial_x f(x(s,x_0))f(x(s,x_0))|, \\
        &\leq |\partial_x f(x(t,x_0))||f(x(t,x_0)) - f(x(s,x_0))| \\&\qquad + |\partial_x f(x(t,x_0)) - \partial_x f(x(s,x_0))||f(x(s,x_0))|,\\
        &\leq 2 M^3|t-s|.
    \end{align*}
    Hence, it follows that
    \begin{align}
        v_k(x_0) &= \frac{1}{\dt}\int_0^\dt \partial_t x(t_k + s, x_0) \, ds, \nonumber\\
        &= \partial_t x(t_k,x_0) + \frac{\dt}{2} \partial_t^2 x(t_k,x_0) + \frac{1}{\dt} \int_0^\dt \int_0^s \left ( \partial_t^2 x(t_k + r, x_0) - \partial_t^2 x(t_k, x_0) \right  ) \, dr \, ds, \nonumber\\
        &= f(x(t_k,x_0)) + \frac{\dt}{2} \partial_x f(x(t_k,x_0))f(x(t_k,x_0)) + O(\dt^2), \label{eq:target-velocity-expansion}
    \end{align}
    with the remainder bounded by $M^3 \dt^2/3$ from the Lipschitz continuity of $t\mapsto \partial_t^2 x(t,x_0)$.
    Expanding the learned velocity field along the interpolant, about $x(t_k,x_0)$, we have
    \begin{align}
        \hat{f}_k(\tau;x_0) &= \hat{f}(x(t_k,x_0) + \tau v_k(x_0)),\nonumber\\
         &=\hat{f}(x(t_k,x_0)) + \tau \partial_x \hat{f}(x(t_k,x_0))v_k(x_0)\nonumber\\
         &\qquad + \int_0^\tau \left (\partial_x \hat{f}(x(t_k,x_0) + s v_k(x_0)) - \partial_x \hat{f}(x(t_k,x_0))\right )v_k(x_0) \, ds,\nonumber\\
         &=\hat{f}(x(t_k,x_0)) + \tau \partial_x \hat{f}(x(t_k,x_0))f(x(t_k,x_0)) + O(\dt^2), \label{eq:learned-velocity-expansion}
    \end{align}
    uniformly in $\tau\in[0,\dt]$, where the integral remainder is bounded by $M|v_k(x_0)|^2\tau^2/2\leq M^3\dt^2/2$, recalling that $|v_k(x_0)|\leq M$ from Lemma~\ref{lemma:chord-estimates}, and where replacing $v_k(x_0)$ with $f(x(t_k,x_0))$ in the first-order term introduces an error bounded by  $\tau M|v_k(x_0) - f(x(t_k,x_0))| \leq M^3\dt^2$, by Lemma~\ref{lemma:chord-estimates} applied in the autonomous setting. Taking the expectation with respect to $\tau$, with $\E_\tau[\tau]=\dt/2$, yields
    \begin{equation}
        \E_\tau[\hat{f}_k(\tau;x_0)] = \hat{f}(x(t_k,x_0)) + \frac{\dt}{2} \partial_x \hat{f}(x(t_k,x_0))f(x(t_k, x_0)) + O(\dt^2), \label{eq:expected-learned-velocity-expansion}
    \end{equation}
    from which we obtain
    \begin{multline*}
     v_k(x_0) - \E_\tau[\hat{f}_k(\tau;x_0)] = f(x(t_k,x_0)) - \hat{f}(x(t_k,x_0)) \\
     + \frac{\dt}{2}\left (\partial_x f(x(t_k,x_0)) - \partial_x \hat{f}(x(t_k,x_0))\right )f(x(t_k,x_0))  + O(\dt^2).
    \end{multline*}
    Taking the squared norm, it holds that
    \begin{multline*}
        | v_k(x_0) - \E_\tau[\hat{f}_k(\tau;x_0)]|^2 = |f(x(t_k,x_0)) - \hat{f}(x(t_k,x_0))|^2 \\
         + \frac{\dt^2}{4} \left | \left  (\partial_x f(x(t_k,x_0)) - \partial_x \hat{f}(x(t_k,x_0)) \right ) f (x(t_k,x_0))  \right |^2  \\
         + \dt \left \langle f(x(t_k,x_0)) - \hat{f}(x(t_k,x_0)), \left (\partial_x f(x(t_k,x_0)) - \partial_x \hat{f}(x(t_k,x_0))\right )f(x(t_k,x_0)) \right \rangle \\
         + 2\langle f(x(t_k,x_0)) - \hat{f}(x(t_k,x_0)), O(\dt^2)\rangle + O(\dt^3).
    \end{multline*}
    Then \eqref{eq:bias-expansion} follows, since, under $|f(x(t_k,x_0)) - \hat{f}(x(t_k,x_0))| \leq C\dt^2$, the zero-order term is bounded by $C^2\dt^4$, while, by Cauchy--Schwarz, the cross-terms are absorbed into $O(\dt^3)$.

    \item \textit{Variance term.} Subtracting \eqref{eq:expected-learned-velocity-expansion} from \eqref{eq:learned-velocity-expansion}, the zero-order term cancels, and we obtain
    \begin{align*}
        \hat{f}_k(\tau;x_0) - \E_\tau[\hat{f}_k(\tau;x_0)] &= \left(\tau - \frac{\dt}{2}\right)\partial_x \hat{f}(x(t_k,x_0))f(x(t_k,x_0)) + O(\dt^2),
    \end{align*}
    from which, taking squared norm and expectation, we obtain
    \begin{align*}
        \var_\tau[\hat{f}_k(\tau;x_0)] &= \E_\tau \left [ \left(\tau - \frac{\dt}{2}\right)^2 \right ] \left|\partial_x \hat{f}(x(t_k,x_0))f(x(t_k,x_0))\right|^2 +  O(\dt^3) ,\\
        &= \frac{\dt^2}{12}|\partial_x \hat{f}(x(t_k,x_0))f(x(t_k,x_0))|^2 + O(\dt^3),
    \end{align*}
    having used the fact that $\var_\tau[\tau] = \dt^2/12$.
\end{enumerate}
\end{proof}

\section{Experimental details}
\label{app:experimental-details}

This appendix details the experiments of Section~\ref{sec:numerical-results}. Section~\ref{app:benchmarks} describes the benchmark systems and the generation of the datasets, Section~\ref{app:architecture} the model architectures, and
Section~\ref{app:training} the training and evaluation procedures. Section~\ref{app:additional-results} provides additional material supporting our results, including the tables of validation and test errors for ordinary differential equations (ODEs) and partial differential equations (PDEs) benchmarks, together with visualizations of the solutions. All experiments were run on a single NVIDIA L40S GPU with 48GB of memory.
The code to reproduce the experiments is available at \href{https://github.com/farenga/velocity-matching}{github.com/farenga/velocity-matching}.

\subsection{Benchmark systems and datasets}
\label{app:benchmarks}

\subsubsection{Ordinary differential equations}
\label{app:benchmarks-ode}

For each ODE system, training, validation and test trajectories were generated from independently sampled initial conditions, by integrating the true dynamics with the DOP853 scheme \citep{hairer08}, with absolute and relative tolerances equal to $10^{-12}$, and were observed on a uniform grid of step size $\dt$ over $[0,T]$. An additional set of test trajectories, observed with step size $10^{-4}$, was used for the continuous-time approximation study of Section~\ref{sec:numerical-results-2}. The number of trajectories, the step sizes and the time horizons of each split are reported in Table~\ref{table:datasets-ode}. Note that test trajectories extend over a time horizon twice as long as the training one, so that test errors also measure extrapolation in time.

\begin{table}[!ht]
\centering
\begin{tabular}{lcccccccccccc}
\toprule
 & \multicolumn{3}{c}{\textbf{Training}} & \multicolumn{3}{c}{\textbf{Validation}} & \multicolumn{3}{c}{\textbf{Test}} & \multicolumn{3}{c}{\textbf{Test} (fine)}\\
\cmidrule(lr){2-4}\cmidrule(lr){5-7}\cmidrule(lr){8-10}\cmidrule(lr){11-13}
\textbf{System} & $N$ & $\dt$ & $T$ & $N$ & $\dt$ & $T$ & $N$ & $\dt$ & $T$ & $N$ & $\dt$ & $T$ \\
\midrule
Euler & 100 & $5\cdot10^{-2}$ & $20$ & 20 & $10^{-2}$ & $20$ & 50 & $5\cdot10^{-3}$ & $40$ & 10 & $10^{-4}$ & $ 20$ \\
Stuart--Landau & 50 & $5\cdot10^{-2}$ & $10$ & 10 & $10^{-2}$ & $10$ & 50 & $2\cdot10^{-3}$ & $20$ & 10 & $10^{-4}$ & $10$ \\
Active M\"uller--Brown & 200 & $5\cdot10^{-2}$ & $10$ & 20 & $10^{-2}$ & $10$ & 50 & $2\cdot10^{-3}$ & $20$ & 10 & $10^{-4}$ & $10$ \\
Lorenz & 80 & $\dt$ & $10$ & -- & -- & -- & 50 & $10^{-3}$ & $20$ & -- & -- & -- \\
\bottomrule
\end{tabular}
\caption{\textbf{ODE datasets.} Number of trajectories $N$, observation step size $\dt$ and time horizon $T$ of each split. For the Lorenz system, the training step size takes the values $\dt \in \{1, 2, 5, 10, 20, 50, 100\}\cdot 10^{-3}$, used to assess the Jacobian spectrum at $C_{-}$, reported in Figure \ref{fig:euler-lorenz}.}
\label{table:datasets-ode}
\end{table}

We report the formulation of the ODE benchmarks below, specifying for each system its governing equations, the parametric configuration, and the distribution $\rho_0$ of the initial conditions.

\paragraph{Euler.} The angular momentum $x\in \R^3$ of a free rigid body evolves according to a Lie--Poisson system \citep{marsden99, hairer10}, reading as
\begin{equation*}
    \dot{x} = x \times \mathrm{I}^{-1} x, \qquad \mathrm{I} = \operatorname{diag}(I_1, I_2, I_3).
\end{equation*}
The kinetic energy $H(x) = \frac12 x^\top \mathrm{I}^{-1} x$ and the Casimir invariant $C(x) = |x|^2$ are conserved, so that trajectories lie on the unit sphere in $\R^3$. We set $(I_1, I_2, I_3) = (2, 1, 2/3)$, and sample the initial conditions uniformly on the unit sphere. Since $|x|$ is conserved, the trajectories carry no information on the radial component of the velocity field, the Jacobian error~\eqref{eq:jac-error} is computed in the tangent directions only, namely as the error between $\partial_x f(x) P(x)$ and $\partial_x \hat{f}(x) P(x)$, where $P(x) = \operatorname{Id} - xx^\top/|x|^2$ is the orthogonal projector onto the tangent space of the sphere at $x$.

\paragraph{Stuart--Landau.} The normal form of a supercritical Hopf bifurcation \citep{perko01} reads as
\begin{equation*}
    \dot{x}_1 = \mu x_1 - x_2 + (a x_1 - b x_2)\,r^2, \qquad \dot{x}_2 = x_1 + \mu x_2 + (b x_1 + a x_2)\,r^2, \qquad r^2 = x_1^2 + x_2^2,
\end{equation*}
and, for $\mu > 0$ and $a < 0$, admits a stable limit cycle of radius $\sqrt{-\mu/a}$. We set $(\mu, a, b) = (1, -1, 1)$, so that the limit cycle is the unit circle, and sample the initial conditions in polar coordinates, with radius $r_0 \sim \mathcal{U}[0.2, 2]$ and angle $\theta_0 \sim \mathcal{U}[0, 2\pi]$, so that trajectories approach the limit cycle from both inside and outside.

\paragraph{Active M\"uller--Brown system.} We consider an active particle of unit mass \citep{romanczuk12}, with position $x\in\R^2$, subject to the M\"uller--Brown potential \citep{muller79} and to Rayleigh--Helmholtz friction, namely
\begin{equation*}
    \ddot{x} = -\kappa \nabla V(x) + (\alpha - \beta |\dot{x}|^2) \dot{x},
\end{equation*}
where
\begin{equation*}
    V(x) = \sum_{i=1}^4 A_i \exp\left \{ a_i (x_1 - \bar{x}_{1,i})^2 + b_i (x_1 - \bar{x}_{1,i})(x_2 - \bar{x}_{2,i}) + c_i (x_2 - \bar{x}_{2,i})^2 \right \},
\end{equation*}
with the standard coefficients of \citet{muller79}. The friction term injects energy at low speed and dissipates it at high speed, and vanishes at the characteristic speed $v_0 = \sqrt{\alpha/\beta}$. We set $(\kappa, \alpha, \beta) = (10^{-2}, 0.25, 1)$, so that $v_0 = 0.5$. With this choice, the system is characterized by self-sustained oscillations within each of the two deepest wells of the potential, whereas the intermediate minimum is only visited transiently. Initial positions are sampled uniformly in $[-1.5, 1.2]\times[-0.2, 2]$, and retained only if $V(q_0) < 0$, which keeps each trajectory within its initial well. Initial velocities have magnitude $v_0$ and a uniformly sampled direction. The resulting velocity field spans several orders of magnitude over the domain.

\paragraph{Lorenz system.} The Lorenz system \citep{lorenz63} reads as
\begin{equation*}
    \dot{x}_1 = \sigma(x_2 - x_1), \qquad \dot{x}_2 = x_1(\rho - x_3) - x_2, \qquad \dot{x}_3 = x_1 x_2 - \beta x_3,
\end{equation*}
and is considered with the classical parameters $(\sigma, \rho, \beta) = (10, 28, 8/3)$, for which the dynamics is chaotic, with the nontrivial equilibria $C_\pm = (\pm\sqrt{\beta(\rho-1)}, \pm\sqrt{\beta(\rho-1)}, \rho - 1)$ being unstable \citep{sparrow82}. Initial conditions are drawn from a localized distribution $\mathcal{N}(\bar{x}_0, 10^{-4} I_3)$, with $\bar{x}_0 = (-9, -3, 34)$, so that the separation induced by the chaotic nature of the system is the only driver of the exploration of the attractor.

\subsubsection{Partial differential equations}
\label{app:benchmarks-pde}

The considered PDE test cases are posed on a one dimensional domain $\Omega = [0,L_\Omega)$, with interval length $L_\Omega$ and periodic boundary conditions, and discretized with $N_h$ equispaced grid points, so that the resulting discretized state is $u_h \in \R^{N_h}$. Time stepping is performed via Fourier pseudo-spectral, second-order exponential time-differencing Runge--Kutta schemes provided by \cite{koehler24}.
Initial conditions are random superpositions of $k$ sine waves, with independent coefficients. Trajectories were generated with a fine time discretization and then subsampled in time for training, while validation and test trajectories were kept at the generation step size.

\begin{table}[!ht]
\centering
\begin{adjustbox}{max width=\textwidth}
\begin{tabular}{lccccccccccc}
\toprule
 & & & \multicolumn{3}{c}{\textbf{Training}} & \multicolumn{3}{c}{\textbf{Validation}} & \multicolumn{3}{c}{\textbf{Test}} \\
\cmidrule(lr){4-6}\cmidrule(lr){7-9}\cmidrule(lr){10-12}
\textbf{Equation} & $L_\Omega$ & $N_h$ & $N$ & $\dt$ & $T$ & $N$ & $\dt$ & $T$ & $N$ & $\dt$ & $T$ \\
\midrule
Burgers & 1 & 1024 & 100 & $10^{-2}$ & $0.5$ & 20 & $5\cdot 10^{-3}$ & $0.5$ & 50 & $10^{-3}$ & $1$ \\
Fisher--KPP & 1 & 1024 & 100 & $5\cdot 10^{-3}$ & $0.5$ & 20 & $5\cdot 10^{-4}$ & $0.5$ & 50 & $5\cdot 10^{-4}$ & $1$ \\
Korteweg--de Vries & 30 & 1024 & 100 & $2\cdot 10^{-2}$ & $2$ & 20 & $2\cdot 10^{-3}$ & $2$ & 50 & $10^{-3}$ & $4$ \\
Kuramoto--Sivashinsky & 64 & 256 & 100 & $1$ & $200$ & -- & -- & -- & 50 & $10^{-1}$ & $ 800$ \\
\bottomrule
\end{tabular}
\end{adjustbox}
\caption{\textbf{PDE datasets.} Domain length $L_\Omega$, number of grid points $N_h$, and number of trajectories $N$, step size $\dt$ and time horizon $T$ of each split. The training step size results from subsampling trajectories generated with a finer step size, by a factor $2$ for the Burgers equation and $10$ for the other equations.}
\label{table:datasets-pde}
\end{table}

We report the PDE benchmarks below, specifying for each equation the spatial domain, the parametric configuration, and the distribution of the initial conditions.

\paragraph{Burgers.} We consider the viscous Burgers equation, reading as
\begin{equation*}
    \partial_t u + u \partial_x u = \nu \partial_x^2 u,
\end{equation*}
with viscosity $\nu = 10^{-2}$, $L_\Omega=1$ and $k=2$, for which the initial waves develop into viscous fronts that propagate and decay over time.

\paragraph{Fisher--KPP.} We consider the following reaction-diffusion equation
\begin{equation*}
    \partial_t u = \nu \partial_x^2 u + r\, u(1-u),
\end{equation*}
characterized by a logistic growth term, combined with diffusion. We set the diffusivity $\nu = 10^{-2}$ and the reaction rate $r = 1$, on a domain of length $L_\Omega = 1$. Initial conditions, with $k=5$, are clipped to the range $[0,1]$.

\paragraph{Korteweg--de Vries.} We consider the hyperviscous Korteweg--de Vries (KdV) equation
\begin{equation*}
    \partial_t u - 6 u\partial_x u + \partial_x^3 u = -\zeta \partial_x^4 u,
\end{equation*}
with hyperviscosity coefficient $\zeta = 10^{-2}$, on a domain of length $L_\Omega=30$. The initial waves, sampled with $k=5$, break into solitary waves propagating at speeds depending on their amplitude, while the hyperviscous term damps small scales.

\paragraph{Kuramoto--Sivashinsky.} We consider the Kuramoto--Sivashinsky equation in conservative form, reading as
\begin{equation*}
    \partial_t u + u\,\partial_x u + \partial_x^2 u + \partial_x^4 u = 0,
\end{equation*}
on a domain of length $L_\Omega = 64$. 
For this domain length, the anti-diffusive second-order term, which destabilizes large scales, the fourth-order term, which dissipates small scales, and the nonlinear term, which transfers energy between them, together result in spatiotemporal chaos.

\subsection{Architectures and normalization}

We detail the architectures used to parameterize the learned velocity field in the ODE and PDE settings, together with the adopted normalization strategy.

\label{app:architecture}

\paragraph{ODEs.} The learned velocity field is a fully connected network of width $256$ for the active M\"uller--Brown system, and $64$ for the remaining test cases. Three residual connections are employed, with each residual block being a gated linear unit, in which a GELU branch is modulated by a $\tanh$ gate. The output layer is initialized to zero, so that training starts from the trivial dynamics $\hat{f} = 0$. Since all systems are autonomous, the network takes the state as its only input.

\paragraph{PDEs.} The learned velocity field is a convolutional residual network acting on the discretized state, as in \citet{stachenfeld22}. In particular, a first convolutional layer lifts the state to $64$ channels, which is then followed by a sequence of residual blocks with the same number of channels, and finally by a GELU activation and a convolution projecting back to a single channel. Each residual block is composed of two GELU activations and two convolutions, and, for the Burgers, Fisher--KPP and KdV equations, of a layer normalization over the channels, applied independently at each grid point before each convolution. We use four residual blocks for the Burgers, Fisher--KPP and KdV equations, and eight residual blocks for the Kuramoto--Sivashinsky equation. All convolutions have kernel size $5$ and circular padding, consistently with the periodic boundary conditions. The last convolution of each residual block and the output convolution are initialized to zero.

\paragraph{Asymmetric normalization.} The input and the output of the learned velocity field are of different nature, namely states and velocities, and are therefore normalized separately, as also noted by \citet{mccabe25}.
Denoting by $(\mu_x, \sigma_x)$ the mean and standard deviation of the training states, and by $(\mu_v, \sigma_v)$ those of the finite difference velocities $(x(t_{k+1}) - x(t_k))/\dt$ computed on the training trajectories, the learned velocity field $\hat{f}$ is parameterized as
\begin{equation*}
    \hat{f}(x) = \sigma_v \odot \tilde{f}\big((x - \mu_x) \oslash \sigma_x\big) + \mu_v,
\end{equation*}
where $\tilde{f}$ is one of the networks described above, and $\odot$ and $\oslash$ denote componentwise multiplication and division, respectively. For ODEs the statistics are computed componentwise, while for PDEs, they are computed over the whole discretized solution.

\subsection{Training and evaluation}
\label{app:training}

\paragraph{Training.} All models were trained with AdamW \citep{loshchilov18}, with weight decay $10^{-5}$ and, for a given system, the same number of iterations for all strategies. For the Euler, Stuart--Landau and Lorenz systems, a constant learning rate $\eta = 10^{-4}$, a batch size of $64$, and $2\cdot 10^4$ iterations were used. For the active M\"uller--Brown system, we used a batch size of $128$ and a cosine schedule decaying the learning rate from $10^{-3}$ to $10^{-6}$ over $5\cdot 10^4$ iterations. 
For the PDEs test cases, we used a batch size of $128$ and a cosine learning rate schedule from $10^{-4}$ to $10^{-7}$ over $2\cdot 10^4$ iterations for the Burgers and Fisher--KPP equations, $4\cdot 10^4$ for the KdV equation, and $10^4$ for the Kuramoto--Sivashinsky equation. These were preceded by $10^3$ warm-up iterations, during which the learning rate increased linearly from $10^{-6}$ to $10^{-4}$. Gradients were clipped to unit norm, except for the Kuramoto--Sivashinsky equation.

\paragraph{Validation.} Every $100$ iterations, the validation trajectory error~\eqref{eq:traj-error} was computed by simulating the learned dynamics with RK4, at the prescribed validation step size, over the full validation horizon. For the ODE benchmarks, the velocity and Jacobian errors~\eqref{eq:vel-error}--\eqref{eq:jac-error} were also computed along the validation trajectories, solely to assess their decay during training and to check the implicit Jacobian matching property of VM, and were not used for model selection.
For each run, the checkpoint attaining the lowest validation trajectory error was retained for testing. For the chaotic systems, no validation was performed, and the last model checkpoint was retained.

\paragraph{Testing.} Test trajectory errors were computed at the step size of the test data, with the implicit midpoint for the Euler and Stuart--Landau systems, and RK4 for all others. For the continuous-time approximation study, the reference trajectories were obtained from an additional test set generated over a fine discretization, sharing the same time horizon of the training set (Table~\ref{table:datasets-ode}), and then by linearly interpolating onto the inference grid.

\newpage
\subsection{Additional results}
\label{app:additional-results}

\begin{figure}
    \centering
    \includegraphics[width=.95\textwidth]{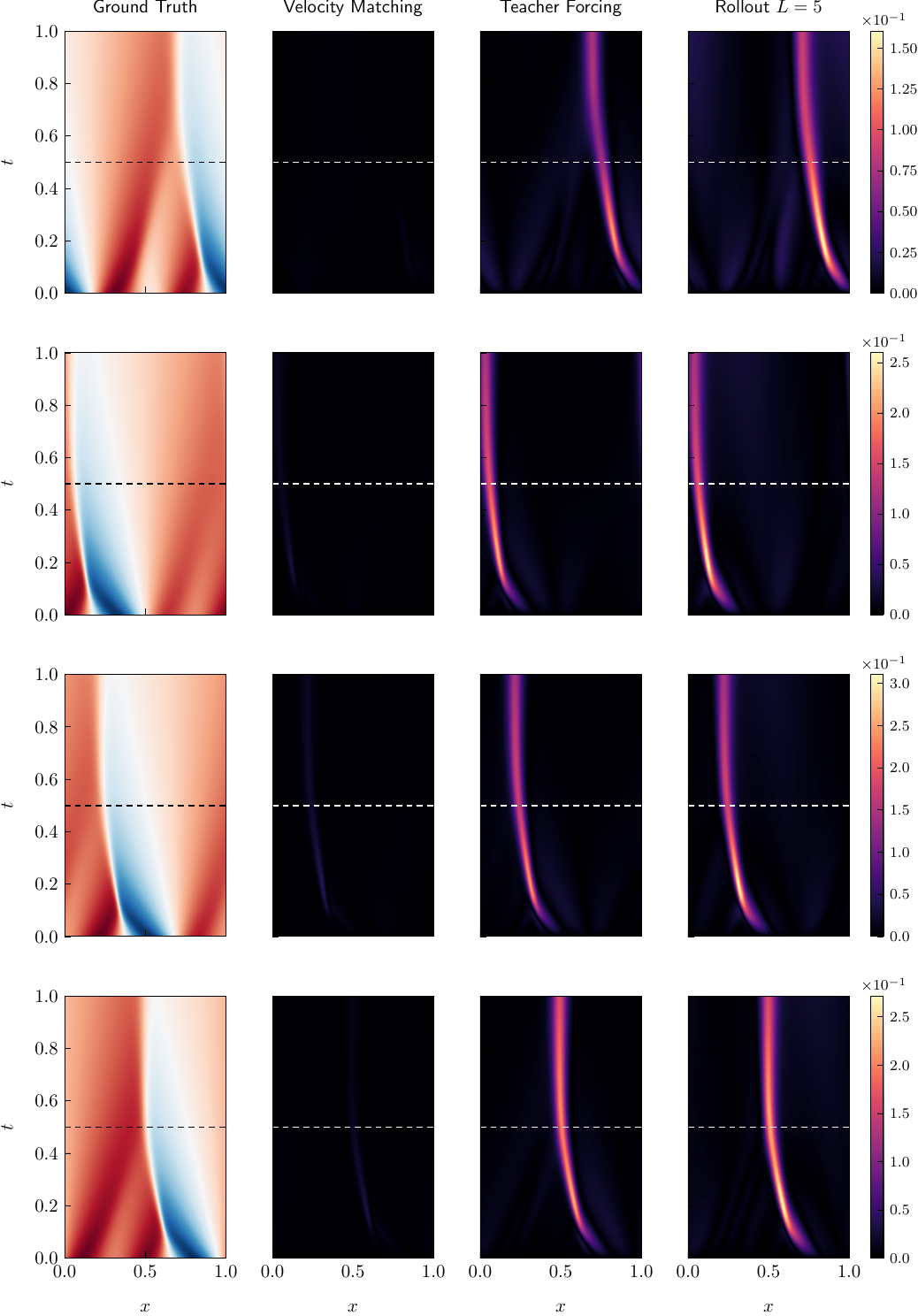}
    \caption{\textbf{Burgers.} The first column reports reference solutions for four test initial conditions. The remaining columns report the pointwise absolute error of the solutions learned by VM, TF, and rollout training with $L = 5$; the dashed line marks the end of the training horizon.}    
    \label{fig:burgers-errors}
\end{figure}

\begin{figure}
    \centering
    \includegraphics[width=0.95\textwidth]{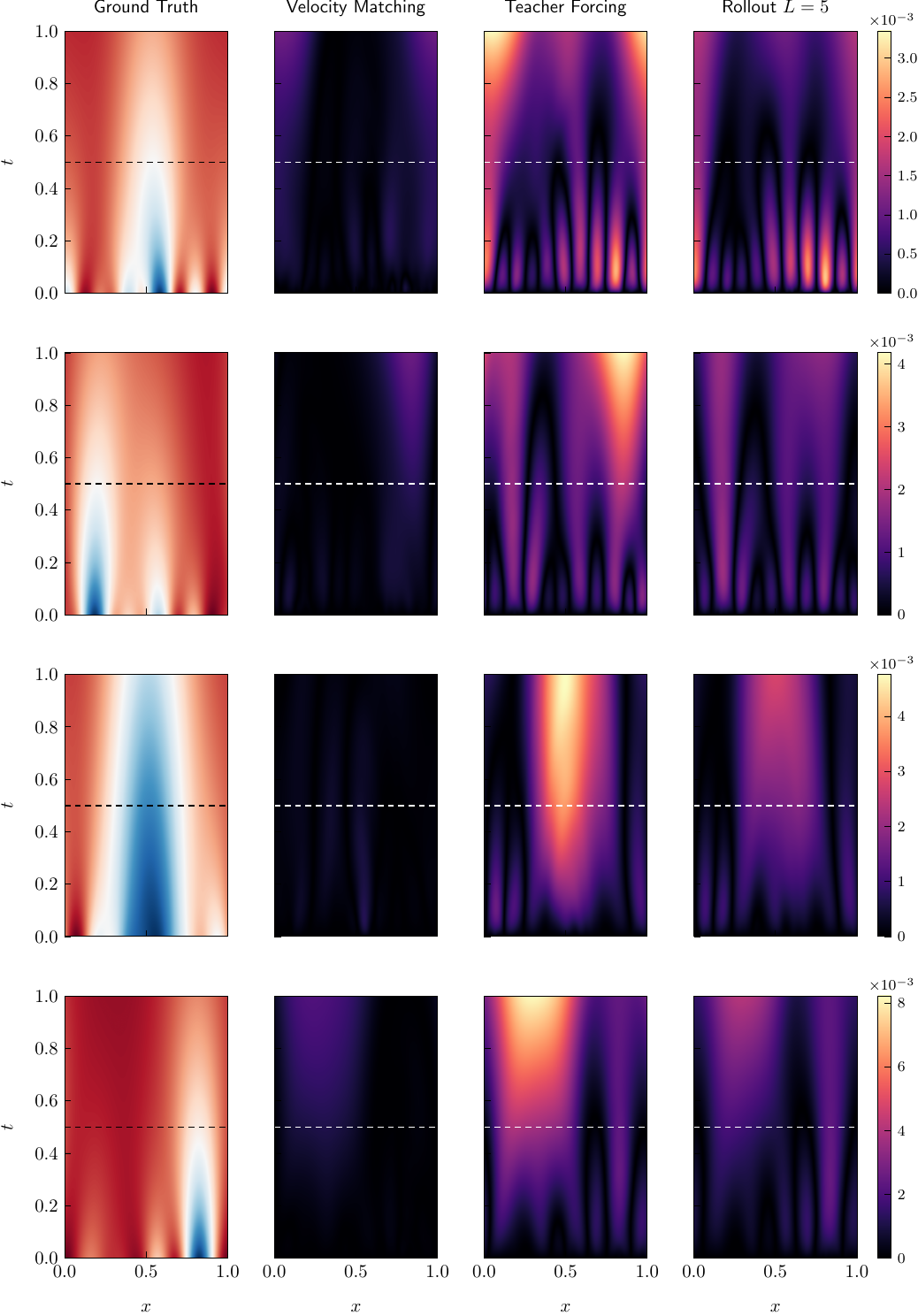}
    \caption{\textbf{Fisher--KPP.} The first column reports reference solutions for four test initial conditions. The remaining columns report the pointwise absolute error of the solutions learned by VM, TF, and rollout training with $L = 5$; the dashed line marks the end of the training horizon.}    \label{fig:fisher-kpp-errors}
\end{figure}

\begin{figure}
    \centering
    \includegraphics[width=.95\textwidth]{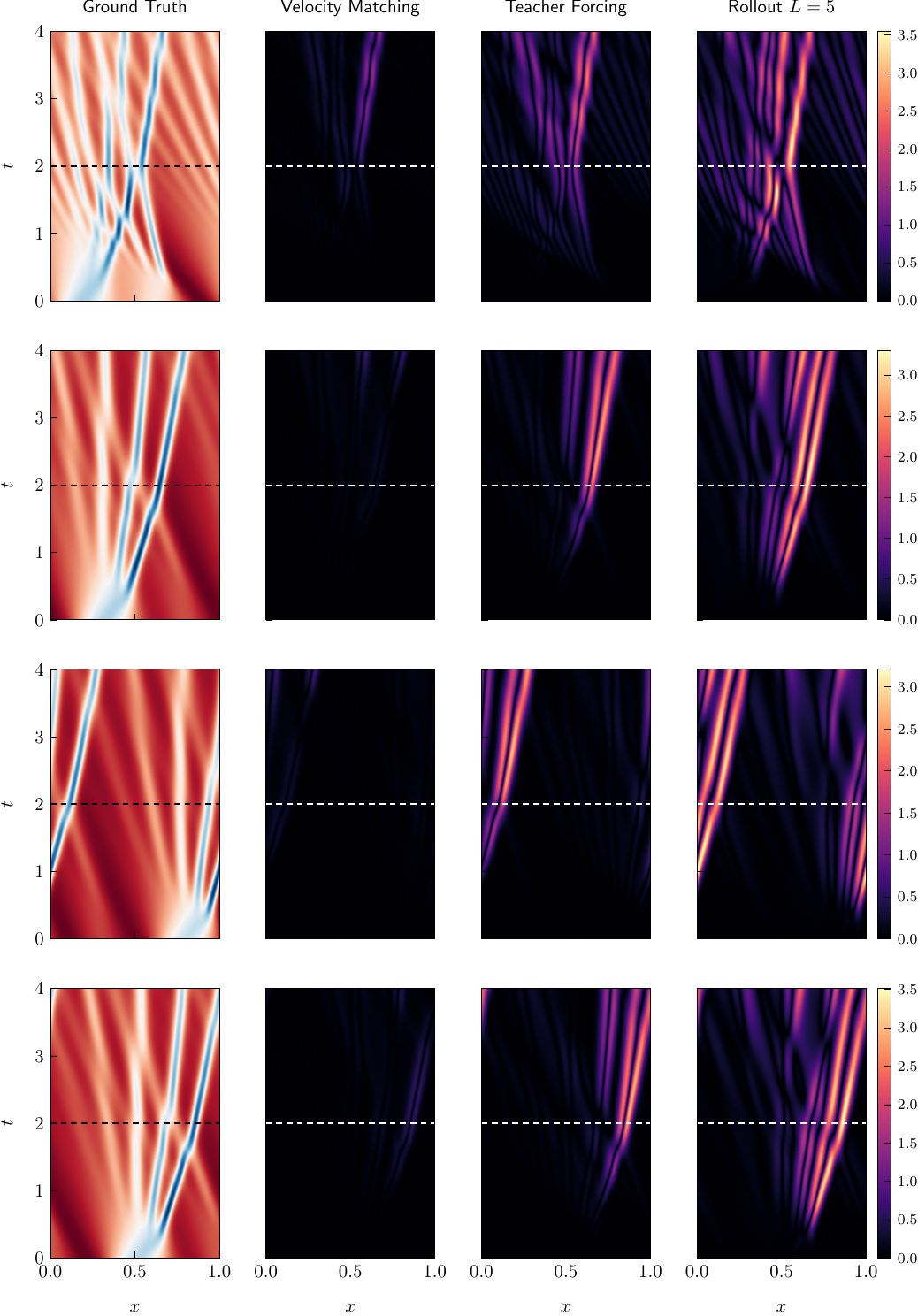}
    \caption{\textbf{Korteweg--de Vries.} The first column reports reference solutions for four test initial conditions. The remaining columns report the pointwise absolute error of the solutions learned by VM, TF, and rollout training with $L = 5$; the dashed line marks the end of the training horizon.}    
    \label{fig:kdv-errors}
\end{figure}

Table~\ref{table:results-ode} reports, for the Euler, Stuart--Landau, and active M\"uller--Brown systems, the validation and test values of the trajectory, velocity, and Jacobian errors~\eqref{eq:traj-error}--\eqref{eq:jac-error}, attained by teacher forcing, rollout training with fixed and scheduled shooting lengths $L \in \{5, 10, 20, 50\}$, alongside VM. Table~\ref{table:results-pde} reports the validation and test trajectory errors~\eqref{eq:traj-error} for the Burgers, Fisher--KPP, and KdV equations, for the same training strategies. In both tables, the validation errors are evaluated over the training horizon $T_\mathrm{train}$, and the test errors over the extended horizon $T_\mathrm{test} = 2T_\mathrm{train}$, as reported in Tables~\ref{table:datasets-ode} and \ref{table:datasets-pde}. 
Finally, Figures~\ref{fig:burgers-errors}--\ref{fig:kdv-errors} report, for each PDE, the reference solutions for four test initial conditions and the pointwise absolute errors of the solutions learned by VM, TF, and rollout training with $L=5$, both within and beyond the training horizon.

\begin{table}[!h]
\centering
\begin{adjustbox}{max width=\textwidth}
\begin{tabular}{llcccccc}
\toprule
 & & \multicolumn{2}{c}{\bf Trajectory} & \multicolumn{2}{c}{\bf Velocity} & \multicolumn{2}{c}{\bf Jacobian} \\
\cmidrule(lr){3-4}
\cmidrule(lr){5-6}
\cmidrule(lr){7-8}
{\bf Test Case} & {\bf Method} & Validation & Test & Validation & Test & Validation & Test \\
\midrule
 & Teacher Forcing & $5.60\cdot 10^{-2}$ & $1.51\cdot 10^{-1}$ & $1.48\cdot 10^{-2}$ & $1.58\cdot 10^{-2}$ & $2.13\cdot 10^{-2}$ & $2.04\cdot 10^{-2}$ \\
 & Rollout $L=5$ & $5.14\cdot 10^{-2}$ & $1.40\cdot 10^{-1}$ & $1.39\cdot 10^{-2}$ & $1.49\cdot 10^{-2}$ & $1.74\cdot 10^{-2}$ & $1.69\cdot 10^{-2}$ \\
 & Rollout $L=10$ & $5.27\cdot 10^{-2}$ & $1.43\cdot 10^{-1}$ & $1.40\cdot 10^{-2}$ & $1.49\cdot 10^{-2}$ & $1.73\cdot 10^{-2}$ & $1.67\cdot 10^{-2}$ \\
 & Rollout $L=20$ (Randomized) & $5.08\cdot 10^{-2}$ & $1.43\cdot 10^{-1}$ & $1.37\cdot 10^{-2}$ & $1.48\cdot 10^{-2}$ & $1.77\cdot 10^{-2}$ & $1.70\cdot 10^{-2}$ \\
 & Rollout $L=50$ (Curriculum) & $5.02\cdot 10^{-2}$ & $1.34\cdot 10^{-1}$ & $1.39\cdot 10^{-2}$ & $1.48\cdot 10^{-2}$ & $1.67\cdot 10^{-2}$ & $1.61\cdot 10^{-2}$ \\
\multirow{-6}{*}{\bf Euler} & \cellcolor{accent-red!20}{\bf Velocity Matching (Ours)} & \cellcolor{accent-red!20}$\bf 7.66\cdot 10^{-3}$ & \cellcolor{accent-red!20}$\bf 2.59\cdot 10^{-2}$ & \cellcolor{accent-red!20}$\bf 3.29\cdot 10^{-3}$ & \cellcolor{accent-red!20}$\bf 3.54\cdot 10^{-3}$ & \cellcolor{accent-red!20}$\bf 5.82\cdot 10^{-3}$ & \cellcolor{accent-red!20}$\bf 5.51\cdot 10^{-3}$ \\
\midrule
 & Teacher Forcing & $4.01\cdot 10^{-1}$ & $8.01\cdot 10^{-1}$ & $5.26\cdot 10^{-2}$ & $5.19\cdot 10^{-2}$ & $8.51\cdot 10^{-2}$ & $8.23\cdot 10^{-2}$ \\
 & Rollout $L=5$ & $4.17\cdot 10^{-1}$ & $8.27\cdot 10^{-1}$ & $5.26\cdot 10^{-2}$ & $5.21\cdot 10^{-2}$ & $8.48\cdot 10^{-2}$ & $8.23\cdot 10^{-2}$ \\
 & Rollout $L=10$ & $4.23\cdot 10^{-1}$ & $8.38\cdot 10^{-1}$ & $5.26\cdot 10^{-2}$ & $5.21\cdot 10^{-2}$ & $8.88\cdot 10^{-2}$ & $8.59\cdot 10^{-2}$ \\
 & Rollout $L=20$ (Randomized) & $2.06\cdot 10^{-1}$ & $4.30\cdot 10^{-1}$ & $5.41\cdot 10^{-2}$ & $5.38\cdot 10^{-2}$ & $8.71\cdot 10^{-2}$ & $8.40\cdot 10^{-2}$ \\
 & Rollout $L=50$ (Curriculum) & $4.21\cdot 10^{-1}$ & $8.35\cdot 10^{-1}$ & $5.34\cdot 10^{-2}$ & $5.27\cdot 10^{-2}$ & $8.73\cdot 10^{-2}$ & $8.39\cdot 10^{-2}$ \\
\multirow{-6}{*}{\bf Stuart--Landau} & \cellcolor{accent-red!20}{\bf Velocity Matching (Ours)} & \cellcolor{accent-red!20}$\bf 1.12\cdot 10^{-2}$ & \cellcolor{accent-red!20}$\bf 1.89\cdot 10^{-2}$ & \cellcolor{accent-red!20}$\bf 9.14\cdot 10^{-3}$ & \cellcolor{accent-red!20}$\bf 7.78\cdot 10^{-3}$ & \cellcolor{accent-red!20}$\bf 3.21\cdot 10^{-2}$ & \cellcolor{accent-red!20}$\bf 2.84\cdot 10^{-2}$ \\
\midrule
 & Teacher Forcing & $3.32\cdot 10^{-1}$ & $4.36\cdot 10^{-1}$ & $1.29\cdot 10^{-1}$ & $1.36\cdot 10^{-1}$ & $6.60\cdot 10^{-2}$ & $6.09\cdot 10^{-2}$ \\
 & Rollout $L=5$ & $3.20\cdot 10^{-1}$ & $4.24\cdot 10^{-1}$ & $1.20\cdot 10^{-1}$ & $1.26\cdot 10^{-1}$ & $6.03\cdot 10^{-2}$ & $5.55\cdot 10^{-2}$ \\
 & Rollout $L=10$ & $3.25\cdot 10^{-1}$ & $4.12\cdot 10^{-1}$ & $1.34\cdot 10^{-1}$ & $1.40\cdot 10^{-1}$ & $6.55\cdot 10^{-2}$ & $5.97\cdot 10^{-2}$ \\
 & Rollout $L=20$ (Randomized) & $3.11\cdot 10^{-1}$ & $4.09\cdot 10^{-1}$ & $1.30\cdot 10^{-1}$ & $1.35\cdot 10^{-1}$ & $7.79\cdot 10^{-2}$ & $7.27\cdot 10^{-2}$ \\
 & Rollout $L=50$ (Curriculum) & $3.12\cdot 10^{-1}$ & $4.25\cdot 10^{-1}$ & $1.27\cdot 10^{-1}$ & $1.32\cdot 10^{-1}$ & $6.19\cdot 10^{-2}$ & $5.61\cdot 10^{-2}$ \\
\multirow{-6}{*}{\bf M\"uller--Brown} & \cellcolor{accent-red!20}{\bf Velocity Matching (Ours)} & \cellcolor{accent-red!20}$\bf 1.61\cdot 10^{-2}$ & \cellcolor{accent-red!20}$\bf 2.59\cdot 10^{-2}$ & \cellcolor{accent-red!20}$\bf 5.06\cdot 10^{-3}$ & \cellcolor{accent-red!20}$\bf 5.05\cdot 10^{-3}$ & \cellcolor{accent-red!20}$\bf 7.61\cdot 10^{-3}$ & \cellcolor{accent-red!20}$\bf 7.70\cdot 10^{-3}$ \\
\bottomrule
\end{tabular}
\end{adjustbox}
\caption{\textbf{Validation and test errors: ODEs.} Relative trajectory, velocity, and Jacobian errors~\eqref{eq:traj-error}--\eqref{eq:jac-error} on the validation and test sets for the Euler, Stuart--Landau, and active M\"uller--Brown systems, averaged over three independent runs; best results in bold. VM attains the lowest errors on all three test cases, including the Jacobian error, consistently with the implicit Jacobian regularization of Proposition~\ref{prop:vm-implicit-regularization}.}
\label{table:results-ode}
\end{table}

\begin{table}[!h]
\centering
\begin{adjustbox}{max width=\textwidth}
\begin{tabular}{lcccccc}
\toprule
 & \multicolumn{2}{c}{\bf Burgers} & \multicolumn{2}{c}{\bf Fisher--KPP} & \multicolumn{2}{c}{\bf Korteweg--de Vries} \\
\cmidrule(lr){2-3}
\cmidrule(lr){4-5}
\cmidrule(lr){6-7}
 {\bf Method} & Validation & Test & Validation & Test & Validation & Test \\
\midrule
Teacher Forcing & $2.32\cdot 10^{-2}$ & $2.69\cdot 10^{-2}$ & $1.51\cdot 10^{-3}$ & $1.59\cdot 10^{-3}$ & $6.53\cdot 10^{-2}$ & $1.91\cdot 10^{-1}$ \\
Rollout $L=5$ & $3.57\cdot 10^{-2}$ & $5.07\cdot 10^{-2}$ & $1.49\cdot 10^{-3}$ & $1.39\cdot 10^{-3}$ & $5.15\cdot 10^{-1}$ & $1.14$ \\
Rollout $L=10$ & $1.00\cdot 10^{-1}$ & $2.05\cdot 10^{-1}$ & $1.43\cdot 10^{-3}$ & $1.23\cdot 10^{-3}$ & $1.16$ & $1.73$ \\
Rollout $L=20$ (Randomized) & $6.40\cdot 10^{-2}$ & $9.41\cdot 10^{-2}$ & $1.55\cdot 10^{-3}$ & $1.32\cdot 10^{-3}$ & $4.72\cdot 10^{-1}$ & $9.12\cdot 10^{-1}$ \\
Rollout $L=50$ (Curriculum) & $3.61\cdot 10^{-2}$ & $8.21\cdot 10^{-2}$ & $1.39\cdot 10^{-3}$ & $1.22\cdot 10^{-3}$ & $1.67\cdot 10^{-1}$ & $3.66\cdot 10^{-1}$ \\
\rowcolor{accent-red!20}{\bf Velocity Matching (Ours)} & $\bf 5.39\cdot 10^{-3}$ & $\bf 5.22\cdot 10^{-3}$ & $\bf 2.80\cdot 10^{-4}$ & $\bf 2.94\cdot 10^{-4}$ & $\bf 1.34\cdot 10^{-2}$ & $\bf 6.64\cdot 10^{-2}$ \\
\bottomrule
\end{tabular}
\end{adjustbox}
\caption{\textbf{Validation and test errors: PDEs.} Relative trajectory errors~\eqref{eq:traj-error} on the validation and test sets for the viscous Burgers, Fisher--KPP and KdV equations, averaged over three independent runs; best results in bold. VM attains the lowest errors on all three test cases, while, for the Burgers and KdV equations, increasing the rollout shooting length does not improve accuracy.}
\label{table:results-pde}
\end{table}

\newpage

\newpage

\newpage

\end{document}